\documentclass{article}

\usepackage[main, final]{neurips_2026}

\usepackage[utf8]{inputenc} 
\usepackage[T1]{fontenc}    
\usepackage{hyperref}       
\usepackage{booktabs}       
\usepackage{amsfonts}       
\usepackage{url}            
\usepackage{nicefrac}       
\usepackage{microtype}      
\usepackage{xcolor}         

\usepackage{amsmath,amssymb,amsthm}
\usepackage{mathtools}
\usepackage{graphicx}
\usepackage{wrapfig}  
\usepackage{booktabs}
\usepackage{multirow}
\usepackage{xcolor}
\usepackage{enumitem}
\usepackage{colortbl} 
\newcommand{\std}[1]{{\scriptsize$\pm$#1}} 
\usepackage{float}
\usepackage{subcaption}
\newtheorem{theorem}{Theorem}[section]
\newtheorem{proposition}[theorem]{Proposition}
\newtheorem{lemma}[theorem]{Lemma}

\newtheorem{definition}[theorem]{Definition}
\newtheorem{remark}[theorem]{Remark}

\newcommand{\R}{\mathbb{R}}

\newcommand{\bA}{\mathbf{A}}
\newcommand{\bD}{\mathbf{D}}

\newcommand{\bX}{\mathbf{X}}
\newcommand{\bP}{\mathbf{P}}
\newcommand{\bY}{\mathbf{Y}}
\newcommand{\bI}{\mathbf{I}}

\newcommand{\bW}{\mathbf{W}}
\newcommand{\bM}{\mathbf{M}}
\newcommand{\tA}{\tilde{\bA}}
\newcommand{\tD}{\tilde{\bD}}

\DeclareMathOperator{\erank}{Erank}
\DeclareMathOperator{\nrank}{NumRank}
\DeclareMathOperator{\diam}{diam}

\DeclareMathOperator{\rank}{rank}
\DeclareMathOperator{\diag}{diag}

\title{Aggregation Dispersion: An Information-Geometric\\ Diagnostic of Oversmoothing in Graph Neural Networks}

\author{%
  Abdessalam Ed-dib\thanks{Corresponding author.} \\
  Tandon School of Engineering\\
  New York University \\
  \texttt{ae2842@nyu.edu} 
  \And
  Amine Mohamed Aboussalah \\
  Tandon School of Engineering \\
  New York University \\
  \texttt{ama10288@nyu.edu} \\
}

\begin{document}

\maketitle
\begin{abstract}
Message-passing neural networks (MPNNs) learn node representations by iteratively aggregating information from neighbors, but stacking too many layers causes all representations to converge, a phenomenon known as oversmoothing. We propose a geometric perspective on this phenomenon. At each depth $k$, the propagation operator assigns to every node $v$ a walk distribution $p_v^{(k)}$, a probability distribution over all nodes that weights source-node contributions to node \(v\). These walk distributions form a point cloud on the probability simplex, whose contraction to a single point characterizes complete mixing of the propagation operator, and thus oversmoothing. To measure this contraction, we exploit the intrinsic geometry of the simplex: the Fisher--Rao metric, the unique Riemannian metric invariant up to a scale under sufficient statistics, whose scaled chordal distance under the square-root embedding $p \mapsto \sqrt{p}$ is the Hellinger distance. The resulting diagnostic, the \emph{aggregation dispersion} $\mathcal{D}_k$, is the variance of the embedded point cloud on the unit sphere. We prove that: (i) $\mathcal{D}_k = 0$ if and only if the $k$-th step propagation matrix has rank one, providing a necessary and sufficient condition for complete operator mixing; (ii) $\mathcal{D}_k$ is monotonically non-increasing with depth; (iii) the spectral gap of the propagation matrix controls the rate of this decay; and (iv) for contractive MPNNs, the aggregation dispersion upper-bounds the representation dispersion up to graph- and initialization-dependent constants, connecting the geometry of the simplex to the geometry of the feature space. In experiments across 14 model configurations and 11 datasets, $\mathcal{D}_k$ attains the highest mean absolute Pearson correlation with test accuracy across depths among eight diagnostics ($|r| = 0.721$, using log-transformed diagnostic values), numerically comparable to the post-training effective rank ($|r| = 0.688$), while requiring no training.
\end{abstract}

\section{Introduction}
\label{sec:intro}
Message-passing neural networks (MPNNs)~\citep{gilmer2017neural} learn node representations by iteratively aggregating information from neighboring nodes in a graph $G = (V, E)$ with $n = |V|$ nodes. In a fixed linear propagation model, this aggregation is governed by a \emph{propagation matrix}~$\bP \in \mathbb{R}^{n \times n}$, a non-negative matrix whose entry~$P_{uv}$ prescribes the weight that node~$u$ assigns to the message received from node~$v$. When every row of~$\bP$ sums to one, a property called \emph{row-stochasticity}, each row is a probability distribution over~$V$. After $k$ layers of propagation, the effective influence of all nodes on node~$v$ is captured by the $v$-th row of the $k$-th matrix power, $p_v^{(k)} = [\bP^k]_{v,:}$, which we call the \emph{walk distribution} from~$v$ at depth~$k$. Each walk distribution is a point on the \emph{probability simplex} $\Delta^{n-1} = \{p \in \mathbb{R}^n : p \geq 0,\; \mathbf{1}^\top p = 1\}$, the space of probability distributions over~$n$ outcomes.

For an irreducible, aperiodic row-stochastic propagation matrix $P$, the Perron--Frobenius theorem guarantees that every point on this simplex converges, as $k \to \infty$, to a unique \emph{stationary distribution}~$\pi$, satisfying $\pi^{\top}P=\pi^{\top}$~\citep{Horn_Johnson_2012}. This convergence means that the $n$~walk distributions $\{p_v^{(k)}\}_{v \in V}$ contract toward a single point~$\pi$ on~$\Delta^{n-1}$. At the point of collapse the matrix~$\bP^k$ has rank one, so every node receives an identical aggregated signal and a shared node-wise function of that signal alone cannot produce distinct node representations. This provides a propagation-driven mechanism for \emph{oversmoothing}~\citep{li2018deeper, oono2020graph}.

Detecting oversmoothing before it erases useful information is therefore a central challenge. Existing diagnostics, however, require model representations: the Dirichlet energy~\citep{cai2020note, digiovanni2023understanding} measures the smoothness of the learned node features, while the effective rank~\citep{deidda2025rethinking} tracks the effective dimensionality of the representation matrix. \citet{deidda2025rethinking} emphasize relating these diagnostics to accuracy degradation across depths, requiring separately trained models and their learned representations. We instead propose a training-free, depth-resolved measure defined solely by the propagation matrix~$\bP$.

The key observation is that the simplex already carries a rich geometric structure that can be exploited for exactly this purpose. The interior
of the simplex carries a canonical Riemannian metric, the \emph{Fisher--Rao
metric}~\citep{amari2016information}, singled out by \v{C}encov's theorem as
the unique (up to scale) metric invariant under sufficient statistics, so that
distances reflect genuine statistical differences rather than artifacts of the
parameterization. Under the square-root embedding $\phi:p\mapsto(\sqrt{p_1},\ldots,\sqrt{p_n})$, the simplex interior $(\operatorname{int}\Delta^{n-1},\frac14 g_{\mathrm{FR}})$ maps isometrically into the unit sphere $\mathbb{S}^{n-1}$, and the chordal distance between embedded points is $\sqrt{2}$ times the Hellinger distance $H(p,q)=\frac{1}{\sqrt{2}}\|\phi(p)-\phi(q)\|_2$, a numerically stable proxy for the Fisher--Rao geodesic that we develop in detail in \S\ref{ssec:hellinger}.

Measuring the Hellinger distance among all pairs of walk distributions yields a
single scalar that we call the \emph{aggregation dispersion} at depth~$k$:
\begin{equation}
\label{eq:dispersion_intro}
    \mathcal{D}_k
    \;=\;
    \frac{1}{n^2}\sum_{u,v} H^2\!\bigl(p_u^{(k)}, p_v^{(k)}\bigr)
    \;=\;
    1 - \Bigl\|\tfrac{1}{n}\textstyle\sum_v \sqrt{p_v^{(k)}}\Bigr\|_2^2.
\end{equation}

The left-hand side is the mean pairwise squared Hellinger distance among all walk distributions. The right-hand side reformulates this mean as the variance of the Hellinger-embedded point cloud on~$\mathbb{S}^{n-1}$, reducing the cost from $O(n^3)$ for the naïve pairwise computation to $O(n^2)$. 

\textbf{Contributions.}
We develop the theory and practice of aggregation dispersion in three parts.
\emph{First} (\S\ref{sec:theory_collapse}), we
prove that $\mathcal{D}_k = 0$ if and only if $\bP^k$ has rank one
(Theorem~\ref{thm:characterisation}), establishing that aggregation dispersion is an exact characterization of complete mixing of the propagation operator. We further show that
$\mathcal{D}_k$ is monotonically non-increasing in~$k$ via the data-processing
inequality (Theorem~\ref{thm:monotone}), confirming that repeated message
passing can only destroy distributional diversity, never create it.
\emph{Second} (\S\ref{sec:theory_decay}--\ref{sec:theory_mpnn}), we link
this monotonic decay to the spectrum of~$\bP$: the spectral gap controls the
rate at which $\mathcal{D}_k$ vanishes
(Theorem~\ref{thm:decay}). For contractive MPNNs, we further show that the
aggregation dispersion upper-bounds the dispersion of the learned
representations (Theorems~\ref{thm:linear_bound}
and~\ref{thm:nonlinear}), connecting the geometry of the simplex to the
geometry of the feature space.
\emph{Third} (\S\ref{sec:experiments}), we show empirically that $\mathcal{D}_k$ attains the highest mean absolute Pearson correlation with test accuracy across depths among eight diagnostics ($|r|=0.721$, using log-transformed diagnostic values) across 14 model configurations and 11 datasets, numerically comparable to the post-training effective rank of~\citet{deidda2025rethinking} ($|r|=0.688$) while requiring no training.

\section{Related Work}
\label{sec:related}

Prior work on the depth limitations of message-passing neural networks has largely focused on two complementary phenomena: oversmoothing and oversquashing, along with corresponding diagnostics and mitigation strategies. We position our work within the ``oversmoothing'' literature and introduce a novel perspective by leveraging information geometry to diagnose it.

\textbf{Oversmoothing.} \citet{li2018deeper} analyzed oversmoothing, observing that node representations in graph convolutional networks become indistinguishable as depth increases. \citet{oono2020graph} formalized this observation by proving the exponential convergence of node representations toward a low-information subspace. Subsequent work quantified oversmoothing through the \emph{Dirichlet energy}, the sum of squared feature variation across edges, showing that it decays to zero as layers are added~\citep{cai2020note}. Recently, \citet{rusch2023survey} provided a survey of these asymptotic results, while \citet{keriven2022not} showed that some smoothing can be beneficial for learning, with excessive smoothing degrading performance in the
studied setting. A key insight from these analyses is that graph topology influences oversmoothing: the propagation operator determines how fast walk distributions converge.

\textbf{Post-training diagnostics.} The diagnostic described above, Dirichlet energy, is computed from the feature matrices and can fail as an oversmoothing indicator for separately trained GNNs of moderate depth. \citet{deidda2025rethinking} proposed two alternatives, the \emph{effective rank} ``$\erank$'' and the \emph{numerical rank} ``$\nrank$'', both derived from the singular-value spectrum of the representation matrix. These rank-based measures track the effective dimensionality of the learned features and correlate strongly with accuracy degradation across depths in the evaluated architectures and datasets, even in regimes where energy-based metrics show no visible change. Their evaluation protocol pairs GCN~\citep{kipf2017semi} and GAT~\citep{velickovic2018graph} with LeakyReLU activations at eight depths across 7~datasets. We extend their protocol to 14~model configurations spanning 11~datasets and show that a training-free measure performs on par with these post-training metrics in average correlation.

\textbf{Training-free measures.}
Classical graph-structural quantities also characterize mixing without training: hitting and commute times, resistance distance, and the spectral gap~\citep{klein1993resistance, lovasz1993random}. Their classical forms are depth-independent and cannot directly track a depth--accuracy curve. Over our tested range, depth-truncated hitting and commute times are approximately linear in $k$, while the log spectral control $\log S_k=2k\log(1-\gamma)$ is exactly linear within each graph and operator (Appendix~\ref{app:structural-baselines}). These baselines therefore largely track depth.  $\mathcal{D}_k$ differs in that it is depth-resolved by construction and follows a nonlinear, spectrally controlled decay that carries information beyond depth (Appendix~\ref{app:depth-controlled}).

\textbf{Information geometry in machine learning.}
Our diagnostic rests on objects from information geometry: the Fisher--Rao metric, the Hellinger distance, and the data-processing inequality~\citep{amari2016information}. The Fisher--Rao metric underpins natural gradients~\citep{amari1998natural, martens2020new}, loss-landscape and data-manifold analyses~\citep{nielsen, ed-dib-etal-2025-gelora}, and geometrically principled generative models~\citep{arvanitidis2018latent}. To our knowledge, this is the first application of Hellinger geometry on the probability simplex to diagnosing oversmoothing.

\textbf{Oversmoothing mitigation.}
Complementary work proposes architectural and training strategies: initial residuals and identity mappings (GCNII,~\citealp{chen2020simple}), random edge removal (DropEdge,~\citealp{rong2020dropedge}), pairwise-distance normalization (PairNorm,~\citealp{zhao2020pairnorm}), and multi-layer aggregation (JKNet,~\citealp{xu2018representation}). We evaluate $\mathcal{D}_k$ on several of these in Appendix~\ref{app:mitigations}.


\section{Setup}
\label{sec:setup}

We now formalise the objects introduced informally in
\S\ref{sec:intro} (graphs, propagation operators, walk distributions, and the
Hellinger distance) and establish the notation used throughout the paper.
\subsection{Graphs and Propagation}
\label{ssec:graphs}
Let $G = (V, E)$ be a connected undirected graph with $n = |V|  \ge 2 $ nodes, $m = |E|$ edges, adjacency matrix $\bA \in \{0,1\}^{n \times n}$, and degree matrix $\bD = \diag(d_1, \ldots, d_n)$. To allow self-loops we set $\tA = \bA + \bI$, $\tD = \bD + \bI$, and write $\diam(G)$ for the diameter. Every standard MPNN architecture defines a \emph{propagation matrix} $\bP \in \mathbb{R}^{n \times n}$ that governs how information flows along edges: GCN uses $\tD^{-1/2}\tA\tD^{-1/2}$~\citep{kipf2017semi}, GraphSAGE aggregates neighbor features and applies separate transformations to root
and neighbor features~\citep{hamilton2017inductive}, and GAT uses learned attention coefficients~\citep{velickovic2018graph}. All share a \emph{locality property}: $[\bP^k]_{vu} \neq 0$ only if $d(v,u) \le k$, so after $k$ layers each node aggregates information from its $k$-hop neighborhood.

\subsection{The Canonical Walk Operator}
\label{ssec:walk}

To study oversmoothing as a structural property of message passing, we abstract away architectural details and analyze the \emph{lazy random walk}
\begin{equation}
\label{eq:lazy_walk}
    \bP \;=\; \tfrac{1}{2}\bigl(\bI + \bD^{-1}\bA\bigr),
\end{equation}
a row-stochastic, reversible, aperiodic operator with eigenvalues $1 = \lambda_1 > \lambda_2 \ge \cdots \ge \lambda_n \ge 0$~\citep[Ch.~1]{chung1997spectral}, spectral gap $\gamma = 1 - \lambda_2$, and stationary distribution $\pi_v = d_v/(2m)$~\citep{levin2017markov}. The lazy step guarantees a non-negative spectrum on every connected graph, with $\lambda_n = 0$ exactly when $G$ is bipartite. We adopt $\bP$ as a parameter-free canonical operator. Appendix~\ref{app:generalization} shows that the analysis holds for any row-stochastic operator satisfying a set of checkable conditions, which we verify for architectures such as GCN. \S\ref{sec:experiments} confirms that $\mathcal{D}_k(\bP)$ remains predictive even on operators outside these conditions, such as GAT's learned, non-reversible attention.

\subsection{Walk Distributions on the Probability Simplex}
\label{ssec:simplex}

Since $\bP$ is row-stochastic, each row of $\bP^k$ defines a probability distribution over nodes. For any node $v \in V$, we define the \emph{walk distribution}
\[
p_v^{(k)} = [\bP^k]_{v,:} \in \Delta^{n-1}, \qquad \Delta^{n-1} = \{p \in \mathbb{R}_{\ge 0}^n : \textstyle\sum_i p_i = 1\},
\]
the endpoint distribution of a $k$-step lazy random walk initialized at $v$. This induces a trajectory on the probability simplex, where each application of $\bP$ evolves the distribution from local concentration toward global equilibrium. At $k=0$ the walk is degenerate, $p_v^{(0)} = \delta_v$, concentrated entirely at the starting node; as $k$ grows, repeated application of $\bP$ redistributes mass over nodes reachable via multi-hop connectivity. Since the induced Markov chain is finite, connected, and aperiodic, the ergodic theorem guarantees convergence to the unique stationary distribution, $p_v^{(k)} \to \boldsymbol{\pi}$ for every $v \in V$. On a disconnected graph, the same convergence holds within each component, while distributions from different components have disjoint supports at every depth. Remark~\ref{rem:components} describes the effect on $\mathcal{D}_k$. Repeated propagation therefore contracts the cloud $\{p_v^{(k)}\}_{v \in V}$ toward a single point on the simplex when $G$ is connected, and one point per component otherwise, leading to oversmoothing. To quantify the rate of this contraction, we next equip the simplex with a distance.

\subsection{The Hellinger Distance}
\label{ssec:hellinger}

Different metrics on $\Delta^{n-1}$ reflect different notions of dissimilarity between distributions, and not all are well-suited to the oversmoothing setting. A principled selection criterion comes from \v{C}encov's theorem~\citep{cencov1982statistical}, which establishes that the \emph{Fisher--Rao metric} is the unique Riemannian metric on the simplex interior (up to a global constant) that is invariant under \emph{sufficient statistics}, that is, transformations that preserve all information relevant to statistical inference. This invariance makes the Fisher--Rao metric a canonical local geometry on the simplex. The simplex interior endowed with this metric $\frac{1}{4} g_{\mathrm{FR}}$ can be embedded isometrically into the unit sphere: the \emph{square-root embedding} $\phi\colon p \mapsto \sqrt{p}$ maps each distribution to a point on the non-negative orthant of the unit sphere $S_+^{n-1} = \{x \in \mathbb{R}^n_{\ge 0} : \|x\|_2 = 1\}$ (since $\|\sqrt{p}\|_2^2 = \sum_i p_i = 1$) equipped with the metric inherited from the ambient Euclidean space $\mathbb{R}^n$. 
\begin{wrapfigure}{r}{0.40\textwidth}
  \centering
  \vspace{-12pt}
  \includegraphics[width=0.38\textwidth]{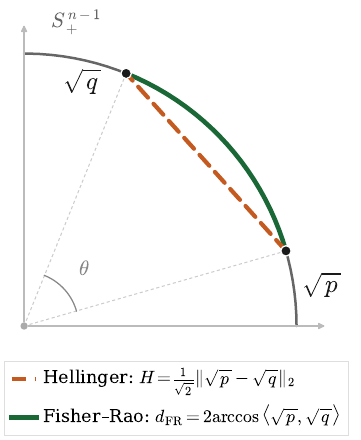}
    \caption{Under $\phi\colon p\mapsto\sqrt{p}$, the Hellinger
    distance is the dashed chord while the Fisher--Rao geodesic
    follows the arc on $S_+^{n-1}$.
    See Eq.~\eqref{eq:hellinger}.}
  \label{fig:hellinger-geodesic}
  \vspace{-20pt}
\end{wrapfigure}
Under this embedding, the Fisher--Rao geodesic distance is twice the spherical arc-length distance, $d_{\mathrm{FR}}(p,q) =2\arccos\!\bigl\langle  \sqrt{p},\sqrt{q}\bigr\rangle$. However, when $p$ and $q$ are close, evaluating $\arccos$ is sensitive to rounding errors near $1$. A natural alternative is a scaled chordal distance between the embedded points, which gives the \emph{Hellinger distance}
\begin{equation}
        \label{eq:hellinger}
            H(p,q) = \frac{1}{\sqrt{2}}\bigl\|\sqrt{p} - \sqrt{q}\bigr\|_2,
            \qquad H \in [0,1].
\end{equation}

The Hellinger distance equals $1/\sqrt{2}$ times the Euclidean distance between $\sqrt{p}$ and $\sqrt{q}$, so computations reduce to inner products on $S_+^{n-1}$. Three properties make it well-suited to diagnosing oversmoothing: it satisfies the data-processing inequality~\citep{cover2005inequalities}, yielding the monotonicity of $\mathcal{D}_k$ with depth (\S\ref{sec:theory_monotone}); it remains finite on disjoint supports, unlike the Kullback--Leibler or $\chi^2$ divergences, which is essential at early depths; and it induces the same topology as total variation~\citep{introtononparam}, linking $\mathcal{D}_k$ to mixing-time theory~\citep{levin2017markov}. Formal statements are in Appendix~\ref{app:hellinger}.

\section{The Aggregation Dispersion}
\label{sec:metric}

We combine the walk distributions and Hellinger distance from~\S\ref{sec:setup} into a single scalar that quantifies the diversity of the walk distributions at each depth.

\begin{definition}[Aggregation dispersion]
\label{def:dispersion}
Let $\bP$ be the lazy walk~\eqref{eq:lazy_walk} on a connected graph with
$n$~nodes, and let $p_v^{(k)} = [\bP^k]_{v,:}$ denote the walk distribution
from node~$v$ at depth~$k$. The \emph{aggregation dispersion} at depth~$k$ is:
\begin{equation}
\label{eq:dispersion_def}
    \mathcal{D}_k
    \;\coloneqq\;
    \frac{1}{n^2}\sum_{u,v \in V} H^2\!\bigl(p_u^{(k)}, p_v^{(k)}\bigr)
    \;=\;
    1 - \bigl\|\bar{\phi}^{(k)}\bigr\|_2^2,
\end{equation}
where $\bar{\phi}^{(k)} = \frac{1}{n}\sum_{v=1}^n \sqrt{p_v^{(k)}}$ is the
\emph{centroid} of the Hellinger-embedded walk distributions.
\end{definition}
The left-hand side of~\eqref{eq:dispersion_def} is the mean pairwise squared Hellinger distance among all $n^2$ pairs of walk distributions: it is large when different nodes ``see'' different parts of the graph, and zero when every node sees the same distribution. The right-hand side reformulates this mean in terms of the centroid $\bar{\phi}^{(k)}$, whose squared norm measures how tightly the embedded points cluster. The equivalence of these two forms is a consequence of the following standard identity.  

\begin{lemma}[Two forms equivalence]
\label{lem:equivalence}
The pairwise form and the centroid form in~\eqref{eq:dispersion_def} are
equal.
\end{lemma}

\textbf{Geometric interpretation.} The equivalence in Lemma~\ref{lem:equivalence} reveals that $\mathcal{D}_k$ is the \emph{extrinsic Euclidean variance} of the point cloud $\{\sqrt{p_v^{(k)}}\}_{v \in V}$ on the unit sphere~$S_+^{n-1}$, with the centroid norm $\|\bar{\phi}^{(k)}\|$ playing the role of a concentration parameter. This norm ranges between two extremes. At one extreme, when all walk distributions coincide, say $p_v^{(k)} = \pi$ for every~$v$, the centroid itself lies on the sphere ($\|\bar{\phi}^{(k)}\| = 1$), so $\mathcal{D}_k = 0$: the point cloud has collapsed to a single point. At the other extreme, at depth $k = 0$, each walk distribution is a Dirac mass $p_v^{(0)} = e_v$, so the embedded points $\sqrt{e_v} = e_v$ are the $n$ standard basis vectors, and the centroid is $\bar{\phi}^{(0)} = \mathbf{1}/n$ with $\|\bar{\phi}^{(0)}\|_2 = 1/\sqrt{n}$, yielding $\mathcal{D}_0 = (n{-}1)/n$, the maximum possible diversity. Between these two extremes, the curve $k \mapsto \mathcal{D}_k$ traces the trajectory from maximum diversity toward complete mixing, providing a depth-resolved diagnostic of propagation-induced oversmoothing.

\textbf{Computation.} For $k\ge1$, each walk distribution $p_v^{(k)} = e_v^\top \bP^k$ is obtained by $k$ sparse matrix--vector products at cost $O(m)$ per product, so all $n$ walk distributions cost $O(nmk)$; the centroid and its squared norm are $O(n^2)$, leaving the total cost at $O(nmk)$. For $n \gtrsim 20{,}000$, we sample $s\ge2$ source nodes independently and uniformly at random with replacement and use the bias-corrected estimator $\hat{\mathcal{D}}_k := \tfrac{s}{s-1}\bigl(1 - \|\hat{\bar\phi}^{(k)}\|_2^2\bigr)$, where $\hat{\bar{\phi}}^{(k)} = \tfrac{1}{s}\sum_{i=1}^s \sqrt{p_{v_i}^{(k)}}$. This estimator is unbiased and concentrates exponentially by McDiarmid's inequality~\citep{McDiarmid_1989}: for each depth $k$, $s = O(\epsilon^{-2}\log(1/\delta))$ samples suffice for $|\hat{\mathcal{D}}_k - \mathcal{D}_k| \le \epsilon$ with probability $\ge 1-\delta$. See Appendix~\ref{app:estimator} for the U-statistic form and proof.
\section{Theoretical Analysis}
\label{sec:theory}
With the aggregation dispersion $\mathcal{D}_k$ defined, we now establish its theoretical properties. The analysis proceeds in three stages. First, we show that $\mathcal{D}_k$ is a faithful and monotone measure of diversity loss (Theorems~\ref{thm:characterisation} and~\ref{thm:monotone}). Second, we connect the rate at which $\mathcal{D}_k$ decays to the spectral gap of the propagation matrix (Theorem~\ref{thm:decay}). Third, under explicit assumptions, we prove that the graph-level quantity $\mathcal{D}_k$ controls the representation-level diversity of the considered MPNNs (Theorems~\ref{thm:linear_bound} and~\ref{thm:nonlinear}), thereby justifying its use as a diagnostic for oversmoothing.
  
\subsection{Characterization of Collapse and Monotonicity}
\label{sec:theory_collapse}
A basic requirement for $\mathcal{D}_k$ is that it identifies complete mixing of the propagation operator: $\mathcal{D}_k$ should vanish exactly when all walk distributions coincide. The following theorem establishes this equivalence and states its consequences for the linear aggregation $\bP^k X$. Representation collapse can also occur before complete mixing, depending on the initial features, the learned weights, and the nonlinearities.

\begin{theorem}[Necessary and sufficient condition for collapse]
\label{thm:characterisation}
For a row-stochastic matrix $\bP^k$ with $n \ge 2$:
\begin{equation}
    \mathcal{D}_k = 0
    \quad\iff\quad
    p_u^{(k)} = p_v^{(k)} \;\;\forall\, u,v \in V
    \quad\iff\quad
    \rank(\bP^k) = 1.
\end{equation}
For every feature matrix $\bX$, $\mathcal{D}_k = 0$ implies $\bP^k\bX$ has identical rows (complete oversmoothing). For fixed $\bP$ and $k$, if $\bX\in\R^{n\times d}$, $d\ge1$, is drawn from a distribution absolutely continuous with respect to Lebesgue measure, then $\mathcal{D}_k > 0$ implies $\bP^k\bX$ is non-constant across nodes almost surely.
\end{theorem}

Theorem~\ref{thm:characterisation} establishes that $\mathcal{D}_k$ is a faithful indicator of collapse: it vanishes if and only if all walk distributions have become identical. This identifies $\mathcal{D}_k$ as a correct endpoint diagnostic for complete mixing. Having characterized \emph{when} collapse occurs, the next question is how diversity evolves with depth: can it increase under further stochastic propagation? The following result shows that aggregation dispersion is monotonically non-increasing with depth.

\label{sec:theory_monotone}

\begin{theorem}[Monotone decay]
\label{thm:monotone}
For a fixed row-stochastic propagation matrix $\bP$, $\mathcal{D}_{k+1} \le \mathcal{D}_k$ for all $k \ge 0$.
\end{theorem}

Theorem~\ref{thm:monotone} shows that diversity cannot increase under iterated aggregation. This reduces the study of propagation-induced oversmoothing to a rate question: how many layers can be applied before diversity becomes negligible?


\subsection{Spectral Control of the Decay Rate}
\label{sec:theory_decay}

We now show that the asymptotic decay rate of $\mathcal{D}_k$ is governed by a single spectral quantity: the gap $\gamma = 1 - \lambda_2$ of the lazy random walk. This rate provides a depth scale for walk mixing; relating it to representation diversity requires the assumptions stated below in \S~\ref{sec:theory_mpnn}.

\begin{theorem}[Decay bound]
\label{thm:decay}
Let $\pi$ be the stationary distribution of the lazy walk on a connected graph with $n\ge2$, $\pi_{\min}=\min_v\pi_v$,
and let $f_1\equiv 1, f_2,\dots,f_n$ be right eigenfunctions of $P$
orthonormal in $L^{2}(\pi)$, with eigenvalues
$1=\lambda_1>\lambda_2=1-\gamma\ge\cdots\ge\lambda_n\ge 0$.
Set $\sigma_2^{2}:=\tfrac{1}{n}\sum_v(f_2(v)-\bar f_2)^{2}$ with
$\bar f_2 = \tfrac{1}{n}\sum_v f_2(v)$.
Then for all $k\ge 0$, using the convention $(1-\gamma)^0=1$,
\begin{enumerate}[label=(\roman*),leftmargin=*,itemsep=2pt]
    \item \textbf{Upper bound:}
    $\displaystyle
    \mathcal{D}_k \;\le\; \pi_{\min}^{-1}(1-\gamma)^{2k}.$

    \item \textbf{Lower bound:}
    $\displaystyle
    \mathcal{D}_k \;\ge\;
    \tfrac{1}{4}\pi_{\min}\,\sigma_2^{2}\,(1-\gamma)^{2k},$
    where $\sigma_2^{2}>0$
    (replaced by $\sum_{i:\lambda_i=\lambda_2}\sigma_i^{2}$
    if $\lambda_2$ has multiplicity greater than one).
\end{enumerate}
\end{theorem}

Theorem~\ref{thm:decay} identifies the spectral gap as the quantity governing diversity decay: graphs with a large gap (strong expansion) have faster asymptotic decay of walk dispersion, while small-gap graphs have slower asymptotic decay. Yet $\mathcal{D}_k$ depends only on walk distributions, and hence only on topology. Whether this topological quantity predicts the behavior of learned GNN representations is the question we address next.
\begin{remark}[Disconnected graphs]
\label{rem:components}
Theorems~\ref{thm:characterisation} and~\ref{thm:monotone} do not require connectedness; Theorem~\ref{thm:decay} applies componentwise under its remaining assumptions. For component sizes $n_j=|C_j|$, cross-component squared Hellinger distances equal $1$, while within-component distances vanish, giving
\begin{equation}
\label{eq}
\mathcal{D}_k \longrightarrow \mathcal{D}_\infty
=1-\sum_{j=1}^{M}\Bigl(\frac{n_j}{n}\Bigr)^2.
\end{equation}
This is the probability that two independent uniform nodes lie in different components, positive for $M>1$. Notice that $\mathcal{D}_k$ carries an $n$-dependent scale and a topology-dependent floor and should not be compared across graphs; Appendix~\ref{app:practical-use} defines a normalized version for that purpose.
\end{remark}


\subsection{From Graph Dispersion to Representation Dispersion}
\label{sec:theory_mpnn}
The results so far establish that $\mathcal{D}_k$ is a faithful, monotone, and spectrally controlled measure of diversity among walk distributions. These are purely \emph{topological} statements: they depend only on the graph and the propagation operator. However, oversmoothing is ultimately a property of \emph{learned representations}. This raises the central question of this section:

\noindent
\emph{Does the graph-level quantity $\mathcal{D}_k$ control the diversity of
node representations produced by MPNNs?}

We answer this question in two steps. We first show that for linear MPNNs, representation diversity is directly controlled by $\mathcal{D}_k$. We then extend this to nonlinear architectures under some assumptions.

\textbf{Representation dispersion.}
To formalise this question, we introduce a measure of diversity among node
representations that parallels $\mathcal{D}_k$.

\begin{definition}[Representation dispersion]
\label{def:rep_dispersion}
For a feature matrix $\bX^{(k)} \in \R^{n \times d}$ whose $v$-th row
$x_v^{(k)}$ is the representation of node~$v$ at layer~$k$, define
\begin{equation}
    R_k \;=\; \frac{1}{n}\sum_{v=1}^n
    \bigl\|x_v^{(k)} - \bar{x}^{(k)}\bigr\|_2^2,
    \qquad \bar{x}^{(k)} = \frac{1}{n}\sum_v x_v^{(k)}.
\end{equation}
When $R_k = 0$, node representations are identical, while $R_k > 0$ means that some diversity remains.
\end{definition}

The challenge is that $\mathcal{D}_k$ and $R_k$ live in different geometries:
$\mathcal{D}_k$ is defined on the probability simplex via Hellinger distance,
while $R_k$ is a Euclidean variance in feature space. A Hellinger--$\ell_2$
equivalence (Lemma~\ref{lem:hellinger-l2}) bridges the two: half the mean pairwise squared $\ell_2$ distance between walk distributions is
sandwiched between $2\mathcal{D}_k^2/n$ and $4\mathcal{D}_k$. Hence
$\mathcal{D}_k$ controls the spread of walk distributions in the Euclidean
geometry relevant for linear transformations, and we can transfer this control
to representation space.

\textbf{Linear MPNNs.}
We begin with linear models, where the relationship is explicit.

\begin{theorem}[Linear MPNN bounds]
\label{thm:linear_bound}
For a linear \(k\)-layer MPNN with
\(x_v^{(k)} = p_v^{(k)}\!\bX^{(0)}\bM_k\),
let \(Z_k = \bX^{(0)}\bM_k\in\R^{n\times d}\)
and let \(\sigma_{\min,k}\ge0\) be the smallest singular value of \(Z_k^\top\) on the mean-zero subspace.
Then
\[
\frac{2\sigma_{\min,k}^2}{n}\,\mathcal{D}_k^2
\;\le\; R_k
\;\le\;
4\|Z_k\|_{\mathrm{op}}^2\,\mathcal{D}_k.
\]
In particular, $\mathcal{D}_k\to0$ implies $R_k\to0$ if
$\sup_k\|Z_k\|_{\mathrm{op}}<\infty$.
Conversely, $R_k\to0$ implies $\mathcal{D}_k\to0$ if
$\inf_k\sigma_{\min,k}>0$.
Thus the limits are equivalent when both uniform conditions hold.

\end{theorem}

The bounds relate representation dispersion to model-dependent factors ($\sigma_{\min,k}$, $\|Z_k\|_{\mathrm{op}}$) and the aggregation dispersion quantity $\mathcal{D}_k$,\footnote{A positive $\sigma_{\min,k}$ requires $Z_k^\top$ to be injective on $\mathbf{1}^{\perp}$, so $d\ge n-1$ is necessary. For $d<n-1$, the lower bound is vacuous.} and establish the equivalence between walk mixing ($\mathcal{D}_k\to0$) and representation collapse ($R_k\to0$) under the stated assumptions.

\textbf{Nonlinear MPNNs.}
We now extend this connection to nonlinear architectures.

\begin{theorem}[Nonlinear MPNN]
\label{thm:nonlinear}
Consider a $k$-layer MPNN defined by
\[
\bX^{(\ell+1)} = \sigma\!\bigl(\bP\,\bX^{(\ell)}\bW^{(\ell)}\bigr),
\qquad \ell=0,\ldots,k-1,
\]
where $\bP$ is the lazy walk on a connected graph,
$\sigma$ is applied row-wise and is $1$-Lipschitz, and
$\|\bW^{(\ell)}\|_{\mathrm{op}} \le 1$ for all layers.
Then $R_k \le C\,\mathcal{D}_k$, where $C\ge0$ is a graph-dependent constant and proportional to the initial dispersion.
\end{theorem}

Together, Theorems~\ref{thm:linear_bound} and~\ref{thm:nonlinear} establish that $\mathcal{D}_k$ bounds representation dispersion under the stated assumptions: conditional two-sided control in the linear case, and a one-sided upper bound in the nonlinear case. The reverse direction $R_k \ge c'\mathcal{D}_k^p$ fails in the nonlinear case without invertibility assumptions: the nonlinearity can collapse representations before the walk has mixed. In both cases, under the stated assumptions, $\mathcal{D}_k \to 0$ forces $R_k \to 0$, linking walk mixing to representation collapse. The next section tests the diagnostic empirically across architectures, datasets, and depths.

\section{Experiments}
\label{sec:experiments}

The theoretical analysis of \S\ref{sec:theory} establishes that $\mathcal{D}_k$ is a faithful, monotone, and spectrally controlled measure of walk-distribution diversity. We now test whether these properties translate into predictive power: does $\mathcal{D}_k$ correlate with test accuracy across depths more strongly than existing diagnostics, and does this hold across architectures, datasets, and graph types?

\subsection{Experimental Setup}
\label{ssec:exp_setup}

We test two questions: \emph{(i)} does $\mathcal{D}_k$, a purely topological diagnostic, correlate with test accuracy degradation of trained GNNs as depth increases? \emph{(ii)} how does it compare with post-training metrics that require access to learned representations? Following~\citet{deidda2025rethinking}, every depth is a separately trained model rather than a layer of a single deep network.

\textbf{Datasets.}
We evaluate on 11 node-classification benchmarks spanning homophily $h \in [0.05, 0.81]$ and scale $n \in [183, 169{,}343]$: the homophilic citation networks Cora, CiteSeer, PubMed~\citep{yang_data}; the heterophilic graphs Chameleon, Squirrel, Wisconsin, Texas, Cornell~\citep{Pei2020Geom-GCN:}; and three large-scale benchmarks Roman-Empire, Amazon-Ratings~\citep{platonov2023a} and ogbn-arxiv~\citep{ogbn-arxiv}. Per-dataset statistics are in Appendix~\ref{app:datasets-and-metrics}.

\textbf{Models.}
We evaluate seven architectures spanning message-passing, attentional, spectral, and decoupled designs: GCN~\citep{kipf2017semi}, GATv1~\citep{velickovic2018graph}, GATv2~\citep{brody2022how}, GraphSAGE~\citep{hamilton2017inductive}, ChebNet~\citep{chebcheb}, JKNet~\citep{xu2018representation}, and APPNP~\citep{appnp}, each with LeakyReLU and Tanh activations (14 configurations). For each configuration we sweep depths $L \in \{2, 4, 6, 8, 10, 12, 16, 24\}$ over 10 seeds, yielding $1{,}120$ trained models per dataset. Oversmoothing-mitigated GCN variants are deferred to Appendix~\ref{app:mitigations}, while GIN experiments are deferred to Appendix~\ref{app:gen-sum} and Appendix~\ref{app:gen-aug}.

\textbf{Training.}
Following~\citet{deidda2025rethinking}: hidden dim 32 (64 for the three large graphs), dropout 0.5, Adam ($\text{lr}{=}10^{-2}$, wd $5{\times}10^{-4}$), up to 200 epochs with early stopping (patience 10).

\textbf{Baselines and evaluation.}
We compare $\mathcal{D}_k$ against seven post-training metrics computed on final-layer representations: Dirichlet energy $E_{\mathrm{Dir}}$~\citep{cai2020note} and its normalized variant $E_{\mathrm{Dir,n}}$; projected energy $E_{\mathrm{Proj}}$ and $E_{\mathrm{Proj,n}}$~\citep{wu2023demystifying}; MAD~\citep{chen2020measuring}; and the effective and numerical ranks $\mathrm{Erank}$, $\mathrm{NumRank}$~\citep{deidda2025rethinking}. Definitions are in Appendix~\ref{app:datasets-and-metrics}. For each (dataset, model) pair we report the absolute Pearson correlation $|r|$ between $\log(\text{metric}_k)$ and test accuracy across eight depths,  and summarize variation across configurations and datasets below.
\subsection{Correlation with Test Accuracy Across Depths}
\label{ssec:main_results}

\begin{table}[H]
\centering
\footnotesize
\setlength{\tabcolsep}{4pt}
\renewcommand{\arraystretch}{1.15}
\caption{\textbf{Mean absolute Pearson correlation $|r|$ between log-transformed metrics and test accuracy across depths}, averaged over configurations. We use $\log(\mathrm{metric}-1)$ for Erank and NumRank, and $\log(\mathrm{metric})$ otherwise. $\mathcal{D}_k$ leads overall and on heterophilic graphs; NumRank leads on homophilic graphs. Uncertainties are SDs across configurations for dataset rows and SEs across dataset means for aggregate rows. \textbf{Bold}/\underline{underline}: highest/second-highest mean per row.}
\label{tab:ranking}
\resizebox{\textwidth}{!}{%
\begin{tabular}{l >{\columncolor{blue!12}}cccccccc}
\toprule
Dataset & $\mathcal{D}_k$ (ours) & Erank & $E_{\mathrm{Proj,n}}$ & NumRank & $E_{\mathrm{Dir,n}}$ & $E_{\mathrm{Dir}}$ & MAD & $E_{\mathrm{Proj}}$ \\
\midrule
\multicolumn{9}{@{}l}{\textsc{Homophilic}} \\
CiteSeer        & \textbf{0.842}\std{0.18}     & \underline{0.819}\std{0.13} & 0.691\std{0.23}             & 0.772\std{0.21}             & 0.696\std{0.26}             & 0.697\std{0.28}          & 0.619\std{0.21} & 0.641\std{0.27} \\
Cora            & \underline{0.744}\std{0.23}  & \textbf{0.756}\std{0.24}    & 0.704\std{0.17}             & 0.729\std{0.24}             & 0.636\std{0.21}             & 0.730\std{0.16}          & 0.538\std{0.21} & 0.641\std{0.22} \\
PubMed          & 0.681\std{0.23}              & 0.627\std{0.33}             & 0.626\std{0.26}             & \textbf{0.768}\std{0.22}    & \underline{0.708}\std{0.24} & 0.513\std{0.36}          & 0.703\std{0.23} & 0.575\std{0.25} \\
ogbn-arxiv      & 0.781\std{0.12}              & \textbf{0.812}\std{0.20}    & 0.717\std{0.29}             & \underline{0.794}\std{0.22} & 0.660\std{0.25}             & 0.544\std{0.31}          & 0.719\std{0.24} & 0.598\std{0.25} \\
\textit{Homophilic avg.} & \underline{0.762}\std{0.034} & 0.754\std{0.044} & 0.685\std{0.020} & \textbf{0.766}\std{0.014} & 0.675\std{0.017} & 0.621\std{0.054} & 0.645\std{0.042} & 0.614\std{0.016} \\
\midrule
\multicolumn{9}{@{}l}{\textsc{Heterophilic}} \\
Amazon-Ratings & \textbf{0.743}\std{0.20} & \underline{0.701}\std{0.25} & 0.677\std{0.31} & 0.652\std{0.32} & 0.587\std{0.29} & 0.667\std{0.34} & 0.594\std{0.34} & 0.540\std{0.30} \\
Chameleon       & \textbf{0.878}\std{0.13}     & 0.758\std{0.25}             & 0.708\std{0.24}             & 0.532\std{0.33}             & 0.624\std{0.29}             & 0.655\std{0.25}          & \underline{0.774}\std{0.14} & 0.661\std{0.25} \\
Cornell         & \underline{0.542}\std{0.23}  & 0.524\std{0.29}             & 0.519\std{0.33}             & \textbf{0.549}\std{0.27}    & 0.527\std{0.33}             & 0.494\std{0.38}          & 0.473\std{0.25}             & 0.510\std{0.35} \\
Roman-Empire & \underline{0.829}\std{0.18} & 0.751\std{0.19} & 0.713\std{0.20} & 0.680\std{0.22} & \textbf{0.893}\std{0.06} & 0.652\std{0.19} & 0.782\std{0.13} & 0.592\std{0.21} \\
Squirrel        & \textbf{0.723}\std{0.27}     & 0.609\std{0.27}             & \underline{0.710}\std{0.27} & 0.440\std{0.28}             & 0.622\std{0.29}             & 0.614\std{0.26}          & 0.520\std{0.36}             & 0.654\std{0.26} \\
Texas           & 0.521\std{0.19}              & 0.501\std{0.28}             & 0.592\std{0.27}             & 0.527\std{0.27}             & 0.514\std{0.31}             & \textbf{0.596}\std{0.30} & 0.488\std{0.28}             & \underline{0.594}\std{0.29} \\
Wisconsin       & 0.649\std{0.25}              & 0.706\std{0.29}             & \underline{0.754}\std{0.28} & 0.676\std{0.24}             & 0.650\std{0.32}             & \textbf{0.757}\std{0.29} & 0.577\std{0.29}             & \underline{0.754}\std{0.30} \\
\textit{Heterophilic avg.} & \textbf{0.698}\std{0.051} & 0.650\std{0.040} & \underline{0.668}\std{0.031} & 0.579\std{0.035} & 0.631\std{0.048} & 0.634\std{0.030} & 0.601\std{0.049} & 0.615\std{0.031} \\
\midrule
\textbf{Overall}
& \textbf{0.721}\std{0.035}
& \underline{0.688}\std{0.033}
& 0.674\std{0.021}
& 0.647\std{0.036}
& 0.647\std{0.031}
& 0.629\std{0.026}
& 0.617\std{0.034}
& 0.614\std{0.020} \\
\bottomrule
\end{tabular}}
\end{table}

Table~\ref{tab:ranking} reports the mean absolute Pearson correlation $|r|$ between $\log(\text{metric})$ and test accuracy across eight depths, averaged over 14 model configurations. The log transform linearizes exponential decay when present; Spearman and Kendall correlations retain $\mathcal{D}_k$ and Erank as the two leading diagnostics (Appendix~\ref{app:rank-correlations}).%
\footnote{We report $|r|$ because the sign of the metric--accuracy relationship can differ across diagnostics and model configurations; $|r|$ therefore measures strength of association only.}
Per-cell standard deviation is taken across configurations; subtotal and \textbf{Overall} standard errors are taken across dataset means within each reported group.

$\mathcal{D}_k$ achieves the highest observed mean absolute correlation ($0.721 \pm 0.035$) and ranks among the top two diagnostics on 7 of 11 benchmarks. Its lead over the runner-up, Erank, is not statistically significant (Appendix~\ref{app:significance}). These results suggest similar observed correlations to the strongest post-training diagnostics while avoiding the cost of obtaining trained representations: each post-training baseline requires representations from a separately trained model at every candidate depth, whereas $\mathcal{D}_k$ depends only on the propagation operator. For example, on ogbn-arxiv, evaluating $\mathcal{D}_k$ across all eight depths takes $128.8$\,s on CPU, compared with approximately $570$\,s for a single (model, seed, depth) training run (Appendix~\ref{app:estimator-runtime}).

The relative rankings further vary with homophily. On homophilic datasets, $\nrank$ leads ($0.766 \pm 0.014$) and $\mathcal{D}_k$ follows ($0.762 \pm 0.034$), a gap of $0.004$ in observed means. On heterophilic datasets, $\mathcal{D}_k$ leads ($0.698 \pm 0.051$), ahead of $E_{\mathrm{Proj,n}}$ ($0.668$) by $0.030$ in observed mean correlation. It ranks first on three of the seven heterophilic graphs, with the largest observed mean gaps on Chameleon ($0.878$ vs.\ $0.774$ for MAD) and Amazon-Ratings ($0.743$ vs.\ $0.701$ for Erank). This asymmetry may reflect heterophilic aggregation blurring class distinctions while leaving feature diversity unchanged, weakening the association between representation-based diagnostics and accuracy. $\mathcal{D}_k$ instead tracks walk mixing directly.

We also compared $\mathcal{D}_k$ with other training-free baselines: the spectral gap power $S_k=(1-\gamma)^{2k}$ from Theorem~\ref{thm:decay}, and depth-truncated hitting and commute times~\citep{lovasz1993random}. These baselines reach at most a mean $|r|=0.702$, but the spectral and mean truncated-walk baselines are closely tied to depth: $\log S_k$ is linear in $k$, while the mean depth-truncated hitting and commute times are evaluated without a log transform because they vary approximately linearly with $k$ over the tested range (Appendix~\ref{app:structural-baselines}). Compared with depth alone, $\mathcal{D}_k$ has a higher observed mean absolute correlation with test accuracy (one-sided paired Wilcoxon test over 11 dataset-level means, $p=0.027$). Adding $\log\mathcal{D}_k$ improves cross-validated $R^2$ by $0.099$ over a model with dataset intercepts and a shared depth slope, and by $0.008$ when slopes are dataset-specific (Appendix~\ref{app:added_value}). Dataset-specific slopes absorb the leading log-linear behavior expected under exponential mixing. At fixed graph and depth, varying the propagation operator yields a mean absolute correlation of $0.763$ between $\log\mathcal{D}_k(\mathbf{T})$ and accuracy (Appendix~\ref{app:teleport-sweep}).
\subsection{Per-Architecture Breakdown}
\label{ssec:per_arch}
 
$\mathcal{D}_k$ is defined from the lazy walk and references no architecture-specific propagation operator. We test whether this architecture-agnostic construction remains associated with test accuracy across depths across seven GNNs grouped into four families: spectral (GCN, ChebNet), attention (GATv1, GATv2), spatial (GraphSAGE), and skip/diffusion (JKNet, APPNP).

\renewcommand{\arraystretch}{0.95}

\begin{table}[H]
\centering
\footnotesize
\setlength{\tabcolsep}{4pt}
\caption{\textbf{Mean absolute correlation $|r|$ between $\log(\text{metric})$ and test accuracy}, averaged over datasets and both activations, per architecture. \textbf{Bold} = best per row, \underline{underline} = second best.}
\label{tab:per_arch}
\resizebox{\textwidth}{!}{%
\begin{tabular}{ll >{\columncolor{blue!12}}cccccccc}
\toprule
Family & Architecture & $\mathcal{D}_k$ (ours) & Erank & NumRank & $E_{\mathrm{Dir}}$ & $E_{\mathrm{Dir,n}}$ & $E_{\mathrm{Proj}}$ & $E_{\mathrm{Proj,n}}$ & MAD \\
\midrule
\multirow{2}{*}{Spectral}        & GCN     & \textbf{0.793}    & 0.705             & 0.671          & 0.553             & \underline{0.719} & 0.496 & 0.544             & 0.584 \\
                                 & ChebNet & \textbf{0.806}    & \underline{0.785} & 0.740          & 0.691             & 0.473             & 0.647 & 0.647             & 0.565 \\
\cmidrule(lr){1-10}
\multirow{2}{*}{Attention}       & GATv1   & \textbf{0.785}    & 0.688             & 0.583          & 0.633             & \underline{0.715} & 0.609 & 0.699             & 0.622 \\
                                 & GATv2   & \textbf{0.798}    & 0.764             & 0.669          & 0.659             & 0.777             & 0.681 & \underline{0.784} & 0.670 \\
\cmidrule(lr){1-10}
Spatial         & SAGE    & \underline{0.812} & \textbf{0.884}    & 0.704          & 0.739             & 0.752             & 0.734 & 0.810             & 0.761 \\
\cmidrule(lr){1-10}
\multirow{2}{*}{Skip/Diffusion}  & JKNet   & 0.503             & 0.489             & \textbf{0.566} & 0.474             & \underline{0.560} & 0.488 & 0.523             & 0.530 \\
                                 & APPNP   & 0.552             & 0.499             & 0.597          & \underline{0.653} & 0.531             & 0.647 & \textbf{0.708}    & 0.585 \\
\midrule
\multicolumn{2}{l}{\textbf{Overall}}
& \textbf{0.721}
& \underline{0.688}
& 0.647
& 0.629
& 0.647
& 0.614
& 0.674
& 0.617 \\
\bottomrule
\end{tabular}}
\end{table}

$\mathcal{D}_k$ has the highest observed mean correlation on all four spectral and attention architectures (GCN $0.793$, ChebNet $0.806$, GATv1 $0.785$, GATv2 $0.798$; family means $0.800$ and $0.792$). These results support using the canonical lazy walk as a surrogate for the evaluated architectures, including those whose propagation is not represented by a single fixed stochastic operator. 

For GraphSAGE, however, Erank ($0.884$) has a higher observed mean correlation than $\mathcal{D}_k$ ($0.812$). GraphSAGE can apply separate learned transformations to root and neighbor features, which the canonical lazy walk does not capture.

Correlations are weaker still for APPNP and JKNet, consistent with \citet{deidda2025rethinking}, who find that oversmoothing diagnostics become less informative when models include mechanisms that mitigate oversmoothing. APPNP restarts toward the input at each layer, producing node-specific personalized-PageRank limits and a positive dispersion floor that the lazy-walk model does not capture (Appendix~\ref{app:gen-appnp}). Computing $\mathcal{D}_k$ from APPNP's restart operator raises its observed mean absolute correlation from $0.552$ to $0.624$ (Appendix~\ref{app:matched-operator}). Meanwhile, JKNet combines representations from multiple layers in its readout, which can weaken the relationship between deepest-layer mixing and final accuracy; even the best diagnostic reaches only $0.566$. Both models also show smaller accuracy drops (roughly $4$~pp for JKNet and $15$~pp for APPNP, versus $23$--$33$~pp elsewhere), which can make correlations more sensitive to noise.


\section{Conclusion}
\label{sec:conclusion}

We introduced \textbf{aggregation dispersion} $\mathcal{D}_k$, a training-free diagnostic for oversmoothing in message-passing neural networks. Viewing each row of $\mathbf{P}^k$ as a probability distribution over nodes, $\mathcal{D}_k$ measures the contraction of the resulting point cloud on the simplex using the Hellinger distance, a scaled chordal counterpart of the Fisher--Rao distance. The diagnostic is computable in $O(nmk)$ time from the propagation matrix alone. We proved that $\mathcal{D}_k = 0$ if and only if $\mathbf{P}^k$ has rank one, that $\mathcal{D}_k$ decays monotonically with depth via the data-processing inequality, that the spectral gap controls the decay rate for the canonical lazy walk on a connected graph, and that $\mathcal{D}_k$ upper-bounds representation dispersion up to graph- and initialization-dependent constants for contractive MPNNs, and thus characterizes representational collapse (oversmoothing). Across 14 model configurations and 11 datasets, $\mathcal{D}_k$ attained the highest mean absolute Pearson correlation with test accuracy across depths ($|r| = 0.721$, using log-transformed diagnostic values), numerically comparable to the strongest post-training diagnostic, while requiring no trained representations. 

Additional analyses reported in the appendix clarify the appropriate use of $\mathcal{D}_k$. The depth-selection analysis shows that accuracy peaks at two layers for most (dataset, model) pairs, and that selecting depth using the evaluated diagnostics does not improve on a fixed two-layer choice (Appendix~\ref{app:depth-selection}). We therefore position $\mathcal{D}_k$ as a complementary structural diagnostic rather than a standalone depth selector: large $\mathcal{D}_k$ with collapsed representations reveals a mismatch between global walk diversity and learned representations, potentially due to the initial features, learned weights, or nonlinearities, whereas small $\mathcal{D}_k$ indicates substantial mixing of the walk distributions, which forces representation collapse as $\mathcal{D}_k\to0$ under the stated assumptions.

To evaluate its practical diagnostic value, we also test whether the observed correlations translate into reliable detection of accuracy degradation. In the pooled task over $1{,}231$ observations, the canonical lazy-walk score attains AUROC $0.602$, compared with $0.780$ for depth and $0.810$ for effective rank. On the $703$-observation subset, the architecture-associated score attains AUROC $0.752$ within fixed graph-and-depth cells, compared with $0.846$ for effective rank (Appendix~\ref{app:detection}). Beyond diagnosis, we explored $\mathcal{D}_k$ as a regularizer for GAT, but observed no statistically significant accuracy improvement in the tested settings (Appendix~\ref{app:regularizer}). A natural next step is to develop node-level diagnostics by analyzing the eigenvalues and eigenvectors of the pairwise Hellinger Gram matrix, to study which nodes lose distinctiveness first.

\begin{ack}
This work was supported by a fellowship from the Department of Finance and Risk Engineering at the NYU Tandon School of Engineering. We thank the anonymous reviewers and the area chair for their constructive feedback and suggestions, which helped improve our manuscript.
\end{ack}

\newpage
\bibliography{references}

\newpage
\appendix

\section{Notation and preliminaries}
\label{app:notation}

This section consolidates the notation used throughout the paper and collects the basic properties of the Hellinger distance invoked in our proofs. It is intended as a reference; readers familiar with the setup of \S\ref{sec:setup} may skip directly to Appendix~\ref{app:proofs}.

\subsection{Notation}
\label{app:notation-table}

We collect the symbols used throughout the paper in four short tables organised by category: graphs and operators (Table~\ref{tab:notation-graph}), the simplex and its embedding (Table~\ref{tab:notation-simplex}), distances (Table~\ref{tab:notation-distances}), and diagnostics (Table~\ref{tab:notation-diagnostics}). MPNN architectural notation ($\mathbf{X}^{(\ell)}$ for layer-$\ell$ features, $\mathbf{W}^{(\ell)}$ for weights, $\sigma$ for an elementwise activation, assumed $1$-Lipschitz in Theorem~\ref{thm:nonlinear}, and $\|\cdot\|_{\mathrm{op}}$ for the spectral norm) follows standard conventions and is not tabulated.

\begin{table}[H]
\centering
\caption{Graphs and propagation operators.}
\label{tab:notation-graph}
\renewcommand{\arraystretch}{1.15}
\begin{tabular}{@{}l>{\raggedright\arraybackslash}p{0.44\textwidth}@{}}
\toprule
$G = (V, E)$, $n = |V|$, $m = |E|$ & connected undirected graph with $n\ge2$ \\
$\mathbf{A}$, $\mathbf{D}$, $\tilde{\mathbf{A}} = \mathbf{A} + \mathbf{I}$, $\tilde{\mathbf{D}} = \mathbf{D} + \mathbf{I}$ & adjacency and degree matrices, and their self-loop-augmented versions \\
$d(u, v)$, $\operatorname{diam}(G)$ & shortest-path distance, diameter \\
$\mathbf{P} = \tfrac{1}{2}(\mathbf{I} + \mathbf{D}^{-1}\mathbf{A})$ & lazy random walk \\
$1 = \lambda_1 > \lambda_2 \geq \cdots \geq \lambda_n \ge 0$ & eigenvalues of $\mathbf{P}$ \\
$\gamma = 1 - \lambda_2$ & spectral gap \\
$\boldsymbol{\pi}$, $\pi_v = \frac{d_v}{2m}$, $\pi_{\min} = \min_v \pi_v$ & stationary distribution \\
$\rho = d_{\max}/d_{\min} = \pi_{\max}/\pi_{\min}$ & degree-heterogeneity ratio (Theorem~\ref{thm:nonlinear}) \\
$\rho_{\pi}$ & stationary-comparability constant, condition (C2) of Appendix~\ref{app:generalization} \\
\bottomrule
\end{tabular}
\end{table}

\begin{table}[H]
\centering
\caption{Walk distributions and the square-root embedding.}
\label{tab:notation-simplex}
\renewcommand{\arraystretch}{1.15}
\begin{tabular}{ll}
\toprule
$\Delta^{n-1} = \{p \in \mathbb{R}^n_{\geq 0} : \sum_i p_i = 1\}$ & probability simplex \\
$p_v^{(k)} = [\mathbf{P}^k]_{v,:}$ & walk distribution from $v$ at depth $k$ \\
$S_+^{n-1} = \{x \in \mathbb{R}^n_{\geq 0} : \|x\|_2 = 1\}$ & nonnegative orthant of unit sphere \\
$\phi: p \mapsto \sqrt{p}$, \quad $\phi_v^{(k)} = \sqrt{p_v^{(k)}}$ & square-root embedding \\
$\bar{\phi}^{(k)} = \tfrac{1}{n}\sum_v \phi_v^{(k)}$ & embedded centroid \\
\bottomrule
\end{tabular}
\end{table}

\begin{table}[H]
\centering
\caption{Distances on the probability simplex.}
\label{tab:notation-distances}
\renewcommand{\arraystretch}{1.15}
\begin{tabular}{ll}
\toprule
$H(p, q) = \tfrac{1}{\sqrt{2}}\|\sqrt{p} - \sqrt{q}\|_2$ & Hellinger, $H \in [0, 1]$ \\
$\mathrm{TV}(p, q) = \tfrac{1}{2}\|p - q\|_1$ & total variation \\
$d_{\mathrm{FR}}(p, q) = 2\arccos\langle\sqrt{p}, \sqrt{q}\rangle$ & Fisher--Rao geodesic \\
\bottomrule
\end{tabular}
\end{table}

\begin{table}[H]
\centering
\caption{Diagnostics.}
\label{tab:notation-diagnostics}
\renewcommand{\arraystretch}{1.15}
\begin{tabular}{@{}l>{\raggedright\arraybackslash}p{0.36\textwidth}@{}}
\toprule
$\mathcal{D}_k = \tfrac{1}{n^2}\sum_{u,v} H^2(p_u^{(k)}, p_v^{(k)}) = 1 - \|\bar{\phi}^{(k)}\|_2^2$ & aggregation dispersion (ours) \\
$\widehat{\mathcal{D}}_k$ & sampled estimator of $\mathcal{D}_k$ \\
$R_k = \tfrac{1}{n}\sum_v \|x_v^{(k)} - \bar{x}^{(k)}\|_2^2$ & representation dispersion (uniform) \\
$R_k^{\boldsymbol{\pi}} = \sum_v \pi_v \|x_v^{(k)} - \bar{x}^{\boldsymbol{\pi}}\|_2^2$ & representation dispersion ($\boldsymbol{\pi}$-weighted) \\
$\Delta_k = \tfrac{1}{n}\sum_v \|p_v^{(k)} - \bar{p}^{(k)}\|_2^2$ & $\ell_2$ dispersion of walk distributions (proofs only) \\
$\widetilde{\Delta}_k = \tfrac{1}{n}\sum_v \sum_j \bigl(p_v^{(k)}(j) - \bar{p}^{(k)}(j)\bigr)^2 / \pi_j$ & $\chi^2$-type ($\boldsymbol{\pi}$-weighted) dispersion (proofs only) \\
\bottomrule
\end{tabular}
\end{table}

Here $\bar{p}^{(k)} = \tfrac{1}{n}\sum_v p_v^{(k)}$ is the mean walk distribution.

\subsection{The lazy random walk}
\label{app:lazy-walk}

We adopt the lazy random walk $\mathbf{P} = \tfrac{1}{2}(\mathbf{I} + \mathbf{D}^{-1}\mathbf{A})$ as the canonical propagation operator throughout our analysis. Three properties make it the natural baseline for studying oversmoothing as a structural phenomenon, independent of architectural choices.

\textit{Row-stochasticity.} Each row of $\mathbf{P}$ sums to one, so each row of $\mathbf{P}^k$ is a probability distribution on $V$. This is what allows us to interpret the $v$-th row as a walk distribution $p_v^{(k)} \in \Delta^{n-1}$.

\textit{Reversibility.} The lazy walk satisfies the detailed-balance condition $\pi_u P_{uv} = \pi_v P_{vu}$ with respect to $\boldsymbol{\pi}$. Reversibility implies that $\mathbf{P}$ is self-adjoint on $L^2(\boldsymbol{\pi})$, so its eigenvalues are real and the symmetrised operator $\mathbf{S} = \mathbf{D}^{1/2} \mathbf{P} \mathbf{D}^{-1/2}$ has the same spectrum as $\mathbf{P}$ but with orthonormal eigenvectors. We exploit this in the proof of Theorem~\ref{thm:decay}(ii).

\textit{Aperiodicity.} The self-loop term $\tfrac{1}{2}\mathbf{I}$ ensures that $\mathbf{P}$ has period one, so the induced Markov chain converges to $\boldsymbol{\pi}$ from every initial distribution. Without this lazy correction, bipartite graphs would exhibit oscillatory behavior ($\mu_n = -1$ for $\mathbf{D}^{-1}\mathbf{A}$) and the walk distributions would not converge to $\boldsymbol{\pi}$; with it, $\lambda_n = \tfrac{1}{2}(1 + \mu_n) \ge 0$ for every connected graph.

The eigenvalue ordering $1 = \lambda_1 > \lambda_2 \geq \cdots \geq \lambda_n \ge 0$ is a consequence of the three properties combined: row-stochasticity gives $\lambda_1 = 1$ with multiplicity one (Perron--Frobenius, using connectedness), reversibility gives real eigenvalues, and the lazy correction maps the spectrum $\{\mu_i\} \subset [-1, 1]$ of $\mathbf{D}^{-1}\mathbf{A}$ to $\lambda_i = \tfrac{1}{2}(1 + \mu_i) \in [0, 1]$, so $\lambda_n \ge 0$ with equality exactly on bipartite graphs. None of the proofs in Appendix~\ref{app:proofs} requires $\lambda_n > 0$; they use only $\lambda_n \ge 0$, so non-bipartiteness is not assumed anywhere in the main results. For the operators $\alpha\mathbf{I}+(1-\alpha)\mathbf{D}^{-1}\mathbf{A}$ in Appendix~\ref{app:gen-ablation}, non-bipartiteness is needed for convergence only when $\alpha=0$. For $0<\alpha<1/2$, the chain is still aperiodic, but negative eigenvalues may affect the mixing rate. Appendix~\ref{app:generalization} extends our analysis to specified row-stochastic operators associated with GCN and GraphSAGE, and treats GIN, APPNP, and learned attention separately.

\subsection{The square-root embedding and the Hellinger distance}
\label{app:hellinger}

The square-root embedding $\phi : \Delta^{n-1} \to S_+^{n-1}$, $p \mapsto (\sqrt{p_1}, \ldots, \sqrt{p_n})$, sends each distribution to a point on the nonnegative orthant of the unit sphere, since $\|\sqrt{p}\|_2^2 = \sum_i p_i = 1$. The Hellinger distance is the normalized chord length between embedded points:
\begin{equation}
H(p, q) = \frac{1}{\sqrt{2}}\|\sqrt{p} - \sqrt{q}\|_2, \qquad H \in [0, 1].
\label{eq:hellinger-def}
\end{equation}
The factor $1/\sqrt{2}$ normalizes so that $H = 1$ when $p$ and $q$ have disjoint supports (then $\langle\sqrt{p}, \sqrt{q}\rangle = 0$ and $\|\sqrt{p} - \sqrt{q}\|_2 = \sqrt{2}$). Three properties of $H$ are used repeatedly in our proofs.

\begin{lemma}[Data-processing inequality]
\label{lem:dpi}
For any row-stochastic matrix $\mathbf{K} \in \mathbb{R}^{n \times n}$ and any two distributions $p, q \in \Delta^{n-1}$,
\begin{equation}
H(p\mathbf{K}, q\mathbf{K}) \leq H(p, q).
\end{equation}
\end{lemma}

The DPI is equivalent to the non-decrease of the Hellinger affinity $\langle\sqrt{p},\sqrt{q}\rangle$ under stochastic post-processing; see \citet{cover2005inequalities} for a textbook treatment. We apply it with $\mathbf{K} = \mathbf{P}$ in the proof of Theorem~\ref{thm:monotone} to establish that $\mathcal{D}_k$ is non-increasing in $k$.

\begin{lemma}[Finiteness on disjoint supports]
\label{lem:disjoint-support}
For any $p, q \in \Delta^{n-1}$, including those with $\mathrm{supp}(p) \cap \mathrm{supp}(q) = \emptyset$, the Hellinger distance is bounded: $H(p, q) \leq 1$.
\end{lemma}

This property distinguishes Hellinger from Kullback--Leibler and $\chi^2$ divergences, both of which diverge to infinity on disjoint supports. Disjoint supports can occur at small depths: for two nodes $u, v$ with $d(u, v) > 2k$, the walk distributions $p_u^{(k)}$ and $p_v^{(k)}$ have disjoint supports because $\mathbf{P}^k$ has the locality property $[\mathbf{P}^k]_{vu} \neq 0 \Rightarrow d(v, u) \leq k$. A divergence that blew up at $k = 0$ could not produce a finite-valued trajectory $k \mapsto \mathcal{D}_k$.

\begin{lemma}[Hellinger--total-variation equivalence]
\label{lem:hellinger-tv}
For any $p, q \in \Delta^{n-1}$,
\begin{equation}
H^2(p, q) \leq \mathrm{TV}(p, q) \leq \sqrt{2}\, H(p, q).
\end{equation}
\end{lemma}

The two inequalities imply that convergence in one metric is equivalent to convergence in the other, although the rates may differ. For example, the left inequality converts the standard spectral mixing bound $\mathrm{TV}(p_v^{(k)}, \boldsymbol{\pi}) \leq \tfrac{1}{2}\pi_v^{-1/2}(1-\gamma)^k$ from Markov-chain theory \citep{levin2017markov} into a Hellinger bound.

\subsection{The probability simplex as a Riemannian manifold}
\label{app:simplex-geometry}

The interior of the simplex carries a canonical Riemannian metric, the Fisher--Rao metric $g_{\mathrm{FR}}$. \v{C}encov's theorem~\citep{cencov1982statistical} characterizes the Fisher--Rao family, up to a common scale, by invariance under sufficient statistics across finite probability simplices. Under the square-root embedding $\phi$, the simplex interior equipped with $\tfrac{1}{4}g_{\mathrm{FR}}$ maps isometrically onto the strictly positive part of $S_+^{n-1}$ equipped with the round metric. Distances on the simplex therefore correspond to scaled spherical distances on $S_+^{n-1}$:

\begin{itemize}
\item The \emph{Fisher--Rao geodesic distance} $d_{\mathrm{FR}}(p, q) = 2\arccos\langle\sqrt{p}, \sqrt{q}\rangle$ is twice the arc length on $S_+^{n-1}$ between $\phi(p)$ and $\phi(q)$.
\item The \emph{Hellinger distance} $H(p, q) = \tfrac{1}{\sqrt{2}}\|\sqrt{p} - \sqrt{q}\|_2$ is the chord length between the same two points divided by $\sqrt{2}$.
\end{itemize}

These distance formulas extend continuously to the simplex boundary. For points on a unit sphere, chord length and arc length are related by $\|x - y\|_2 = 2\sin(\theta/2)$ where $\theta$ is the arc-length angle, so $H = \sqrt{2}\sin(d_{\mathrm{FR}}/4)$ and $H \sim d_{\mathrm{FR}}/(2\sqrt{2})$ as $d_{\mathrm{FR}}\to0$. We work with $H$ rather than $d_{\mathrm{FR}}$ throughout because it admits the closed-form centroid identity
\begin{equation}
\frac{1}{n^2}\sum_{u,v} H^2(p_u, p_v) = 1 - \|\bar{\phi}\|_2^2,
\label{eq:centroid-identity}
\end{equation}
which reduces the computation of $\mathcal{D}_k$ from $O(n^3)$ to $O(n^2)$ once the walk distributions are available and is the cornerstone of our estimator (Appendix~\ref{app:estimator}). The arccosine of an inner product does not yield this centroid identity.

\subsection{Two notions of representation dispersion}
\label{app:two-dispersions}

Our representation-level results state contraction bounds in two related but distinct quantities. The \emph{uniform-mean representation dispersion}
\begin{equation}
R_k = \frac{1}{n} \sum_{v=1}^n \|x_v^{(k)} - \bar{x}^{(k)}\|_2^2, \qquad \bar{x}^{(k)} = \frac{1}{n} \sum_v x_v^{(k)},
\end{equation}
is the natural Euclidean variance of the node representations under uniform node weighting. The \emph{$\boldsymbol{\pi}$-weighted representation dispersion}
\begin{equation}
R_k^{\boldsymbol{\pi}} = \sum_{v=1}^n \pi_v \|x_v^{(k)} - \bar{x}^{\boldsymbol{\pi}}\|_2^2, \qquad \bar{x}^{\boldsymbol{\pi}} = \sum_v \pi_v x_v^{(k)},
\end{equation}
is the variance under the stationary measure $\boldsymbol{\pi}$ and is the quantity that contracts cleanly under the lazy walk, because $\boldsymbol{\pi}^\top \mathbf{P} = \boldsymbol{\pi}^\top$ and the operator norm of $\mathbf{P}$ on the $\boldsymbol{\pi}$-mean-zero subspace of $L^2(\boldsymbol{\pi})$ is exactly $\lambda_2 = 1 - \gamma$.

The two dispersions are related by the following inequalities.
\begin{equation}
R_k \;\le\; \frac{d_{\text{avg}}}{d_{\min}}\; R_k^{\boldsymbol{\pi}},
\qquad
R_k^{\boldsymbol{\pi}} \;\le\; \frac{d_{\max}}{d_{\text{avg}}}\; R_k,
\label{eq:dispersion-relations}
\end{equation}
where
\(d_{\text{avg}} = \frac{1}{n}\sum_v d_v\),
\(d_{\min} = \min_v d_v\),
\(d_{\max} = \max_v d_v\).
On regular graphs both constants equal $1$ and the two notions coincide. In the proof of Theorem~\ref{thm:nonlinear}, we establish contraction in $R_k^{\boldsymbol{\pi}}$ and transfer to $R_k$ using the first bound.

\begin{proof}
For a connected undirected graph, the stationary distribution of the lazy walk is $\pi_v = d_v / (2|E|)$. Note that $d_{\text{avg}} = 2|E|/n$. Fix $k$ and suppress the depth superscripts on features and their means.

First we prove $R_k \le \frac{d_{\text{avg}}}{d_{\min}} R_k^{\boldsymbol{\pi}}$. Because the uniform mean $\bar{x}$ minimizes the uniform variance, we have
\[
R_k = \frac{1}{n}\sum_v \|x_v - \bar{x}\|^2 \;\le\; \frac{1}{n}\sum_v \|x_v - \bar{x}^{\boldsymbol{\pi}}\|^2.
\]
Rewriting the right-hand side using the weights $\pi_v$ gives
\[
\frac{1}{n}\sum_v \|x_v - \bar{x}^{\boldsymbol{\pi}}\|^2
= \sum_v \pi_v \cdot \frac{1}{n\pi_v}\|x_v - \bar{x}^{\boldsymbol{\pi}}\|^2
\;\le\; \left(\max_v \frac{1}{n\pi_v}\right) \sum_v \pi_v \|x_v - \bar{x}^{\boldsymbol{\pi}}\|^2.
\]
Since $\pi_v = d_v/(2|E|)$, we have $\frac{1}{n\pi_v} = \frac{2|E|}{n d_v} = \frac{d_{\text{avg}}}{d_v}$. Thus $\max_v \frac{1}{n\pi_v} = \frac{d_{\text{avg}}}{d_{\min}}$ and the first inequality follows.

For the reverse inequality, start from the definition of $R_k^{\boldsymbol{\pi}}$ and use that $\bar{x}^{\boldsymbol{\pi}}$ minimizes the weighted variance:
\[
R_k^{\boldsymbol{\pi}} = \sum_v \pi_v \|x_v - \bar{x}^{\boldsymbol{\pi}}\|^2
\;\le\; \sum_v \pi_v \|x_v - \bar{x}\|^2.
\]
Now write $\pi_v = \frac{d_v}{2|E|} = \frac{d_v}{n d_{\text{avg}}}$. Then
\[
\sum_v \pi_v \|x_v - \bar{x}\|^2
= \frac{1}{n d_{\text{avg}}} \sum_v d_v \|x_v - \bar{x}\|^2
\;\le\; \frac{d_{\max}}{d_{\text{avg}}} \cdot \frac{1}{n} \sum_v \|x_v - \bar{x}\|^2
= \frac{d_{\max}}{d_{\text{avg}}}\,R_k.
\]
This completes the proof.
\end{proof}

\section{Proofs of main results}
\label{app:proofs}

This appendix collects the proofs of the main-paper theorems and the equivalence lemma. The order follows the main text: \S\ref{app:proof-equivalence} establishes the two-form equivalence underlying Definition~\ref{def:dispersion}; \S\ref{app:proof-collapse} and \S\ref{app:proof-monotone} prove the structural properties (Theorems~\ref{thm:characterisation} and~\ref{thm:monotone}); \S\ref{app:proof-decay} proves the spectral decay bound (Theorem~\ref{thm:decay}, both parts); \S\ref{app:proof-hellinger-l2} establishes a Hellinger--$\ell_2$ equivalence used in the representation-level results; and \S\ref{app:proof-linear} and \S\ref{app:proof-nonlinear} prove the representation-dispersion bounds (Theorems~\ref{thm:linear_bound} and~\ref{thm:nonlinear}).

\subsection{Lemma~\ref{lem:equivalence}: two-form equivalence}
\label{app:proof-equivalence}

\begin{lemma}[Two-form equivalence]
\label{lem:equivalence_appendix}
The pairwise form and the centroid form in~\eqref{eq:dispersion_def} are
equal.
\end{lemma}

\begin{proof}
Since every walk distribution sums to one, each embedded point has unit norm:
$\|\sqrt{p_v^{(k)}}\|_2^2 = \sum_i p_{v,i}^{(k)} = 1$. The variance of
these unit-norm points about their centroid is therefore
\begin{align*}
    \frac{1}{n}\sum_v \bigl\|\sqrt{p_v^{(k)}} - \bar{\phi}^{(k)}\bigr\|_2^2
    &= \frac{1}{n}\sum_v \Bigl(
    \bigl\|\sqrt{p_v^{(k)}}\bigr\|^2 - 2\langle \sqrt{p_v^{(k)}},
    \bar{\phi}^{(k)}\rangle + \bigl\|\bar{\phi}^{(k)}\bigr\|^2\Bigr) \\
    &= 1 - 2\bigl\|\bar{\phi}^{(k)}\bigr\|^2 + \bigl\|\bar{\phi}^{(k)}\bigr\|^2
    = 1 - \bigl\|\bar{\phi}^{(k)}\bigr\|^2.
\end{align*}
The standard variance--covariance identity relates this variance to the mean
pairwise squared distance:
\[
    \frac{1}{n}\sum_v \bigl\|\sqrt{p_v^{(k)}} - \bar{\phi}^{(k)}\bigr\|^2
    = \frac{1}{2n^2}\sum_{u,v}\bigl\|\sqrt{p_u^{(k)}} - \sqrt{p_v^{(k)}}
    \bigr\|^2
    = \frac{1}{n^2}\sum_{u,v} H^2(p_u^{(k)}, p_v^{(k)}),
\]
where the last equality uses the definition of the Hellinger distance,
$H^2(p,q) = \tfrac{1}{2}\|\sqrt{p} - \sqrt{q}\|_2^2$. Combining the two
displays yields the claimed identity.
\end{proof}

\subsection{Theorem~\ref{thm:characterisation}: necessary and sufficient condition for collapse}
\label{app:proof-collapse}

\begin{theorem}[Necessary and sufficient condition for collapse]
\label{thm:characterisation_appendix}
For a row-stochastic matrix $\bP^k$ with $n \ge 2$:
\begin{equation}
    \mathcal{D}_k = 0
    \quad\iff\quad
    p_u^{(k)} = p_v^{(k)} \;\;\forall\, u,v \in V
    \quad\iff\quad
    \rank(\bP^k) = 1.
\end{equation}
For every feature matrix $\bX$, $\mathcal{D}_k = 0$ implies $\bP^k\bX$ has identical rows (complete oversmoothing). For fixed $\bP$ and $k$, if $\bX\in\R^{n\times d}$, $d\ge1$, is drawn from a distribution absolutely continuous with respect to Lebesgue measure, then $\mathcal{D}_k > 0$ implies $\bP^k\bX$ is non-constant across nodes almost surely.
\end{theorem}

\begin{proof}
\textit{(}$\Leftarrow$\textit{).} Suppose $p_v^{(k)} = q$ for all $v$. Then $\phi_v^{(k)} = \sqrt{q}$ for all $v$, the centroid is $\bar{\phi}^{(k)} = \sqrt{q}$, and $\|\bar{\phi}^{(k)}\|_2^2 = \|\sqrt{q}\|_2^2 = 1$, so $\mathcal{D}_k = 1 - 1 = 0$.

\textit{(}$\Rightarrow$\textit{).} Suppose $\mathcal{D}_k = 0$. By Lemma~\ref{lem:equivalence} in its variance form, $\frac{1}{n}\sum_v \|\phi_v^{(k)} - \bar{\phi}^{(k)}\|_2^2 = 0$. Each summand is non-negative, so every summand vanishes: $\phi_v^{(k)} = \bar{\phi}^{(k)}$ for all $v$. The coordinatewise square root $\mathbb{R}^n_{\geq 0} \to \mathbb{R}^n_{\geq 0}$ is injective, so $p_u^{(k)} = p_v^{(k)}$ for all $u, v \in V$.

A row-stochastic matrix with all rows equal to a common distribution $q$ has the form $\mathbf{1}q^\top$, which has rank one. Conversely, a row-stochastic rank-one matrix must have the form $\mathbf{1}q^\top$ for some $q \in \Delta^{n-1}$ (the unique row, repeated), so its rows are all equal.

\textit{Consequently clause.} When $\mathrm{rank}(\mathbf{P}^k) = 1$, the matrix has the form $\mathbf{1}q^\top$, so $\mathbf{P}^k\mathbf{X} = \mathbf{1}(q^\top\mathbf{X})$ is the same row repeated regardless of $\mathbf{X}$. When $\mathrm{rank}(\mathbf{P}^k) \geq 2$, there exist two distinct rows $a_u \neq a_v$ of $\mathbf{P}^k$. For fixed $\mathbf{P}$ and $k$, the set
$\{\mathbf{X}\in\mathbb{R}^{n\times d}:(a_u-a_v)\mathbf{X}=0\}$ is a proper linear subspace and has Lebesgue measure zero. Absolute continuity of the feature distribution therefore gives this set probability zero.
Hence $(\mathbf{P}^k\mathbf{X})_u \neq (\mathbf{P}^k\mathbf{X})_v$ almost surely.
\end{proof}

\subsection{Theorem~\ref{thm:monotone}: monotone decay}
\label{app:proof-monotone}

\begin{theorem}[Monotone decay]
\label{thm:monotone_appendix}
For a fixed row-stochastic propagation matrix $\bP$, $\mathcal{D}_{k+1} \le \mathcal{D}_k$ for all $k \ge 0$.
\end{theorem}

\begin{proof}
The Hellinger distance satisfies the data-processing inequality: for any row-stochastic matrix $\mathbf{K} \in \mathbb{R}^{n \times n}$ (a Markov kernel) and any two distributions $p, q \in \Delta^{n-1}$,
\begin{equation}
H(p\mathbf{K}, q\mathbf{K}) \leq H(p, q),
\label{eq:dpi-restated}
\end{equation}
see Lemma~\ref{lem:dpi} in Appendix~\ref{app:hellinger}. The walk distributions evolve under one additional application of the lazy walk via $p_v^{(k+1)} = p_v^{(k)}\mathbf{P}$. Applying \eqref{eq:dpi-restated} with $\mathbf{K} = \mathbf{P}$ to each pair $(u,v)$:
\begin{equation*}
H^2(p_u^{(k+1)}, p_v^{(k+1)}) = H^2(p_u^{(k)}\mathbf{P}, p_v^{(k)}\mathbf{P}) \leq H^2(p_u^{(k)}, p_v^{(k)}).
\end{equation*}
Summing over all $n^2$ pairs and dividing by $n^2$:
\begin{equation*}
\mathcal{D}_{k+1} = \frac{1}{n^2}\sum_{u,v} H^2(p_u^{(k+1)}, p_v^{(k+1)}) \leq \frac{1}{n^2}\sum_{u,v} H^2(p_u^{(k)}, p_v^{(k)}) = \mathcal{D}_k. \qedhere
\end{equation*}
\end{proof}

\subsection{Theorem~\ref{thm:decay}: spectral decay bound}
\label{app:proof-decay}

\begin{theorem}[Decay bound]
\label{thm:decay_appendix}
Let $\pi$ be the stationary distribution of the lazy walk on a connected graph with $n\ge2$, $\pi_{\min}=\min_v\pi_v$,
and let $f_1\equiv 1, f_2,\dots,f_n$ be right eigenfunctions of $P$
orthonormal in $L^{2}(\pi)$, with eigenvalues
$1=\lambda_1>\lambda_2=1-\gamma\ge\cdots\ge\lambda_n\ge 0$.
Set $\sigma_2^{2}:=\tfrac{1}{n}\sum_v(f_2(v)-\bar f_2)^{2}$ with
$\bar f_2 = \tfrac{1}{n}\sum_v f_2(v)$.
Then for all $k\ge 0$, using the convention $(1-\gamma)^0=1$,
\begin{enumerate}[label=(\roman*),leftmargin=*,itemsep=2pt]
    \item \textbf{Upper bound:}
    $\displaystyle
    \mathcal{D}_k \;\le\; \pi_{\min}^{-1}(1-\gamma)^{2k}.$

    \item \textbf{Lower bound:}
    $\displaystyle
    \mathcal{D}_k \;\ge\;
    \tfrac{1}{4}\pi_{\min}\,\sigma_2^{2}\,(1-\gamma)^{2k},$
    where $\sigma_2^{2}>0$
    (replaced by $\sum_{i:\lambda_i=\lambda_2}\sigma_i^{2}$
    if $\lambda_2$ has multiplicity greater than one).
\end{enumerate}
\end{theorem}

\begin{proof}
We establish the two bounds one after the other.

\textit{Part (i) -- Upper bound.}

Let \(p_v^{(k)} = P^k(v,\cdot)\) and define the density
\(g_k := p_v^{(k)}/\pi\).
For the reversible lazy walk, the \(L^2(\pi)\) contraction
(see \citet[Chapter~12]{levin2017markov}) gives
\[
\|g_k - 1\|_{2,\pi}
\le (1-\gamma)^k \|g_0 - 1\|_{2,\pi}.
\]
Squaring yields the familiar \(\chi^2\)-divergence bound
\[
\chi^2\!\left(p_v^{(k)} \,\|\, \pi\right)
   = \|g_k - 1\|_{2,\pi}^2
   \le (1-\gamma)^{2k}\|g_0 - 1\|_{2,\pi}^2.
\]
A direct computation shows
\[
\|g_0 - 1\|_{2,\pi}^2
= \frac{1-\pi(v)}{\pi(v)},
\]
hence
\[
\chi^2\!\left(p_v^{(k)} \,\|\, \pi\right)
   \le \frac{1-\pi(v)}{\pi(v)}(1-\gamma)^{2k}.
\]

Using the inequality \(H^2(p,\pi) \le \tfrac{1}{2}\chi^2(p \,\|\, \pi)\),
we obtain
\[
H^2\!\left(p_v^{(k)},\pi\right)
   \le \tfrac{1}{2}\,\frac{1-\pi(v)}{\pi(v)}(1-\gamma)^{2k}.
\]

By Lemma~\ref{lem:equivalence}, the centroid \(\bar{\phi}^{(k)}\) minimizes
the mean squared deviation, so
\[
\mathcal{D}_k
\le \frac{1}{n}\sum_v \|\phi_v^{(k)}-\sqrt{\pi}\|_2^2
= \frac{2}{n}\sum_v H^2\!\left(p_v^{(k)},\pi\right)
\le \frac{1}{n}\sum_v \frac{1-\pi(v)}{\pi(v)}(1-\gamma)^{2k}.
\]

In particular, using \((1-\pi(v))/\pi(v)\le\pi_{\min}^{-1}-1\) yields
\[
\mathcal{D}_k \le (\pi_{\min}^{-1}-1)(1-\gamma)^{2k},
\]
and the cruder bound
\(\mathcal{D}_k\le\pi_{\min}^{-1}(1-\gamma)^{2k}\)
follows immediately.

\medskip
\textit{Part (ii) -- Lower bound.}

The pointwise inequality
$(\sqrt{a}-\sqrt{b})^2\ge\tfrac14(a-b)^2$
for $a,b\in[0,1]$, summed coordinate-wise and averaged over vertex pairs, gives
\begin{equation}
\mathcal{D}_k \;\ge\; \tfrac{1}{4}\,\Delta_k,
\qquad
\Delta_k=\frac{1}{n}\sum_v
\bigl\|p_v^{(k)}-\bar p^{(k)}\bigr\|_2^2,
\label{eq:proof-dispersion-l2-lower}
\end{equation}
where $\bar p^{(k)}=\tfrac{1}{n}\sum_v p_v^{(k)}$ is the uniform vertex average.
It therefore suffices to lower-bound $\Delta_k$.

\smallskip
\emph{Spectral decomposition in the natural inner product.}
Let $f_1\equiv1,f_2,\dots,f_n$ denote the right eigenfunctions of the
lazy random walk operator $P$, orthonormal in $L^2(\pi)$, with eigenvalues
$1=\lambda_1>\lambda_2\ge\cdots\ge\lambda_n\ge0$
(so $\lambda_2=1-\gamma$).
For reversible chains \citep[Chapter~12]{levin2017markov} one has
\[
p_v^{(k)}(j)
=\pi(j)\sum_{i=1}^n\lambda_i^k f_i(v)f_i(j)
=\pi(j)+\pi(j)\sum_{i\ge2}\lambda_i^k f_i(v)f_i(j).
\]
Writing $\bar f_i:=\tfrac{1}{n}\sum_v f_i(v)$ and subtracting the
vertex-averaged version yields
\begin{equation}
p_v^{(k)}(j)-\bar p^{(k)}(j)
=\pi(j)\sum_{i\ge2}\lambda_i^k
\bigl(f_i(v)-\bar f_i\bigr)f_i(j).
\label{eq:proof-centered-spectral}
\end{equation}
Orthogonality holds in the $\pi$-weighted inner product:
$\sum_j\pi(j)f_i(j)f_h(j)=\delta_{ih}$.
We therefore introduce the $\chi^2$-type quantity
\[
\widetilde{\Delta}_k
:=\frac{1}{n}\sum_v\sum_j
\frac{\bigl(p_v^{(k)}(j)-\bar p^{(k)}(j)\bigr)^2}{\pi(j)},
\]
which carries exactly the right weighting to diagonalize \eqref{eq:proof-centered-spectral}.

\smallskip
\emph{Diagonalization.}
Substituting \eqref{eq:proof-centered-spectral}
and exchanging sums,
\begin{align*}
\widetilde{\Delta}_k
&=\frac{1}{n}\sum_v\sum_j\pi(j)
\Bigl(\sum_{i\ge2}\lambda_i^k(f_i(v)-\bar f_i)f_i(j)\Bigr)^2\\
&=\frac{1}{n}\sum_v\sum_{i,h\ge2}
\lambda_i^k\lambda_h^k
(f_i(v)-\bar f_i)(f_h(v)-\bar f_h)
\underbrace{\sum_j\pi(j)f_i(j)f_h(j)}_{=\,\delta_{ih}}\\
&=\sum_{i\ge2}\lambda_i^{2k}\sigma_i^2,
\qquad
\sigma_i^2:=\frac{1}{n}\sum_v\bigl(f_i(v)-\bar f_i\bigr)^2.
\end{align*}
The cross-terms vanish \emph{exactly}, and the coefficients
$\sigma_i^2\ge0$ are independent of $k$, depending on the graph and the chosen eigenbasis.

\smallskip
\emph{Non-degeneracy of the second mode.}
We claim $\sigma_2^2>0$. Indeed,
$\langle f_2,f_1\rangle_\pi=\sum_v\pi(v)f_2(v)=0$
together with $\|f_2\|_\pi=1$ forces $f_2$ to
be non-constant on $V$, hence its uniform empirical variance
$\sigma_2^2$ is strictly positive.
(If $\lambda_2$ has multiplicity $r>1$, the same argument applied to any
orthonormal basis of the second eigenspace gives
$\sum_{i:\lambda_i=\lambda_2}\sigma_i^2>0$,
which suffices in what follows.)

\smallskip
\emph{Lower bound at every depth}.
Retaining only the $i=2$ term and using $\lambda_2=1-\gamma$,
\[
\widetilde{\Delta}_k
\ge \sigma_2^2(1-\gamma)^{2k}
\qquad\text{for all }k\ge0.
\]

Since $\sum_jx_j^2\ge\pi_{\min}\sum_jx_j^2/\pi(j)$ for every vector $x$,
converting back to the unweighted Euclidean norm yields
\[
\Delta_k
\ge\pi_{\min}\widetilde{\Delta}_k
\ge\pi_{\min}\sigma_2^2(1-\gamma)^{2k}.
\]
Combining with \eqref{eq:proof-dispersion-l2-lower},
\[
\mathcal{D}_k
\ge\frac{\pi_{\min}\sigma_2^2}{4}(1-\gamma)^{2k}
\qquad\text{for every }k\ge0.
\]
Retaining all terms with $\lambda_i=\lambda_2$
gives the stated bound with the summed coefficient.
This completes the proof.
\end{proof}

\subsection{Hellinger--\texorpdfstring{$\ell_2$}{l2} equivalence}
\label{app:proof-hellinger-l2}

The proof of Theorem~\ref{thm:linear_bound} transfers the geometric content of $\mathcal{D}_k$ from
the Hellinger metric on the simplex to the $\ell_2$ metric on $\mathbb{R}^n$.
The following lemma provides the bridge.

\begin{lemma}[Hellinger--\(\ell_2\) equivalence]
\label{lem:hellinger-l2}
For any probability distributions \(p,q\in\Delta^{n-1}\),
\begin{equation}
\frac{4}{n}\,H^4(p,q)\leq\|p-q\|_2^2\leq8\,H^2(p,q).
\label{eq:hellinger-l2-basic}
\end{equation}
Consequently, let
\[
\mathcal{D}_k=\frac{1}{n^2}\sum_{u,v}H^2(p_u^{(k)},p_v^{(k)}),
\qquad
\Delta_k=\frac{1}{2n^2}\sum_{u,v}
\|p_u^{(k)}-p_v^{(k)}\|_2^2.
\]
Then
\begin{equation}
\frac{2\,\mathcal{D}_k^2}{n}\leq\Delta_k\leq4\,\mathcal{D}_k.
\label{eq:hellinger-l2-averaged}
\end{equation}
\end{lemma}

\begin{proof}
We proceed to show each bound separately, and then we conclude with the average bounds.

\textbf{Upper bound.}
For each coordinate \(i\) we factor
\[
p_i-q_i=(\sqrt{p_i}+\sqrt{q_i})(\sqrt{p_i}-\sqrt{q_i}).
\]
Squaring and summing over \(i\) gives
\[
\|p-q\|_2^2
=\sum_{i=1}^n
(\sqrt{p_i}+\sqrt{q_i})^2(\sqrt{p_i}-\sqrt{q_i})^2.
\]
Because \(p_i,q_i\in[0,1]\), we have
\(\sqrt{p_i},\sqrt{q_i}\in[0,1]\), hence
\((\sqrt{p_i}+\sqrt{q_i})^2\le4\).
Thus
\[
\|p-q\|_2^2
\le4\sum_{i=1}^n(\sqrt{p_i}-\sqrt{q_i})^2
=4\,\|\sqrt{p}-\sqrt{q}\|_2^2.
\]
By definition of the Hellinger distance,
\(\|\sqrt{p}-\sqrt{q}\|_2^2=2H^2(p,q)\).
Consequently,
\[
\|p-q\|_2^2\le8\,H^2(p,q),
\]
which is the right-hand inequality of~\eqref{eq:hellinger-l2-basic}.

\medskip
\noindent
\textbf{Lower bound.}
The Cauchy--Schwarz inequality yields
\[
\|p-q\|_1=\sum_{i=1}^n|p_i-q_i|
\le\sqrt{n}\,\|p-q\|_2.
\]
The Hellinger--total-variation inequality
(Lemma~\ref{lem:hellinger-tv}) states
\(\|p-q\|_1\ge2H^2(p,q)\).
Combining,
\[
\|p-q\|_2
\ge\frac{1}{\sqrt{n}}\|p-q\|_1
\ge\frac{2}{\sqrt{n}}\,H^2(p,q).
\]
Squaring gives
\[
\|p-q\|_2^2\ge\frac{4}{n}\,H^4(p,q),
\]
the left-hand inequality of~\eqref{eq:hellinger-l2-basic}.

\medskip
\noindent
\textbf{Averaged bounds for \(\Delta_k\).}
Using the upper bound for every pair \((u,v)\) and averaging,
\[
\Delta_k
=\frac{1}{2n^2}\sum_{u,v}\|p_u^{(k)}-p_v^{(k)}\|_2^2
\le\frac{1}{2n^2}\sum_{u,v}8\,H^2(p_u^{(k)},p_v^{(k)})
=4\;\underbrace{\frac{1}{n^2}\sum_{u,v}
H^2(p_u^{(k)},p_v^{(k)})}_{\mathcal{D}_k}
=4\,\mathcal{D}_k.
\]
For the lower bound we similarly insert the pointwise inequality:
\[
\Delta_k
\ge\frac{1}{2n^2}\sum_{u,v}\frac{4}{n}
H^4(p_u^{(k)},p_v^{(k)})
=\frac{2}{n}\;\frac{1}{n^2}\sum_{u,v}
H^4(p_u^{(k)},p_v^{(k)}).
\]
Jensen's inequality (applied to the convex function \(t\mapsto t^2\)) implies
\[
\frac{1}{n^2}\sum_{u,v}H^4(p_u^{(k)},p_v^{(k)})
\ge
\left(\frac{1}{n^2}\sum_{u,v}H^2(p_u^{(k)},p_v^{(k)})\right)^2
=\mathcal{D}_k^2.
\]
Hence
\[
\Delta_k\ge\frac{2}{n}\,\mathcal{D}_k^2,
\]
completing the proof of~\eqref{eq:hellinger-l2-averaged}.
\end{proof}

\subsection{Theorem~\ref{thm:linear_bound}: linear MPNN bound}
\label{app:proof-linear}

\begin{theorem}[Linear MPNN bounds]
\label{thm:thm:linear_bound_appendix}
Consider a linear \(k\)-layer MPNN with output
\(x_v^{(k)}=p_v^{(k)}\bX^{(0)}\bM_k\)
and let
\(Z_k=\bX^{(0)}\bM_k\in\mathbb{R}^{n\times d}\).
Denote by \(\sigma_{\min,k}\ge0\)
the smallest singular value of \(Z_k^\top\) on the
subspace of mean-zero vectors, i.e.
\[
\sigma_{\min,k}
=\inf\bigl\{\|Z_k^\top z\|_2:
\mathbf{1}^\top z=0,\;\|z\|_2=1\bigr\}.
\]
Then
\begin{equation}
\frac{2\sigma_{\min,k}^2}{n}\,\mathcal{D}_k^2
\le R_k
\le4\,\|{Z_k}\|_{\mathrm{op}}^2\,\mathcal{D}_k.
\end{equation}
In particular, $\mathcal{D}_k\to0$ implies $R_k\to0$
if $\sup_k\|Z_k\|_{\mathrm{op}}<\infty$.
Conversely, $R_k\to0$ implies $\mathcal{D}_k\to0$
if $\inf_k\sigma_{\min,k}>0$. Thus the limits are equivalent when both uniform conditions hold.
\end{theorem}

\begin{proof}
The linear output difference is
\(x_u^{(k)}-x_v^{(k)}
=(p_u^{(k)}-p_v^{(k)})Z_k\).
Averaging the squared \(\ell_2\) norm over all pairs gives
\[
R_k
=\frac{1}{2n^2}\sum_{u,v}
\|(p_u^{(k)}-p_v^{(k)})Z_k\|_2^2
\le\|Z_k\|_{\mathrm{op}}^2\,\Delta_k
\le4\|Z_k\|_{\mathrm{op}}^2\,\mathcal{D}_k,
\]
where the last inequality uses the upper bound
\(\Delta_k\le4\mathcal{D}_k\)
from Lemma~\ref{lem:hellinger-l2}.

For the lower bound, note that
\(p_u^{(k)}-p_v^{(k)}\) is mean-zero. Hence
\[
\|(p_u^{(k)}-p_v^{(k)})Z_k\|_2
\ge\sigma_{\min,k}\,
\|p_u^{(k)}-p_v^{(k)}\|_2.
\]
Squaring and averaging yields
\(R_k\ge{\sigma_{\min,k}^2}\,\Delta_k\).
Lemma~\ref{lem:hellinger-l2} gives
\(\Delta_k\ge\frac{2}{n}\mathcal{D}_k^2\),
and combining these proves the claimed lower bound.

If $\sup_k\|Z_k\|_{\mathrm{op}}<\infty$,
the upper bound gives
$\mathcal{D}_k\to0\Rightarrow R_k\to0$.
Conversely, if $R_k\to0$ and $\inf_k\sigma_{\min,k}>0$, then
\[
\mathcal{D}_k^2
\le\frac{n}{2\sigma_{\min,k}^2}R_k
\longrightarrow0,
\]
so \(\mathcal{D}_k\to0\).
\end{proof}

\subsection{Theorem~\ref{thm:nonlinear}: nonlinear MPNN contraction}
\label{app:proof-nonlinear}

\begin{theorem}[Nonlinear MPNN]
\label{thm:nonlinear_appendix}
Consider a $k$-layer MPNN defined by
\[
\bX^{(\ell+1)}
=\sigma\!\bigl(\bP\,\bX^{(\ell)}\bW^{(\ell)}\bigr),
\qquad \ell=0,\ldots,k-1,
\]
where $\bP$ is the lazy walk on a connected graph with $n\ge2$, $\sigma$ is applied row-wise and is $1$-Lipschitz, and $\|\bW^{(\ell)}\|_{\mathrm{op}}\le1$ for all layers.
Then
\[
R_k\le C\,\mathcal{D}_k,
\]
where $C\ge0$ is a graph-dependent constant
proportional to the initial dispersion.
\end{theorem}

\begin{proof}
Let \(R_\pi(\bY)\) denote the stationary-mean dispersion of a feature matrix
\(\bY\) with rows \(y_v\):
\[
R_\pi(\bY)
=\sum_{v=1}^n\pi_v\,\|y_v-\bar y_\pi\|_2^2,
\qquad
\bar y_\pi=\sum_{v=1}^n\pi_v y_v.
\]
Write $R_k^\pi=R_\pi(\bX^{(k)})$.

We show that for every layer \(\ell\),
\[
R_\pi(\bX^{(\ell+1)})
\le(1-\gamma)^2R_\pi(\bX^{(\ell)}).
\]

\textit{Propagation.}
The stationary distribution satisfies \(\pi^\top\bP=\pi^\top\).
Define the centered matrix
\(\bY_c=\bY-\mathbf{1}\bar y_\pi\);
its \(\pi\)-weighted rows sum to zero. Because the lazy walk is reversible,
\(\bP\) is self-adjoint on \(L_2(\pi)\), and its operator norm restricted to
the \(\pi\)-mean-zero subspace is exactly \(\lambda_2(\bP)=1-\gamma\).
Thus
\[
R_\pi(\bP\bY)
=\|\bP\bY_c\|_{L_2(\pi)}^2
\le\lambda_2(\bP)^2\|\bY_c\|_{L_2(\pi)}^2
=(1-\gamma)^2R_\pi(\bY),
\]
where
$\|\mathbf{B}\|_{L_2(\pi)}^2
:=\sum_v\pi_v\|\mathbf{B}_{v,:}\|_2^2$.

\textit{Weight multiplication.}
Since \(\|\bW^{(\ell)}\|_{\mathrm{op}}\le1\), we have
\[
R_\pi(\bY\bW^{(\ell)})
\le\|\bW^{(\ell)}\|_{\mathrm{op}}^2R_\pi(\bY)
\le R_\pi(\bY).
\]

\textit{Activation.}
Using the pairwise identity
\(R_\pi(\bY)=\frac12\sum_{u,v}\pi_u\pi_v\|y_u-y_v\|_2^2\)
and the \(1\)-Lipschitz property of \(\sigma\),
\[
R_\pi(\sigma(\bY))
=\frac12\sum_{u,v}\pi_u\pi_v
\|\sigma(y_u)-\sigma(y_v)\|_2^2
\le\frac12\sum_{u,v}\pi_u\pi_v\|y_u-y_v\|_2^2
=R_\pi(\bY).
\]

Chaining the three operations gives
\[
R_\pi(\bX^{(\ell+1)})
\le(1-\gamma)^2R_\pi(\bX^{(\ell)}),
\]
and iterating over \(k\) layers yields
\begin{equation}
R_k^\pi\le(1-\gamma)^{2k}R_0^\pi.
\label{eq:proof-weighted-contraction}
\end{equation}

Using the degree-based relations proved in
Appendix~\ref{app:two-dispersions}
(inequality~\eqref{eq:dispersion-relations}), we have
\[
R_k\le\frac{d_{\text{avg}}}{d_{\min}}R_k^\pi,
\qquad
R_0^\pi\le\frac{d_{\max}}{d_{\text{avg}}}R_0.
\]
Combining these with \eqref{eq:proof-weighted-contraction},
we obtain
\[
R_k
\le\frac{d_{\text{avg}}}{d_{\min}}(1-\gamma)^{2k}R_0^\pi
\le\frac{d_{\text{avg}}}{d_{\min}}(1-\gamma)^{2k}
\frac{d_{\max}}{d_{\text{avg}}}R_0
=\frac{d_{\max}}{d_{\min}}(1-\gamma)^{2k}R_0.
\]
Denoting $\rho=d_{\max}/d_{\min}=\pi_{\max}/\pi_{\min}$, we have
\begin{equation}
R_k\le\rho(1-\gamma)^{2k}R_0.
\label{eq:proof-uniform-contraction}
\end{equation}

Theorem~\ref{thm:decay_appendix} (ii) gives the positive constant
$c=\pi_{\min}\sigma_2^2/4$, depending only on the graph and the chosen normalized eigenfunction, such that
\[
\mathcal{D}_k\ge c(1-\gamma)^{2k}
\qquad\text{for all }k\ge0.
\]
Hence $(1-\gamma)^{2k}\le c^{-1}\mathcal{D}_k$.
Substituting this into \eqref{eq:proof-uniform-contraction} yields
\[
R_k
\le\frac{\rho R_0}{c}\,\mathcal{D}_k
=\frac{4\rho R_0}{\pi_{\min}\sigma_2^2}\,\mathcal{D}_k.
\]
This completes the proof.
\end{proof}

\section{Datasets and baseline diagnostics}
\label{app:datasets-and-metrics}

This appendix provides the experimental details for all results reported in \S\ref{sec:experiments} and Appendix~\ref{app:extended-results}: dataset statistics (\S\ref{app:dataset-stats}), train/validation/test splits (\S\ref{app:dataset-splits}), and formal definitions of the seven post-training baseline diagnostics we compare against (\S\ref{app:baseline-definitions}).

\subsection{Dataset statistics}
\label{app:dataset-stats}

We evaluate on 11 node-classification benchmarks spanning nearly three orders of magnitude in node count ($183 \le n \le 169{,}343$) and a broad range of edge homophily ($h \in [0.05, 0.81]$). Table~\ref{tab:dataset-stats} summarizes the key structural statistics of these datasets.

\begin{table}[H]
\centering
\caption{Dataset statistics. $n$ is the number of nodes; $m$ is the number of distinct undirected edges after symmetrization and self-loop removal; $C$ is the number of classes; $d_{\mathrm{feat}}$ is the input feature dimension; $\bar{d}$ is the mean degree; \(\operatorname{diam}(G)\) denotes the diameter of the largest connected component for disconnected graphs and the graph diameter otherwise; $h$ is the edge homophily, defined as the fraction of edges connecting nodes of the same class. Datasets are grouped by homophily regime and ordered within each group by node count.}
\label{tab:dataset-stats}
\renewcommand{\arraystretch}{1.10}
\small
\begin{tabular}{lrrrrrrrl}
\toprule
Dataset & $n$ & $m$ & $C$ & $d_{\mathrm{feat}}$ & $\bar{d}$ & $\operatorname{diam}(G)$ & $h$ & Domain \\
\midrule
\multicolumn{9}{l}{\textit{Homophilic} ($h \geq 0.5$)} \\
Cora           &   2{,}708 &    5{,}278 &  7 & 1{,}433 &  3.90 & 19 & 0.81 & Citation \\
CiteSeer       &   3{,}327 &    4{,}552 &  6 & 3{,}703 &  2.74 & 28 & 0.74 & Citation \\
PubMed         &  19{,}717 &   44{,}324 &  3 &     500 &  4.50 & 18 & 0.80 & Citation \\
ogbn-arxiv     & 169{,}343 & 1{,}157{,}799 & 40 &     128 & 13.67 & 23 & 0.65 & Citation \\
\midrule
\multicolumn{9}{l}{\textit{Heterophilic} ($h < 0.5$)} \\
Texas          &     183 &      279 &  5 & 1{,}703 &  3.05 &  8 & 0.11 & WebKB \\
Cornell        &     183 &      277 &  5 & 1{,}703 &  3.03 &  8 & 0.30 & WebKB \\
Wisconsin      &     251 &      450 &  5 & 1{,}703 &  3.59 &  8 & 0.21 & WebKB \\
Chameleon      &   2{,}277 &   31{,}371 &  5 & 2{,}325 & 27.55 & 11 & 0.23 & Wikipedia \\
Squirrel       &   5{,}201 &  198{,}353 &  5 & 2{,}089 & 76.28 & 10 & 0.22 & Wikipedia \\
Roman-Empire    &  22{,}662 &   32{,}927 & 18 &     300 &  2.91 & 6{,}824 & 0.05 & Wikipedia \\
Amazon-Ratings  &  24{,}492 &   93{,}050 &  5 &     300 &  7.60 & 46 & 0.38 & Co-purchase \\
\bottomrule
\end{tabular}
\end{table}

\paragraph{Sources.}
The citation networks Cora, CiteSeer, and PubMed follow \citet{yang_data}; the WebKB datasets (Texas, Cornell, Wisconsin) and the Wikipedia heterophilic graphs Chameleon and Squirrel follow \citet{Pei2020Geom-GCN:}; RomanEmpire and AmazonRatings follow \citet{platonov2023a}; ogbn-arxiv follows \citet{ogbn-arxiv}.

\paragraph{Note on Chameleon and Squirrel.}
We use the original \citet{Pei2020Geom-GCN:} releases of Chameleon and Squirrel ($n = 2{,}277$ and $n = 5{,}201$) rather than the filtered versions of \citet{platonov2023a} ($n = 890$ and $n = 2{,}223$). \citet{platonov2023a} document that the original graphs contain a substantial number of duplicate nodes that can leak between train and test partitions; their filtered versions remove this leakage. Averaging over ten random splits reduces variance in our reported accuracies but does not remove this leakage, since the duplicates are a property of the graph rather than of any particular split. Absolute accuracies on these two datasets are therefore not directly comparable to those obtained on the filtered versions. Leakage may also affect accuracy differently across depths and architectures, and therefore alter the within-dataset correlations evaluated here. We flag this as a limitation and direct readers to the filtered versions for any cross-paper comparison of absolute accuracy.

Excluding the Chameleon and Squirrel datasets, the mean absolute correlations over the remaining nine datasets are approximately $0.704$ for $\mathcal{D}_k$ and $0.689$ for Erank, an observed difference of approximately $0.015$. This exclusion reduces the observed gap and shows that the aggregate comparison is sensitive to benchmark composition. However, the aggregation dispersion still achieved higher observed mean correlation compared to the effective rank.

\subsection{Splits and training protocol}
\label{app:dataset-splits}

We use the standard Planetoid splits for Cora, CiteSeer, and PubMed \citep{yang_data}; the predefined split (split 0) of \citet{platonov2023a} for RomanEmpire and AmazonRatings; and the official time-based split for ogbn-arxiv. For Texas, Wisconsin, Cornell, Chameleon, and Squirrel, we generate ten random 60\%/20\%/20\% train/validation/test splits, one per seed, and average results over the ten seeds.

We report results averaged over 10 random seeds per (dataset, model, depth) configuration. The hidden dimension is 32 for most datasets and 64 for RomanEmpire, AmazonRatings, and ogbn-arxiv. We use dropout of 0.5 and the Adam optimizer with learning rate $10^{-2}$ and weight decay $5 \times 10^{-4}$. Models are trained for up to 200 epochs with early stopping (patience 10) based on validation performance, and we restore the best checkpoint.

\subsection{Baseline diagnostic definitions}
\label{app:baseline-definitions}

We compare $\mathcal{D}_k$ against seven post-training diagnostics computed on the
final-layer representation matrix $\mathbf{X}^{(L)} \in \mathbb{R}^{n \times d}$
produced by an $L$-layer GNN. To fix notation: $\mathbf{x}_v \in \mathbb{R}^d$ denotes
the $v$-th row of $\mathbf{X}^{(L)}$; $\mathbf{u} \in \mathbb{R}^n$ is a strictly positive dominant right
eigenvector of the propagation matrix $\mathbf{T}$,
normalized so that
$\|\mathbf{u}\|_2 = 1$ (for the symmetric
 GCN propagation matrix $\mathbf{T} =
\tilde{\mathbf{D}}^{-1/2}\tilde{\mathbf{A}}\tilde{\mathbf{D}}^{-1/2}$,
$\mathbf{u}$ is proportional to $(\sqrt{1+d_i})_{i=1}^{n}$; for the
row-stochastic aggregators used by GAT, $\mathbf{u} \propto \mathbf{1}$);
$\boldsymbol{\Pi}_{\mathbf{u}} = \mathbf{u}\mathbf{u}^\top$ is the orthogonal projection onto
$\mathrm{span}\{\mathbf{u}\}$; and $\sigma_1 \geq \sigma_2 \geq \cdots \geq
\sigma_{\min(n,d)} \geq 0$ are the singular values of $\mathbf{X}^{(L)}$.
We follow the diagnostic formulations of \citet{deidda2025rethinking}. The normalized energies and rank diagnostics below assume
$\mathbf{X}^{(L)}\neq0$; the MAD formula additionally assumes nonzero node representations on the evaluated edges.

\paragraph{Dirichlet energy ($\mathcal{E}_{\mathrm{Dir}}$, $\mathcal{E}_{\mathrm{Dir,n}}$).}
\citet{cai2020note} use the Dirichlet energy as a smoothness functional
aligned with the dominant eigenspace of the aggregation matrix:
\begin{equation}
\mathcal{E}_{\mathrm{Dir}}(\mathbf{X}^{(L)})
= \sum_{(i,j) \in E}
\left\| \frac{\mathbf{x}_i}{u_i} - \frac{\mathbf{x}_j}{u_j} \right\|_2^2.
\end{equation}
On a connected graph, the eigenvector rescaling makes
$\mathcal{E}_{\mathrm{Dir}}$ vanish exactly when
features align with $\mathbf{u}$; for GCN it is proportional to the standard degree-normalized form, and for GAT (where $\mathbf{u} \propto \mathbf{1}$) to the unweighted edge-difference sum. Following \citet{deidda2025rethinking}, we also
report the scale-invariant variant
\begin{equation}
\mathcal{E}_{\mathrm{Dir,n}}(\mathbf{X}^{(L)})
= \mathcal{E}_{\mathrm{Dir}}(\mathbf{X}^{(L)})
\big/ \|\mathbf{X}^{(L)}\|_F^2,
\end{equation}
which removes the dependency of $\mathcal{E}_{\mathrm{Dir}}$ on the global
magnitude of the representations.

\paragraph{Projected energy ($\mathcal{E}_{\mathrm{Proj}}$, $\mathcal{E}_{\mathrm{Proj,n}}$).}
Following \citet{deidda2025rethinking}, we measure the squared deviation of features from the dominant-eigenvector subspace,
\begin{equation}
\mathcal{E}_{\mathrm{Proj}}(\mathbf{X}^{(L)})
= \|\mathbf{X}^{(L)}
- \boldsymbol{\Pi}_{\mathbf{u}}\mathbf{X}^{(L)}\|_F^2
= \|(\mathbf{I}-\boldsymbol{\Pi}_{\mathbf{u}})
\mathbf{X}^{(L)}\|_F^2,
\qquad
\boldsymbol{\Pi}_{\mathbf{u}}=\mathbf{u}\mathbf{u}^\top,
\end{equation}
where $\boldsymbol{\Pi}_{\mathbf{u}}$ is the orthogonal projector onto $\mathrm{span}\{\mathbf{u}\}$. For $\mathbf{u}=\mathbf{1}/\sqrt{n}$, this is the square of the dispersion measure used by \citet{wu2023demystifying}. We also report the normalized counterpart
\begin{equation}
\mathcal{E}_{\mathrm{Proj,n}}(\mathbf{X}^{(L)})
= \mathcal{E}_{\mathrm{Proj}}(\mathbf{X}^{(L)})
\big/ \|\mathbf{X}^{(L)}\|_F.
\end{equation}
This follows the normalization convention of
\citet{deidda2025rethinking}.

On a connected graph with $u_i>0$,
$\mathcal{E}_{\mathrm{Proj}}$ and $\mathcal{E}_{\mathrm{Dir}}$ are equivalent up to constants depending on the graph and $\mathbf{u}$
\citep[Lemma A.1]{deidda2025rethinking}.
On disconnected graphs, zero Dirichlet energy allows
a different rescaled feature vector in each component, whereas the
displayed projected energy uses a single global vector.
Interpreting these metrics relative to a common propagation eigenspace
assumes a shared dominant right eigenvector across layers.

\paragraph{Mean Average Distance (MAD).}
\citet{chen2020measuring} define MAD
as the mean cosine distance over a target set of node pairs. We adopt the
edge-set instantiation used by \citet{deidda2025rethinking},
\begin{equation}
\mathrm{MAD}(\mathbf{X}^{(L)})
= \frac{1}{|E|}\sum_{(i,j)\in E}
\left(1-
\frac{\langle\mathbf{x}_i,\mathbf{x}_j\rangle}
{\|\mathbf{x}_i\|_2\|\mathbf{x}_j\|_2}\right),
\end{equation}
which measures angular dispersion of representations along the graph's edges and
is invariant to positive rescaling of individual rows.
Unlike $\mathcal{E}_{\mathrm{Dir}}$ and
$\mathcal{E}_{\mathrm{Proj}}$, MAD does not require a reference eigenvector
to be defined.

\paragraph{Effective rank (Erank).}
Following \citet{roy2007effective}, the
effective rank is the exponential of the Shannon entropy of the normalized
singular-value distribution. Writing
$\bar{\sigma}_i=\sigma_i/\sum_j\sigma_j$ and using the convention $0\log0=0$,
\begin{equation}
\mathrm{Erank}(\mathbf{X}^{(L)})
=\exp\!\left(-\sum_i\bar{\sigma}_i\log\bar{\sigma}_i\right).
\end{equation}
$\mathrm{Erank}$ ranges from $1$
(one direction concentrates all spectral mass)
to $\min(n,d)$ (uniform spectrum), and continuously interpolates between these
extremes. \citet{deidda2025rethinking} advocate it as an oversmoothing diagnostic
because, unlike $\mathcal{E}_{\mathrm{Dir}}$ and
$\mathcal{E}_{\mathrm{Proj}}$, it
does not require a known dominant eigenspace and is scale-invariant in
$\mathbf{X}^{(L)}$.

\paragraph{Numerical rank (NumRank).}
\citet{deidda2025rethinking} use the
continuous Frobenius-to-spectral ratio
\begin{equation}
\mathrm{NumRank}(\mathbf{X}^{(L)})
=\frac{\|\mathbf{X}^{(L)}\|_F^2}{\|\mathbf{X}^{(L)}\|_2^2}
=\frac{\sum_i\sigma_i^2}{\sigma_1^2}.
\end{equation}
$\mathrm{NumRank}$ is bounded between $1$ (rank-one features) and $\min(n,d)$,
is scale-invariant, and converges to one whenever $\sigma_2/\sigma_1\to0$.
The bound
$\mathrm{NumRank}(\mathbf{X})
\leq1+\|(\mathbf{I}-\boldsymbol{\Pi}_{\mathbf{u}})
\mathbf{X}\|_F^2/\|\mathbf{X}\|_2^2$
\citep[eq.~4]{deidda2025rethinking} bounds its excess above one by the projected energy
relative to the squared spectral norm.

All representation-based diagnostics are computed on the final hidden representation before the output classifier, in evaluation mode with dropout disabled. For each dataset, configuration, and depth, we first average test accuracy and each raw diagnostic over the available seeds. We then apply \(\log M\) to each averaged diagnostic, using \(\log(M-1)\) for Erank and NumRank, and compute the Pearson correlation across the available depths. We take the absolute value of this correlation before averaging across configurations and datasets.
\section{Estimator and computation}
\label{app:estimator}

This section addresses the practical question of computing $\mathcal{D}_k$ at scale. We first establish the unbiasedness and U-statistic form of the sampling estimator (\S\ref{app:estimator-unbiased}), then prove its exponential concentration via McDiarmid's inequality (\S\ref{app:estimator-concentration}), assess sampling error empirically (\S\ref{app:estimator-convergence}), and finally report runtime measurements across our benchmark suite (\S\ref{app:estimator-runtime}).

\subsection{Unbiased estimator and U-statistic form}
\label{app:estimator-unbiased}

For graphs with $n \gtrsim 20{,}000$ nodes, exact computation of $\mathcal{D}_k$ becomes expensive: each walk distribution $p_v^{(k)} = e_v^\top \mathbf{P}^k$ requires $k$ sparse matrix--vector products, each costing $O(m)$, so the full computation is $O(nmk)$. Sampling reduces the number of propagated source distributions from $n$ to $s$. We therefore sample $s$ source nodes $v_1,\ldots,v_s$ independently and uniformly with replacement from $V$ and form the empirical centroid
\begin{equation}
\widehat{\bar{\phi}}^{(k)}
= \frac{1}{s}\sum_{i=1}^s\phi_{v_i}^{(k)},
\qquad
\phi_{v_i}^{(k)}:=\sqrt{p_{v_i}^{(k)}}.
\end{equation}
We reuse the same sampled sources across all evaluated depths.
The bias-corrected estimator is
\begin{equation}
\widehat{\mathcal{D}}_k
:=\frac{s}{s-1}
\left(1-\|\widehat{\bar{\phi}}^{(k)}\|_2^2\right).
\end{equation}
The factor $s/(s-1)$ corrects the downward bias inherent in plugging an empirical mean into a centred quadratic, the same correction that converts the biased sample variance to its unbiased form.

\begin{proposition}[Unbiased estimator and U-statistic form]
\label{prop:unbiased}
The estimator
$\widehat{\mathcal{D}}_k
:=\frac{s}{s-1}\bigl(1-\|\widehat{\bar{\phi}}^{(k)}\|_2^2\bigr)$
satisfies
$\mathbb{E}[\widehat{\mathcal{D}}_k]=\mathcal{D}_k$
for $s\ge2$, and admits the representation
\[
\widehat{\mathcal{D}}_k
=\frac{1}{s(s-1)}\sum_{i\ne j}
H^2\!\bigl(p_{v_i}^{(k)},p_{v_j}^{(k)}\bigr).
\]
\end{proposition}

\begin{proof}
Expanding the squared norm of the empirical centroid:
\[
\begin{aligned}
\|\widehat{\bar{\phi}}^{(k)}\|_2^2
&=\Bigl\|\frac{1}{s}\sum_{i=1}^s\phi_{v_i}^{(k)}\Bigr\|_2^2\\
&=\frac{1}{s^2}\sum_{i=1}^s\sum_{j=1}^s
\langle\phi_{v_i}^{(k)},\phi_{v_j}^{(k)}\rangle\\
&=\frac{1}{s^2}\sum_{i=1}^s\|\phi_{v_i}^{(k)}\|_2^2
+\frac{1}{s^2}\sum_{i\ne j}
\langle\phi_{v_i}^{(k)},\phi_{v_j}^{(k)}\rangle.
\end{aligned}
\]
Since $\|\phi_{v_i}^{(k)}\|_2^2=1$ for every $i$, the diagonal sum equals $1/s$, giving
\begin{equation}
\label{eq:diag-offdiag}
\|\widehat{\bar{\phi}}^{(k)}\|_2^2
=\frac{1}{s}
+\frac{1}{s^2}\sum_{i\ne j}
\langle\phi_{v_i}^{(k)},\phi_{v_j}^{(k)}\rangle.
\end{equation}

For $i\ne j$, the samples $v_i$ and $v_j$ are independent and each uniform on $V$, so their joint distribution satisfies
$\Pr(v_i=u,v_j=w)=1/n^2$ for all $u,w\in V$. Taking expectations and using bilinearity,
\begin{equation}
\label{eq:expected-inner}
\begin{aligned}
\mathbb{E}\bigl[
\langle\phi_{v_i}^{(k)},\phi_{v_j}^{(k)}\rangle
\bigr]
&=\frac{1}{n^2}\sum_{u\in V}\sum_{w\in V}
\langle\phi_u^{(k)},\phi_w^{(k)}\rangle\\
&=\Bigl\|\frac{1}{n}\sum_{v\in V}\phi_v^{(k)}\Bigr\|_2^2
=\|\bar{\phi}^{(k)}\|_2^2.
\end{aligned}
\end{equation}

Taking expectations of~\eqref{eq:diag-offdiag} and applying~\eqref{eq:expected-inner} to each of the $s(s-1)$ off-diagonal pairs:
\[
\mathbb{E}\bigl[\|\widehat{\bar{\phi}}^{(k)}\|_2^2\bigr]
=\frac{1}{s}
+\frac{s(s-1)}{s^2}\|\bar{\phi}^{(k)}\|_2^2
=\frac{1}{s}
+\frac{s-1}{s}\|\bar{\phi}^{(k)}\|_2^2,
\]
so that
\[
\begin{aligned}
\mathbb{E}\bigl[1-\|\widehat{\bar{\phi}}^{(k)}\|_2^2\bigr]
&=1-\frac{1}{s}-\frac{s-1}{s}\|\bar{\phi}^{(k)}\|_2^2\\
&=\frac{s-1}{s}\bigl(1-\|\bar{\phi}^{(k)}\|_2^2\bigr)
=\frac{s-1}{s}\mathcal{D}_k.
\end{aligned}
\]
The plug-in $1-\|\widehat{\bar{\phi}}^{(k)}\|_2^2$ is thus downward biased by the factor $(s-1)/s$. Multiplying by $s/(s-1)$:
\[
\mathbb{E}[\widehat{\mathcal{D}}_k]
=\frac{s}{s-1}\frac{s-1}{s}\mathcal{D}_k
=\mathcal{D}_k.
\]

Substituting~\eqref{eq:diag-offdiag} into the estimator:
\begin{align*}
\widehat{\mathcal{D}}_k
&=\frac{s}{s-1}
\Bigl(
\frac{s-1}{s}
-\frac{1}{s^2}\sum_{i\ne j}
\langle\phi_{v_i}^{(k)},\phi_{v_j}^{(k)}\rangle
\Bigr)\\
&=1-\frac{1}{s(s-1)}\sum_{i\ne j}
\langle\phi_{v_i}^{(k)},\phi_{v_j}^{(k)}\rangle.
\end{align*}
Recalling that
$H^2(p,q)=1-\sum_u\sqrt{p(u)q(u)}
=1-\langle\sqrt{p},\sqrt{q}\rangle$,
we substitute
$\langle\phi_{v_i}^{(k)},\phi_{v_j}^{(k)}\rangle
=1-H^2(p_{v_i}^{(k)},p_{v_j}^{(k)})$:
\begin{align}
\widehat{\mathcal{D}}_k
&=1-\frac{1}{s(s-1)}\sum_{i\ne j}
\bigl[1-H^2(p_{v_i}^{(k)},p_{v_j}^{(k)})\bigr]
\notag\\
&=1-1+\frac{1}{s(s-1)}\sum_{i\ne j}
H^2(p_{v_i}^{(k)},p_{v_j}^{(k)})
\notag\\
&=\frac{1}{s(s-1)}\sum_{i\ne j}
H^2(p_{v_i}^{(k)},p_{v_j}^{(k)}),
\end{align}
which is a degree-2 U-statistic with kernel
$h(v_i,v_j)=H^2(p_{v_i}^{(k)},p_{v_j}^{(k)})$.
\end{proof}

The U-statistic form makes two facts transparent. First, $\widehat{\mathcal{D}}_k$ is a degree-2 U-statistic with bounded kernel
$h(v_i,v_j)=H^2(p_{v_i}^{(k)},p_{v_j}^{(k)})\in[0,1]$,
which places it in a setting where standard concentration inequalities apply.
Second, the centroid form
$\frac{s}{s-1}(1-\|\widehat{\bar{\phi}}^{(k)}\|_2^2)$
and the pairwise form
$\frac{1}{s(s-1)}\sum_{i\ne j}H^2$
are algebraically identical, but, once the sampled walk distributions are available, the centroid form is computed in $O(sn)$ time,
whereas direct evaluation of the pairwise form requires $O(s^2n)$ time. We always use the centroid form in practice.

\subsection{Concentration}
\label{app:estimator-concentration}

We now show that $\widehat{\mathcal{D}}_k$ concentrates exponentially around
$\mathcal{D}_k$ at each fixed depth, with the standard
$\epsilon^{-2}\log(1/\delta)$ sample complexity.

\begin{proposition}[Concentration via McDiarmid's inequality]
\label{prop:concentration}
For $s\ge2$ and any $\epsilon>0$,
\[
\Pr\bigl(
|\widehat{\mathcal{D}}_k-\mathcal{D}_k|\ge\epsilon
\bigr)
\le
2\exp\!\left(-\frac{\epsilon^2(s-1)^2}{8s}\right).
\]
Consequently, $s=O(\epsilon^{-2}\log(1/\delta))$ samples suffice for
$|\widehat{\mathcal{D}}_k-\mathcal{D}_k|\le\epsilon$
with probability at least $1-\delta$.
\end{proposition}

\begin{proof}
Replace the $\ell$-th sample $v_\ell$ with $v_\ell'$.
The empirical centroids
$\widehat{\bar{\phi}}^{(k)}$ and
$\widehat{\bar{\phi}}^{(k)\prime}$ differ by
\[
\widehat{\bar{\phi}}^{(k)\prime}
-\widehat{\bar{\phi}}^{(k)}
=\frac{1}{s}
\bigl(\phi_{v_\ell'}^{(k)}-\phi_{v_\ell}^{(k)}\bigr).
\]
By the triangle inequality and $\|\phi_v^{(k)}\|_2=1$ for all $v$:
\[
\begin{aligned}
\|\widehat{\bar{\phi}}^{(k)\prime}
-\widehat{\bar{\phi}}^{(k)}\|_2
&=\frac{1}{s}
\|\phi_{v_\ell'}^{(k)}-\phi_{v_\ell}^{(k)}\|_2\\
&\le\frac{1}{s}
\bigl(\|\phi_{v_\ell'}^{(k)}\|_2
+\|\phi_{v_\ell}^{(k)}\|_2\bigr)
=\frac{2}{s}.
\end{aligned}
\]

By definition,
$\widehat{\mathcal{D}}_k
=\frac{s}{s-1}(1-\|\widehat{\bar{\phi}}^{(k)}\|_2^2)$,
so
\[
|\widehat{\mathcal{D}}_k'
-\widehat{\mathcal{D}}_k|
=\frac{s}{s-1}
\bigl|
\|\widehat{\bar{\phi}}^{(k)}\|_2^2
-\|\widehat{\bar{\phi}}^{(k)\prime}\|_2^2
\bigr|.
\]
Applying the difference of squares identity
$\|a\|^2-\|b\|^2=\langle a+b,a-b\rangle$
followed by Cauchy--Schwarz:
\[
\begin{aligned}
\bigl|
\|\widehat{\bar{\phi}}^{(k)}\|_2^2
-\|\widehat{\bar{\phi}}^{(k)\prime}\|_2^2
\bigr|
&=
\bigl|
\langle
\widehat{\bar{\phi}}^{(k)}
+\widehat{\bar{\phi}}^{(k)\prime},
\widehat{\bar{\phi}}^{(k)}
-\widehat{\bar{\phi}}^{(k)\prime}
\rangle
\bigr|\\
&\le
\|\widehat{\bar{\phi}}^{(k)}
+\widehat{\bar{\phi}}^{(k)\prime}\|_2
\,
\|\widehat{\bar{\phi}}^{(k)}
-\widehat{\bar{\phi}}^{(k)\prime}\|_2.
\end{aligned}
\]
Since $\widehat{\bar{\phi}}^{(k)}$ is a convex combination of unit vectors,
$\|\widehat{\bar{\phi}}^{(k)}\|_2\le1$,
and likewise
$\|\widehat{\bar{\phi}}^{(k)\prime}\|_2\le1$,
so
$\|\widehat{\bar{\phi}}^{(k)}
+\widehat{\bar{\phi}}^{(k)\prime}\|_2\le2$.
Combining with the preceding bound:
\[
|\widehat{\mathcal{D}}_k'
-\widehat{\mathcal{D}}_k|
\le
\frac{s}{s-1}\cdot2\cdot\frac{2}{s}
=\frac{4}{s-1}
=:c.
\]
This holds uniformly for every coordinate $\ell\in\{1,\dots,s\}$.

The samples $v_1,\dots,v_s$ are independent and
$\mathbb{E}[\widehat{\mathcal{D}}_k]=\mathcal{D}_k$
by Proposition~\ref{prop:unbiased}.
The sum of squared bounded-difference constants is
\[
\sum_{\ell=1}^s c^2
=s\frac{16}{(s-1)^2}
=\frac{16s}{(s-1)^2}.
\]
McDiarmid's inequality states that for independent $X_1,\dots,X_s$ and any
function $f$ with
$\sup_{x_\ell,x_\ell'}
|f(\dots,x_\ell,\dots)-f(\dots,x_\ell',\dots)|\le c_\ell$:
\[
\Pr\bigl(|f-\mathbb{E}[f]|\ge\epsilon\bigr)
\le
2\exp\!\left(
-\frac{2\epsilon^2}{\sum_\ell c_\ell^2}
\right).
\]
Substituting:
\[
\begin{aligned}
\Pr\bigl(
|\widehat{\mathcal{D}}_k-\mathcal{D}_k|\ge\epsilon
\bigr)
&\le
2\exp\!\left(
-\frac{2\epsilon^2}{16s/(s-1)^2}
\right)\\
&=
2\exp\!\left(
-\frac{\epsilon^2(s-1)^2}{8s}
\right).
\end{aligned}
\]

Setting the bound to at most $\delta$:
\[
2\exp\!\left(-\frac{\epsilon^2(s-1)^2}{8s}\right)
\le\delta
\quad\Longleftrightarrow\quad
\frac{\epsilon^2(s-1)^2}{8s}
\ge\ln\frac{2}{\delta}.
\]
For $s\ge2$, we have $(s-1)^2/s\ge(s-1)/2$.
Therefore a sufficient condition is
\[
\frac{\epsilon^2(s-1)}{16}
\ge\ln\frac{2}{\delta}
\quad\Longleftrightarrow\quad
s\ge1+\frac{16}{\epsilon^2}\ln\frac{2}{\delta},
\]
confirming $s=O(\epsilon^{-2}\log(1/\delta))$.
\end{proof}

\paragraph{Beyond McDiarmid: a variance-aware refinement.}
Proposition~\ref{prop:concentration} is dimension-free, but it discards
the variance of the kernel
$h(u,w)=H^2(p_u^{(k)},p_w^{(k)})$
and retains only its worst-case range through the bounded-difference
constant $c=4/(s-1)$. For relative error
$\eta:=\epsilon/\mathcal{D}_k$ at a depth with $\mathcal{D}_k>0$,
the absolute-error guarantee yields the sufficient sample-size bound
$s=O\!\bigl(
\eta^{-2}\mathcal{D}_k^{-2}\log(1/\delta)
\bigr)$. A Bernstein-type bound improves this dependence on $\mathcal{D}_k$.
Hoeffding's variance decomposition for the degree-2 U-statistic
of Proposition~\ref{prop:unbiased} gives
\[
\begin{aligned}
\mathrm{Var}(\widehat{\mathcal{D}}_k)
&=
\frac{4(s-2)}{s(s-1)}\,\zeta_1
+\frac{2}{s(s-1)}\,\zeta_2,\\
\zeta_1&:=\mathrm{Var}\bigl(g(V_1)\bigr),
\qquad
\zeta_2:=\mathrm{Var}\bigl(h(V_1,V_2)\bigr),
\end{aligned}
\]
with $g(u):=\mathbb{E}_w[h(u,w)]$, where $V_1,V_2$ and $w$ are independent uniform vertices.
Expanding
$H^2(p_u^{(k)},p_w^{(k)})
=\tfrac12\|\phi_u^{(k)}-\phi_w^{(k)}\|_2^2$
around the centroid yields the closed form
\[
g(u)
=\tfrac12\|\phi_u^{(k)}-\bar{\phi}^{(k)}\|_2^2
+\tfrac12\mathcal{D}_k.
\]
Since $\phi_u^{(k)}$ and $\bar{\phi}^{(k)}$ have nonnegative entries
with $\|\phi_u^{(k)}\|_2=1$ and $\|\bar{\phi}^{(k)}\|_2\le1$, we have
$\|\phi_u^{(k)}-\bar{\phi}^{(k)}\|_2^2
\le1+\|\bar{\phi}^{(k)}\|_2^2\le2$,
and the self-bound
$\|\phi_u^{(k)}-\bar{\phi}^{(k)}\|_2^4
\le2\|\phi_u^{(k)}-\bar{\phi}^{(k)}\|_2^2$
gives, on taking expectations,
\[
\begin{aligned}
\zeta_1
&=
\tfrac14\mathbb{E}_u
\|\phi_u^{(k)}-\bar{\phi}^{(k)}\|_2^4
-\tfrac14\mathcal{D}_k^2
\le\tfrac12\mathcal{D}_k,\\
\zeta_2
&\le\mathbb{E}[h^2]
\le\mathbb{E}[h]
=\mathcal{D}_k,
\end{aligned}
\]
where $\mathbb{E}[h^2]\le\mathbb{E}[h]$ uses $h\in[0,1]$. Both variance terms are therefore bounded by multiples of $\mathcal{D}_k$, which McDiarmid's proof does not exploit.

To obtain a tail bound, let $b=\lfloor s/2\rfloor$
and let $\tau$ be an independent uniform permutation of
$\{1,\ldots,s\}$. Define
\[
A_\tau
:=\frac{1}{b}\sum_{j=1}^b
h(v_{\tau(2j-1)},v_{\tau(2j)}).
\]
Averaging over permutations gives
$\widehat{\mathcal{D}}_k
=\mathbb{E}_\tau[A_\tau\mid v_1,\ldots,v_s]$.
For each fixed permutation, the $b$ summands are independent,
lie in $[0,1]$, and have variance $\zeta_2\le\mathcal{D}_k$.
Jensen's inequality applied to the exponential therefore bounds
the moment-generating function of the centred U-statistic by that
of the corresponding paired average. Applying the usual Bernstein
argument to this average gives
\[
\Pr\bigl(
|\widehat{\mathcal{D}}_k-\mathcal{D}_k|\ge\epsilon
\bigr)
\le
2\exp\!\left(
-\frac{\lfloor s/2\rfloor\,\epsilon^2}
{2(\zeta_2+\epsilon/3)}
\right)
\le
2\exp\!\left(
-\frac{\lfloor s/2\rfloor\,\epsilon^2}
{2(\mathcal{D}_k+\epsilon/3)}
\right).
\]
For $\mathcal{D}_k>0$ and $0<\eta\le1$,
setting $\epsilon=\eta\mathcal{D}_k$ yields the sufficient requirement
\[
s=O\!\left(
\frac{1}{\eta^2\mathcal{D}_k}\log(1/\delta)
\right).
\]
This improves the dependence on $\mathcal{D}_k$
from $\mathcal{D}_k^{-2}$ to $\mathcal{D}_k^{-1}$ in these sufficient bounds. On connected graphs with $0<\lambda_2<1$, Theorem~\ref{thm:decay} implies that this halves the resulting exponential dependence on depth. Neither bound is asserted to be tight.

\subsection{Empirical convergence of the sampling estimator}
\label{app:estimator-convergence}

\begin{figure}[H]
\centering
\includegraphics[width=\textwidth]{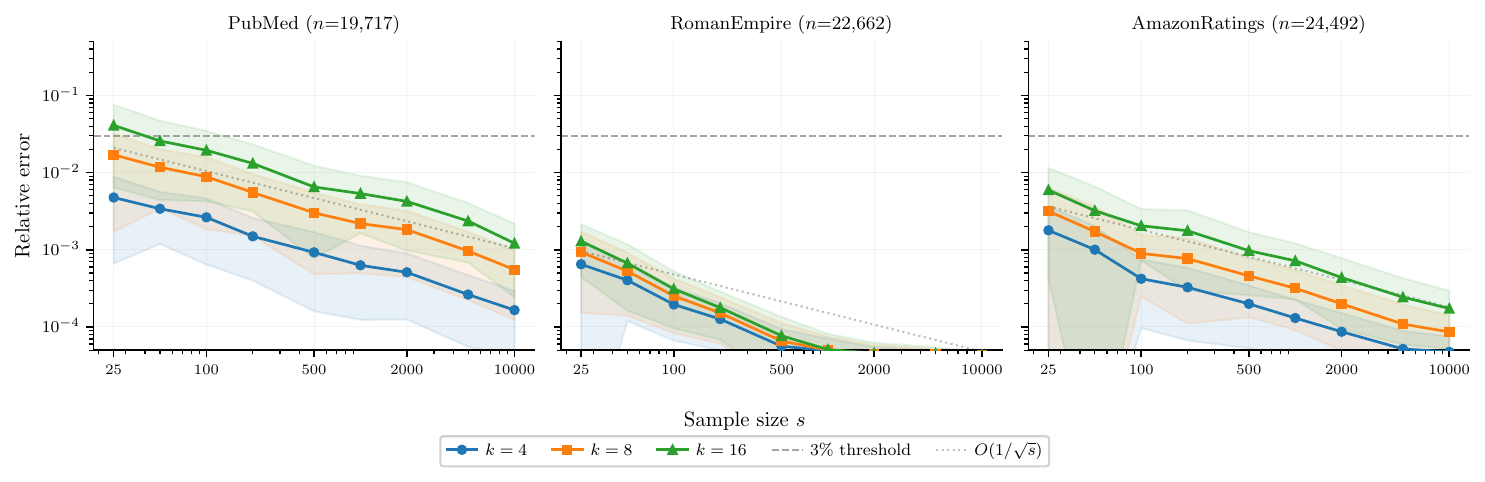}
\caption{Sampling estimator convergence. Mean relative error vs.\ sample size $s$ for three datasets at depths $k\in\{4,8,16\}$. Error bars: $\pm1$ s.d.\ over 50 trials. The default $s=200$ yields mean relative error below $3\%$ (dashed grey line) at all tested depths on these datasets.}
\label{fig:convergence}
\end{figure}

Proposition~\ref{prop:concentration} gives an absolute-error bound of order $s^{-1/2}$ at fixed confidence.
Figure~\ref{fig:convergence} examines the dependence on sample size on three datasets: PubMed, RomanEmpire, and AmazonRatings.

For each dataset and trial, we draw one sample of $s$
source nodes independently and uniformly with replacement and reuse it across depths. We compute $\widehat{\mathcal{D}}_k$ from the centroid form
and report the relative error
$|\widehat{\mathcal{D}}_k-\mathcal{D}_k|/\mathcal{D}_k$
averaged over 50 trials, with the exact $\mathcal{D}_k$ as ground truth. We do this at three representative depths $k\in\{4,8,16\}$, covering the earlier and later parts of the tested depth range.

Three empirical observations align with the theory.
First, the mean relative error decreases with $s$, broadly consistent with the $s^{-1/2}$ reference trend (Figure~\ref{fig:convergence}). Second, the mean relative error is below $3\%$ for all tested $s\ge50$: on RomanEmpire ($n=22{,}662$) and AmazonRatings ($n=24{,}492$), even $s=25$ suffices for $<3\%$ mean error at each tested depth; on PubMed ($n=19{,}717$), $s=50$ brings the mean error below $3\%$ at the most challenging depth ($k=16$, relative error $2.6\%$). Third, deeper walks tend to have larger mean relative
errors in these experiments. Both the absolute error, through the kernel variance, and the normalizing factor $\mathcal{D}_k$ depend on depth.

We adopt $s=200$ as the default for $n\ge20{,}000$ in all main-paper experiments. For ogbn-arxiv ($n=169{,}343$), this represents a sampling rate of $0.12\%$. Reusing intermediate walk vectors requires $24$ sparse propagation steps per sampled source, with centroid reductions at the eight evaluated depths.

\textbf{A note on the $sn$ memory cost.}
Storing all $s$ walk distributions of length $n$
simultaneously uses $O(sn)$ auxiliary storage, excluding the graph and temporary arrays.
On \texttt{ogbn-arxiv}, with $s=200$, this corresponds to approximately $135$\,MB in \texttt{float32}, which is well within the limits of
commodity hardware.

For larger graphs, the walk distributions can be streamed:
each $\phi_{v_i}^{(k)}$ contributes additively to
$\widehat{\bar{\phi}}^{(k)}$ and can be discarded once accumulated, reducing the auxiliary storage to $O(n)$ for a fixed number of evaluated depths without increasing the asymptotic time complexity. We did not require this optimization for any benchmark in our suite.

\subsection{Runtime}
\label{app:estimator-runtime}

All experiments were run on an Apple M2 Max with 64\,GB RAM.
Walk distributions are computed via repeated sparse matrix--vector products
using \texttt{scipy.sparse}; the squared centroid norm is computed
in a single dense reduction.

Table~\ref{tab:runtime} reports wall-clock time for computing $\mathcal{D}_k$ across all eight depths $k\in\{2,4,6,8,10,12,16,24\}$  on each benchmark. We compute exactly when $n<20{,}000$ and switch to the sampling estimator with $s=200$ source nodes otherwise.

\begin{table}[H]
\centering
\caption{Runtime for computing $\mathcal{D}_k$ across all eight depths on the Apple M2 Max CPU. Sampling is used for $n\ge20{,}000$. $m$ counts distinct undirected edges after symmetrization and self-loop removal. Total time across the full 11-dataset suite is under 30 minutes.}
\label{tab:runtime}
\begin{tabular}{lrrlr}
\toprule
Dataset & $n$ & $m$ & Mode & Time (all 8 depths) \\
\midrule
Texas          &     183 & 279  & exact & 0.01\,s \\
Wisconsin      &     251 & 450 & exact & 0.02\,s \\
Cornell        &     183 &      277 & exact   &     0.01\,s \\
Cora           &   2{,}708 &    5{,}278 & exact   & 3.30\,s \\
CiteSeer       &   3{,}327 &    4{,}552 & exact   & 1.90\,s \\
Chameleon      &   2{,}277 &   31{,}371 & exact   & 5.03\,s \\
Squirrel       &   5{,}201 &  198{,}353 & exact   & 57.16\,s \\
PubMed         &  19{,}717 &   44{,}324 & exact   & 959.27\,s \\
AmazonRatings  &  24{,}492 &   93{,}050 & sampled & 2.20\,s \\
RomanEmpire    &  22{,}662 &   32{,}927 & sampled & 0.11\,s \\
ogbn-arxiv     & 169{,}343 & 1{,}157{,}799 & sampled & 128.81\,s \\
\bottomrule
\end{tabular}
\end{table}

For comparison, training a single GCN model on ogbn-arxiv at depth $24$
takes approximately $566$\,seconds on CPU
(37 epochs with early stopping), plus $4$\,seconds to extract hidden
representations and compute Erank via SVD, a total of $570$\,seconds
for one (model, seed, depth) configuration.
Computing $\mathcal{D}_k$ across all eight depths on the same graph
costs $128.8$\,seconds with the sampled estimator at $s=200$, about $4.4\times$ less time than this combined
training-and-Erank computation. With sampling at $s=200$, the corresponding times are
$4.78$\,s on \textsc{PubMed}, $2.20$\,s on \textsc{Amazon-ratings},
and $0.11$\,s on \textsc{Roman-empire}
(Table~\ref{tab:runtime} reports exact computation for PubMed). A single trained model is one of the
$14\times10\times8=1{,}120$ configurations we evaluate per dataset to compute the post-training baselines.

\begin{remark}
Tables~\ref{tab:dataset-stats} and~\ref{tab:runtime} report distinct undirected edges after
symmetrization, merging duplicate edges, and removing
self-loops. The canonical diagnostic uses the lazy walk
constructed from this processed graph. The trained models
receive the same processed graph as input, with normalization
and self-loop additions determined by each architecture.
Self-loops introduced by the propagation operators are
not included in the reported edge counts.
\end{remark}
\section{Generalization across architectures}
\label{app:generalization}

The theoretical analysis in \S\ref{sec:theory} uses the lazy random walk $\mathbf{P} = \tfrac{1}{2}(\mathbf{I} + \mathbf{D}^{-1}\mathbf{A})$ as a parameter-free baseline. This appendix states how far the analysis extends beyond it. The spectral analysis extends, up to graph- and operator-dependent constants, to any reversible row-stochastic operator satisfying a set of checkable conditions, whose applicability we examine for row-normalized GCN and an idealized GraphSAGE-mean operator; APPNP's restart propagation and learned attention fall outside these conditions and are treated separately (\S\ref{app:gen-rowstoch}). The appendix then identifies sum aggregation as a regime where the simplex-geometric diagnostic does not directly apply (\S\ref{app:gen-sum}); proposes an augmented-scale variant $\mathcal{D}_k^{\mathrm{aug}}$ tailored to GIN and validates it empirically (\S\ref{app:gen-aug}); and reports an ablation over the choice of propagation operator $\mathbf{P}$ underlying the diagnostic itself (\S\ref{app:gen-ablation}).

\subsection{Row-stochastic operators: GCN, SAGE, APPNP, and attention}
\label{app:gen-rowstoch}

We extend the analysis of $\mathcal{D}_k$ from the lazy walk to associated propagation operators by way of a Dirichlet-form comparison. The argument is structured as follows: \S\ref{app:gen-conditions} states the admissibility conditions on a propagation operator $\mathbf{T}$ that suffice for $\mathcal{D}_k^{(\mathbf{T})}$ to track $\mathcal{D}_k^{(\mathrm{lazy})}$ at the level of exponential rates; \S\ref{app:gen-theorem} states and proves the universality theorem; \S\ref{app:gen-archcheck} verifies the conditions for the operators that admit a reversible closed form; \S\ref{app:gen-appnp} and \S\ref{app:gen-attention} treat APPNP and learned attention, to which the theorem does not apply.

\subsubsection{Admissibility conditions}
\label{app:gen-conditions}

Let $G$ be a connected, undirected, loop-free graph with $n\ge2$ nodes. Let $\mathbf{T}$ be a row-stochastic matrix on $V$ with stationary distribution $\tilde{\boldsymbol{\pi}}$, and write $Q_{vw}^{(\mathbf{T})} = \tilde{\pi}_v T_{vw}$ for its edge conductance on $(v,w) \in E$. The lazy walk has uniform conductance $Q_{vw}^{(\mathrm{lazy})} = 1/(2\mathrm{vol}(G))$ on every edge.

We assume the following.

\begin{itemize}
    \item \textbf{(C1) Stochasticity, graph support, irreducibility, and reversibility.}
    $\mathbf{T}\mathbf{1} = \mathbf{1}$, $T_{vw} \geq 0$, and $T_{vw} > 0$ for $v \neq w$ only if $(v,w) \in E$.
    The chain is irreducible and reversible with respect to its stationary distribution $\tilde{\boldsymbol{\pi}}$, i.e.\ $\tilde{\pi}_v T_{vw} = \tilde{\pi}_w T_{wv}$ for all $v,w$.

    \item \textbf{(C2) Stationary distribution comparability.}
    There exists $\rho_{\pi} \geq 1$ such that $\rho_{\pi}^{-1} \pi_v \leq \tilde{\pi}_v \leq \rho_{\pi} \pi_v$ for all $v$.

    \item \textbf{(C3) Conductance comparability.}
    There exist $0 < \alpha \leq \beta < \infty$ such that $\alpha Q_{vw}^{(\mathrm{lazy})} \leq Q_{vw}^{(\mathbf{T})} \leq \beta Q_{vw}^{(\mathrm{lazy})}$ for every $(v,w) \in E$.

    \item \textbf{(C4) Non-negative spectrum.}
    $\lambda_{\min}(\mathbf{T}) \geq 0$.

    \item \textbf{(C4$'$) Self-loop mass.}
    $2\min_v T_{vv} \geq \beta \rho_{\pi}^2 \gamma_{\mathrm{lazy}}$. This is a checkable sufficient substitute for (C4); see Lemma~\ref{lem:selfloop-mass} and the proof of Theorem~\ref{thm:lazy-universality}.

    \item \textbf{(C5) Rate-comparison condition.}
    $\beta \rho_{\pi}^2 \gamma_{\mathrm{lazy}} < 1$, where $\gamma_{\mathrm{lazy}} = 1 - \lambda_2(\mathbf{P}_{\mathrm{lazy}})$.
\end{itemize}

Conditions (C1)--(C3), together with (C4) or (C4$'$), are sufficient for the following rate characterization: $\mathcal{D}_k^{(\mathbf{T})}$ and $\mathcal{D}_k^{(\mathrm{lazy})}$ vanish together, at rates set by their respective spectral gaps. Conditions (C1)--(C3) alone do not guarantee convergence: on a bipartite graph the non-lazy walk $\mathbf{D}^{-1}\mathbf{A}$ satisfies (C1)--(C3) yet never collapses. Condition (C5) is required for the explicit interval comparison of decay rates and is graph-dependent.

\begin{lemma}[Self-loop mass bounds the negative spectrum]
\label{lem:selfloop-mass}
Let $\mathbf{T}$ be row-stochastic and reversible, and let $t = \min_v T_{vv}$. Then
\[
    \lambda_{\min}(\mathbf{T}) \;\geq\; 2t - 1,
    \qquad\text{equivalently}\qquad
    1 + \lambda_{\min}(\mathbf{T})
    \;\geq\; 2\min_v T_{vv}.
\]
\end{lemma}

\begin{proof}
If $t = 1$ then $\mathbf{T} = \mathbf{I}$ and the claim is trivial. Otherwise $\mathbf{S} := (\mathbf{T} - t\mathbf{I})/(1-t)$ is entrywise non-negative, row-stochastic, and reversible with respect to the same stationary distribution, so its eigenvalues are real and lie in $[-1,1]$. Every eigenvalue of $\mathbf{T}$ has the form $t + (1-t)\mu$ with $\mu$ an eigenvalue of $\mathbf{S}$, hence $\lambda_{\min}(\mathbf{T}) \geq t - (1-t) = 2t - 1$.
\end{proof}

\subsubsection{Universality theorem}
\label{app:gen-theorem}

\begin{lemma}[Spectral gap comparison]
\label{lem:gap}
Under (C1)--(C3), with $\gamma_{\mathbf{T}} = 1 - \lambda_2(\mathbf{T})$ and $\gamma_{\mathrm{lazy}} = 1 - \lambda_2(\mathbf{P}_{\mathrm{lazy}})$,
\begin{equation*}
    \frac{\alpha}{\rho_{\pi}^2}\gamma_{\mathrm{lazy}}
    \leq \gamma_{\mathbf{T}}
    \leq \beta\rho_{\pi}^2\gamma_{\mathrm{lazy}}.
\end{equation*}
\end{lemma}

\begin{proof}
The Dirichlet forms of the two reversible chains are
\[
    \mathcal{E}_{\mathbf{T}}(f,f)
    = \sum_{\{v,w\}\in E}
      Q_{vw}^{(\mathbf{T})}(f(v)-f(w))^2,
    \qquad
    \mathcal{E}_{\mathrm{lazy}}(f,f)
    = \sum_{\{v,w\}\in E}
      Q_{vw}^{(\mathrm{lazy})}(f(v)-f(w))^2.
\]
By (C3), for all $f:V\to\mathbb{R}$,
\begin{equation}
\label{eq:dirichlet_sandwich}
    \alpha\mathcal{E}_{\mathrm{lazy}}(f,f)
    \leq \mathcal{E}_{\mathbf{T}}(f,f)
    \leq \beta\mathcal{E}_{\mathrm{lazy}}(f,f).
\end{equation}

The spectral gaps admit the variational characterization
\[
    \gamma_{\mathbf{T}}
    = \min_{\operatorname{Var}_{\tilde{\pi}}(f)>0}
      \frac{\mathcal{E}_{\mathbf{T}}(f,f)}
           {\operatorname{Var}_{\tilde{\pi}}(f)},
    \qquad
    \gamma_{\mathrm{lazy}}
    = \min_{\operatorname{Var}_{\pi}(f)>0}
      \frac{\mathcal{E}_{\mathrm{lazy}}(f,f)}
           {\operatorname{Var}_{\pi}(f)}.
\]
Using
\[
    \operatorname{Var}_{\mu}(f)
    = \frac{1}{2}\sum_{v,w}
      \mu_v\mu_w(f(v)-f(w))^2
\]
and condition (C2), we obtain
\[
    \rho_{\pi}^{-2}\operatorname{Var}_{\pi}(f)
    \leq \operatorname{Var}_{\tilde{\pi}}(f)
    \leq \rho_{\pi}^{2}\operatorname{Var}_{\pi}(f).
\]
Combining this with \eqref{eq:dirichlet_sandwich} gives
\[
    \gamma_{\mathbf{T}}
    = \min_f
      \frac{\mathcal{E}_{\mathbf{T}}(f,f)}
           {\operatorname{Var}_{\tilde{\pi}}(f)}
    \geq \min_f
      \frac{\alpha\mathcal{E}_{\mathrm{lazy}}(f,f)}
           {\rho_{\pi}^2\operatorname{Var}_{\pi}(f)}
    = \frac{\alpha}{\rho_{\pi}^2}\gamma_{\mathrm{lazy}}.
\]
The upper bound is symmetric.
\end{proof}

\begin{lemma}[Single-chain decay rate]
\label{lem:decay-single}
Let $\mathbf{M}$ be a reversible irreducible chain on $G$ with stationary distribution $\mu$, spectral gap $\gamma = 1 - \lambda_2(\mathbf{M})$, and $\max_{i\geq2}|\lambda_i(\mathbf{M})| = \lambda_2(\mathbf{M})$ (which holds in particular when the spectrum is non-negative), and let $\{f_i\}$ be an $L^2(\mu)$-orthonormal eigenbasis with $f_1\equiv1$. Then
\begin{equation*}
    c_{\mathbf{M}}(1-\gamma)^{2k}
    \leq \mathcal{D}_k^{(\mathbf{M})}
    \leq C_{\mathbf{M}}(1-\gamma)^{2k},
\end{equation*}
for $k\geq0$, with the convention $0^0=1$, and with $C_{\mathbf{M}} = \tfrac{2}{n}\sum_v(\mu_v^{-1}-1)$ and $c_{\mathbf{M}} = \tfrac{1}{4}\inf_v\mu_v\cdot\operatorname{Var}_{\mathrm{unif}}(f_2)$. In particular, if $0<\lambda_2(\mathbf{M})<1$, then $\mathcal{D}_k^{(\mathbf{M})} = \Theta((1-\gamma)^{2k})$.
\end{lemma}

\begin{proof}
We will prove the two sides of the inequality one by one.

\noindent\textbf{Upper bound.}
By the Hellinger triangle inequality, for any reference distribution $\mu^*$,
\[
    H^2(p_v^{(k)},p_u^{(k)})
    \leq 2\bigl(
        H^2(p_v^{(k)},\mu^*)
        + H^2(p_u^{(k)},\mu^*)
    \bigr).
\]
Choosing $\mu^*=\mu$ and averaging over all pairs $(v,u)$ yields
\[
    \mathcal{D}_k^{(\mathbf{M})}
    = \frac{1}{n^2}\sum_{v,u}
      H^2(p_v^{(k)},p_u^{(k)})
    \leq \frac{4}{n}\sum_v H^2(p_v^{(k)},\mu).
\]
By the Hellinger-chi-squared inequality, $H^2(p,q)\leq\frac{1}{2}\chi^2(p\|q)$. The spectral decomposition gives
\[
    \frac{p_v^{(k)}(w)}{\mu(w)}
    = \sum_{i=1}^n\lambda_i^k f_i(v)f_i(w),
\]
from which
\[
    \chi^2(p_v^{(k)}\|\mu)
    = \sum_w\mu(w)
      \biggl(\frac{p_v^{(k)}(w)}{\mu(w)}-1\biggr)^2
    = \sum_{i\geq2}\lambda_i^{2k}f_i(v)^2
    \leq \lambda_2^{2k}
      \biggl(\frac{1}{\mu_v}-1\biggr),
\]
where the inequality uses $|\lambda_i|\leq\lambda_2$ for $i\geq2$ and $\sum_{i\geq1}f_i(v)^2=1/\mu_v$.
Therefore
\[
    \mathcal{D}_k^{(\mathbf{M})}
    \leq \frac{4}{n}\sum_v
      \frac{1}{2}\lambda_2^{2k}
      \biggl(\frac{1}{\mu_v}-1\biggr)
    = \frac{2}{n}\sum_v
      \biggl(\frac{1}{\mu_v}-1\biggr)(1-\gamma)^{2k}.
\]

\medskip
\noindent\textbf{Lower bound.}
Write $p_v^{(k)}(w)=\mu(w)(1+g_v(w))$, where
\[
    g_v(w)
    = \sum_{i\geq2}\lambda_i^k f_i(v)f_i(w).
\]
Define $\psi_v(w)=\sqrt{p_v^{(k)}(w)/\mu(w)}=\sqrt{1+g_v(w)}$. Then
\[
    H^2(p_v^{(k)},p_u^{(k)})
    = \frac{1}{2}\sum_w\mu(w)
      \bigl(\psi_v(w)-\psi_u(w)\bigr)^2.
\]
Since $\psi_v(w)\leq1/\sqrt{\mu(w)}$ and $\psi_u(w)\leq1/\sqrt{\mu(w)}$, the identity
\[
    g_v(w)-g_u(w)
    = \bigl(\psi_v(w)-\psi_u(w)\bigr)
      \bigl(\psi_v(w)+\psi_u(w)\bigr)
\]
implies
\[
    \bigl(\psi_v(w)-\psi_u(w)\bigr)^2
    \geq \frac{\mu(w)}{4}
      \bigl(g_v(w)-g_u(w)\bigr)^2.
\]
Hence
\[
    H^2(p_v^{(k)},p_u^{(k)})
    \geq \frac{1}{8}\sum_w\mu(w)^2
      \bigl(g_v(w)-g_u(w)\bigr)^2
    \geq \frac{\inf_w\mu(w)}{8}
      \sum_w\mu(w)\bigl(g_v(w)-g_u(w)\bigr)^2.
\]
Projecting onto the $f_2$ direction and using $L^2(\mu)$-orthonormality,
\[
    \sum_w\mu(w)
    \bigl(\lambda_2^k(f_2(v)-f_2(u))f_2(w)\bigr)^2
    = \lambda_2^{2k}(f_2(v)-f_2(u))^2.
\]
Therefore
\[
    H^2(p_v^{(k)},p_u^{(k)})
    \geq \frac{\inf_w\mu(w)}{8}
      \lambda_2^{2k}(f_2(v)-f_2(u))^2.
\]
Averaging over all pairs $(v,u)$ and using
\[
    \frac{1}{n^2}\sum_{v,u}(f_2(v)-f_2(u))^2
    = 2\operatorname{Var}_{\mathrm{unif}}(f_2)
\]
gives
\[
    \mathcal{D}_k^{(\mathbf{M})}
    \geq \frac{\inf_v\mu_v}{4}
      \operatorname{Var}_{\mathrm{unif}}(f_2)
      (1-\gamma)^{2k}.
    \qedhere
\]
\end{proof}

\begin{theorem}[Lazy-walk universality for reversible operators]
\label{thm:lazy-universality}
Let $\mathbf{T}$ satisfy (C1)--(C3) and either (C4) or (C4$'$).
\begin{itemize}
    \item[(i)] Both dispersions vanish:
    $\mathcal{D}_k^{(\mathbf{T})}\to0$ if and only if
    $\mathcal{D}_k^{(\mathrm{lazy})}\to0$.

    \item[(ii)] Under the additional condition (C5)
    and $n\geq3$, both chains satisfy
    $\mathcal{D}_k=\Theta((1-\gamma)^{2k})$ with their respective gaps, and
    \begin{equation*}
    \frac{\log(1-\alpha\rho_{\pi}^{-2}\gamma_{\mathrm{lazy}})}
         {\log(1-\gamma_{\mathrm{lazy}})}
    \leq
    \lim_{k\to\infty}
    \frac{\log\mathcal{D}_k^{(\mathbf{T})}}
         {\log\mathcal{D}_k^{(\mathrm{lazy})}}
    \leq
    \frac{\log(1-\beta\rho_{\pi}^{2}\gamma_{\mathrm{lazy}})}
         {\log(1-\gamma_{\mathrm{lazy}})}.
    \end{equation*}
\end{itemize}
\end{theorem}

\begin{proof}
Because $\mathbf{T}$ is finite, reversible, and irreducible, its stationary distribution is unique and strictly positive. We first check the spectral hypothesis of Lemma~\ref{lem:decay-single}, $\max_{i\geq2}|\lambda_i|=\lambda_2$. For $\mathbf{P}_{\mathrm{lazy}}$, and for $\mathbf{T}$ under (C4), the spectrum is non-negative and there is nothing to prove. Under (C4$'$), Lemma~\ref{lem:selfloop-mass} gives $\lambda_{\min}(\mathbf{T})\geq2\min_vT_{vv}-1$ and Lemma~\ref{lem:gap} gives $\gamma_{\mathbf{T}}\leq\beta\rho_{\pi}^2\gamma_{\mathrm{lazy}}$, so
\[
    \lambda_2(\mathbf{T})
    = 1-\gamma_{\mathbf{T}}
    \geq 1-\beta\rho_{\pi}^2\gamma_{\mathrm{lazy}}
    \geq 1-2\min_vT_{vv}
    \geq -\lambda_{\min}(\mathbf{T}),
\]
and every eigenvalue other than $\lambda_1=1$ lies in
$[\lambda_{\min}(\mathbf{T}),\lambda_2(\mathbf{T})]
\subseteq[-\lambda_2(\mathbf{T}),\lambda_2(\mathbf{T})]$.
By Lemma~\ref{lem:decay-single} applied to $\mathbf{T}$ and to $\mathbf{P}_{\mathrm{lazy}}$, each chain has dispersion converging to zero, which proves part (i).

For part (ii), condition (C5) implies $\beta\rho_{\pi}^2\gamma_{\mathrm{lazy}}<1$, hence by Lemma~\ref{lem:gap}
\[
    0 < \frac{\alpha}{\rho_{\pi}^2}\gamma_{\mathrm{lazy}}
    \leq \gamma_{\mathbf{T}}
    \leq \beta\rho_{\pi}^2\gamma_{\mathrm{lazy}}
    < 1.
\]
Thus $0<\gamma_{\mathbf{T}}<1$. Moreover $0<\gamma_{\mathrm{lazy}}<1$ whenever $n\geq3$: the eigenvalues of $\mathbf{D}^{-1}\mathbf{A}$ sum to zero, so its second-largest eigenvalue exceeds $-1$ and $\lambda_2(\mathbf{P}_{\mathrm{lazy}})>0$. Hence the two-sided exponential estimates from Lemma~\ref{lem:decay-single} apply to both chains. Taking logarithms and dividing by $2k$ gives
\[
    \lim_{k\to\infty}
    \frac{\log\mathcal{D}_k^{(\mathbf{M})}}{2k}
    = \log(1-\gamma_{\mathbf{M}})
\]
for $\mathbf{M}\in\{\mathbf{T},\mathbf{P}_{\mathrm{lazy}}\}$. Therefore
\[
    \lim_{k\to\infty}
    \frac{\log\mathcal{D}_k^{(\mathbf{T})}}
         {\log\mathcal{D}_k^{(\mathrm{lazy})}}
    =
    \frac{\log(1-\gamma_{\mathbf{T}})}
         {\log(1-\gamma_{\mathrm{lazy}})}.
\]
Substituting the gap bounds from Lemma~\ref{lem:gap} and using that $x\mapsto\log(1-x)$ is decreasing on $(0,1)$ yields the stated interval.
\end{proof}

\begin{remark}[What happens without (C4) or (C4$'$)]
\label{rem:C4prime}
An eigenmode with eigenvalue $\lambda$ decays as $|\lambda|^k$, so when $0<r_\star:=\max_{i\geq2}|\lambda_i(\mathbf{T})|<1$, the dispersion decays as $\Theta(r_\star^{2k})$; conditions (C4) and (C4$'$) serve only to guarantee $r_\star=\lambda_2(\mathbf{T})$. When neither holds but $\min_vT_{vv}>0$, Lemma~\ref{lem:selfloop-mass} still gives $r_\star<1$, so $\mathcal{D}_k^{(\mathbf{T})}\to0$, and the upper bound of Lemma~\ref{lem:decay-single} holds with $r_\star^{2k}$ in place of $(1-\gamma)^{2k}$.
\end{remark}

\subsubsection{Architecture verification}
\label{app:gen-archcheck}

Table~\ref{tab:arch-conditions} summarises the conditions for the reversible closed-form operators related to our architectures. Conductances are compared against $Q_{vw}^{(\mathrm{lazy})}=1/(2\mathrm{vol}(G))$.

\begin{table}[H]
\centering
\caption{Conditions (C1)--(C4) for the reversible closed-form operators related to our architectures. The lazy kernel in the last row is a one-step surrogate for APPNP, whose actual propagation is not a power of a fixed operator (\S\ref{app:gen-appnp}). Self-loop notation: $\hat{\mathbf{A}}=\mathbf{A}+\mathbf{I}$, $\hat{d}_v=d_v+1$, $\hat{\pi}_v=\hat{d}_v/(\mathrm{vol}(G)+n)$. Condition (C5) depends on $\gamma_{\mathrm{lazy}}$ and is graph-dependent.}
\label{tab:arch-conditions}
\renewcommand{\arraystretch}{1.25}
\small
\begin{tabular}{lccccc}
\toprule
Architecture & Associated $\mathbf{T}$ & Reversible? & $\tilde{\boldsymbol{\pi}}$ & $\alpha=\beta$ & (C4) \\
\midrule
SAGE-mean proxy
& $\tfrac{1}{2}(\mathbf{I}+\mathbf{D}^{-1}\mathbf{A})$
& Yes & $\boldsymbol{\pi}$ & $1$ & Yes \\
GCN (row-norm)
& $\hat{\mathbf{D}}^{-1}\hat{\mathbf{A}}$
& Yes & $\hat{\boldsymbol{\pi}}$
& $\frac{2\mathrm{vol}(G)}{\mathrm{vol}(G)+n}$
& Not auto. \\
Lazy kernel
& $\alpha_{\mathrm{tel}}\mathbf{I}+(1-\alpha_{\mathrm{tel}})\hat{\mathbf{D}}^{-1}\hat{\mathbf{A}}$
& Yes & $\hat{\boldsymbol{\pi}}$
& $\frac{2(1-\alpha_{\mathrm{tel}})\mathrm{vol}(G)}{\mathrm{vol}(G)+n}$
& Yes if $\alpha_{\mathrm{tel}}\geq1/2$ \\
\bottomrule
\end{tabular}
\end{table}

\paragraph{Derivations.}
In what follows, we derive the quantities listed in Table~\ref{tab:arch-conditions} for the three operators.

For the two operators built on $\hat{\mathbf{A}}$ (row-normalized GCN and the lazy kernel), condition (C2) holds with $\rho_{\pi}=\rho_{\mathrm{sl}}$, where
\begin{equation*}
\rho_{\mathrm{sl}}
=
\max\left\{
\frac{\mathrm{vol}(G)}{\mathrm{vol}(G)+n}
\left(1+\frac{1}{d_{\min}}\right),
\quad
\frac{\mathrm{vol}(G)+n}{\mathrm{vol}(G)}
\cdot\frac{d_{\max}}{d_{\max}+1}
\right\}.
\end{equation*}

\smallskip\noindent
\emph{SAGE-mean.}
Here $T_{vw}=1/(2d_v)$ for $w\sim v$ and $T_{vv}=1/2$.
The chain is reversible with respect to the ordinary stationary distribution
$\boldsymbol{\pi}$ and
\[
    Q_{vw}^{(\mathbf{T})}
    = \pi_vT_{vw}
    = \frac{d_v}{\operatorname{vol}(G)}
      \cdot\frac{1}{2d_v}
    = \frac{1}{2\operatorname{vol}(G)}.
\]
Hence $\alpha=\beta=1$ and $\rho_{\pi}=1$. This operator coincides exactly with the lazy random walk. It is the aggregation operator of an idealized GraphSAGE-mean layer in which the root and mean-neighbor terms receive equal weight and share one feature transformation. The \texttt{SAGEConv} layer used in our experiments learns separate transformations for the two terms, so the theory covers this idealized operator, while the full learned architecture is evaluated empirically (Table~\ref{tab:per_arch}).

\smallskip\noindent
\emph{GCN (row-normalised).}
For every $w\sim v$ or $w=v$, $T_{vw}=1/\hat{d}_v$.
The chain is reversible with respect to
$\hat{\boldsymbol{\pi}}$ (the self-loop corrected distribution) and
\[
    Q_{vw}^{(\mathrm{GCN})}
    = \hat{\pi}_vT_{vw}
    = \frac{1}{\operatorname{vol}(G)+n}
\]
for every edge $(v,w)\in E$.
Since $Q_{vw}^{(\mathrm{lazy})}=1/(2\operatorname{vol}(G))$, we obtain
\[
    \alpha=\beta
    = \frac{2\operatorname{vol}(G)}{\operatorname{vol}(G)+n}.
\]
Condition~(C4) (non-negative spectrum) is not automatic.
For the three-vertex path one has, for instance,
\[
    \hat{\mathbf{D}}^{-1}\hat{\mathbf{A}}
    =
    \begin{pmatrix}
    1/2 & 1/2 & 0 \\
    1/3 & 1/3 & 1/3 \\
    0   & 1/2 & 1/2
    \end{pmatrix},
\]
whose eigenvalues are $1$, $1/2$ and $-1/6$.
Lemma~\ref{lem:selfloop-mass} quantifies how negative the spectrum can be: here $T_{vv}=1/\hat{d}_v$, so $\lambda_{\min}\geq2/(d_{\max}+1)-1$ (on the three-vertex path, $-1/6\geq-1/3$), and (C4$'$) reads $2/(d_{\max}+1)\geq\beta\rho_{\pi}^2\gamma_{\mathrm{lazy}}$. This is checkable from the graph alone, but it is restrictive: it can fail on graphs with large maximum degree, and the general decay rate is governed by $r_\star=\max_{i\geq2}|\lambda_i|$, which need not equal $\lambda_2$ (Remark~\ref{rem:C4prime}).

\smallskip\noindent
\emph{Lazy kernel with holding probability $\alpha_{\mathrm{tel}}$ (one-step surrogate for APPNP).}
With $\hat{\mathbf{A}}=\mathbf{A}+\mathbf{I}$ and holding probability $0\leq\alpha_{\mathrm{tel}}<1$, the operator is
\[
    \mathbf{T}
    = \alpha_{\mathrm{tel}}\mathbf{I}
      +(1-\alpha_{\mathrm{tel}})
      \hat{\mathbf{D}}^{-1}\hat{\mathbf{A}}.
\]
It is reversible with respect to $\hat{\boldsymbol{\pi}}$ and, for
$(v,w)\in E$,
\[
    Q_{vw}^{(\mathbf{T})}
    = \hat{\pi}_v
      \frac{1-\alpha_{\mathrm{tel}}}{\hat{d}_v}
    = \frac{1-\alpha_{\mathrm{tel}}}
           {\operatorname{vol}(G)+n}.
\]
Hence
\[
    \alpha=\beta
    = \frac{2(1-\alpha_{\mathrm{tel}})
             \operatorname{vol}(G)}
           {\operatorname{vol}(G)+n}.
\]
Writing $\mathbf{K}=\hat{\mathbf{D}}^{-1}\hat{\mathbf{A}}$, every eigenvalue of $\mathbf{K}$ lies in $[-1,1]$. Consequently, the eigenvalues of
$\mathbf{T}=\alpha_{\mathrm{tel}}\mathbf{I}
+(1-\alpha_{\mathrm{tel}})\mathbf{K}$ lie in
\[
    [2\alpha_{\mathrm{tel}}-1,\,1].
\]
Thus (C4) holds whenever $\alpha_{\mathrm{tel}}\geq1/2$; for smaller $\alpha_{\mathrm{tel}}$ it may fail on certain graphs. This kernel keeps mass at the current node during each step. It is a laziness sweep and not the APPNP propagation, which restarts toward the input signal at every layer and is analyzed in \S\ref{app:gen-appnp}.

\begin{remark}[Special cases of Lemma~\ref{lem:selfloop-mass}]
\label{rem:criterion-C4}
If $\mathbf{T}$ is reversible and $T_{vv}\geq1/2$ for every $v$, Lemma~\ref{lem:selfloop-mass} gives $\lambda_{\min}(\mathbf{T})\geq0$, that is, (C4). This applies immediately to SAGE-mean, and to the lazy kernel whenever $\alpha_{\mathrm{tel}}\geq1/2$. It does not apply to GCN in general.
\end{remark}

\begin{remark}[What the GCN bridge transfers]
\label{rem:gcn-bridge}
GCN as implemented uses the symmetric operator
$\mathbf{S}=\hat{\mathbf{D}}^{-1/2}\hat{\mathbf{A}}\hat{\mathbf{D}}^{-1/2}$,
which is related to the row-normalized operator
$\mathbf{T}_{\mathrm{rw}}=\hat{\mathbf{D}}^{-1}\hat{\mathbf{A}}$
of Table~\ref{tab:arch-conditions} by the diagonal similarity
$\mathbf{S}=\hat{\mathbf{D}}^{1/2}\mathbf{T}_{\mathrm{rw}}\hat{\mathbf{D}}^{-1/2}$.
Diagonal similarity preserves eigenvalues, so the same nontrivial eigenvalues control convergence of the powers of both operators to their rank-one limits. For $\mathbf{S}$, the limiting rows are proportional to $\sqrt{\hat{\boldsymbol{\pi}}}^{\top}$ and need not be identical.
It does \emph{not} transfer dispersion values: $\mathbf{S}$ need not have probability rows, and $\mathcal{D}_k$ applies the nonlinear square-root map to rows, neither of which is invariant under a diagonal change of coordinates. The bridge is therefore a statement about decay rates and not about numerical values of $\mathcal{D}_k$; the analogous comparison for the lazy walk is made empirically in Table~\ref{tab:ablation-sym}.
\end{remark}

\begin{remark}[Condition (C5) must be checked separately]
\label{rem:C5}
The table checks (C1)--(C3) and reports when (C4) holds. The explicit logarithmic rate interval in Theorem~\ref{thm:lazy-universality} also requires (C5), namely $\beta\rho_{\pi}^2\gamma_{\mathrm{lazy}}<1$. This is a graph-architecture pair condition and can fail even when (C1)--(C4) hold.
For example, for GCN on the complete graph $K_n$, one has $\rho_{\pi}=1$,
$\beta=2(n-1)/n$, and $\gamma_{\mathrm{lazy}}=n/(2(n-1))$, so
$\beta\rho_{\pi}^2\gamma_{\mathrm{lazy}}=1$.
\end{remark}

\begin{remark}[Role of $\rho_{\pi}$]
\label{rem:rho}
For the SAGE-mean proxy, $\rho_{\pi}=1$. For the two self-loop-corrected operators above, $\rho_{\pi}=\rho_{\mathrm{sl}}\leq2$, since $d_v\geq1$. Thus stationary-distribution comparability is uniformly bounded for all three operators.
\end{remark}

\subsubsection{APPNP requires its own operator}
\label{app:gen-appnp}

APPNP does not propagate by powers of a fixed operator. With input signal $\mathbf{H}^{(0)}$, a one-step row-stochastic kernel $\mathbf{K}$, and restart probability $0<\alpha\leq1$, its layers read
$\mathbf{H}^{(\ell+1)}
=\alpha\mathbf{H}^{(0)}
+(1-\alpha)\mathbf{K}\mathbf{H}^{(\ell)}$.
Writing $\mathbf{H}^{(k)}=\mathbf{Q}_k\mathbf{H}^{(0)}$, the depth-$k$ operator satisfies $\mathbf{Q}_0=\mathbf{I}$ and
$\mathbf{Q}_{\ell+1}
=\alpha\mathbf{I}+(1-\alpha)\mathbf{K}\mathbf{Q}_\ell$,
and by repeated substitution
\begin{equation}
\label{eq:appnp-qk}
    \mathbf{Q}_k
    = \alpha\sum_{i=0}^{k-1}(1-\alpha)^i\mathbf{K}^i
      +(1-\alpha)^k\mathbf{K}^k.
\end{equation}
The finite sum collects the input signal re-injected at every layer, and the last term is the part that has not been restarted by depth $k$.
$\mathbf{Q}_k$ is row-stochastic, but for $0<\alpha<1$ and $\mathbf{K}\neq\mathbf{I}$ it is not the $k$-th power of a single operator ($\mathbf{Q}_2\neq\mathbf{Q}_1^2$), so Theorem~\ref{thm:lazy-universality} does not apply to it.
As $k\to\infty$,
$\mathbf{Q}_k\to\mathbf{Q}_\infty
=\alpha(\mathbf{I}-(1-\alpha)\mathbf{K})^{-1}$,
whose rows are node-specific personalized-PageRank distributions.
Since $\mathbf{Q}_\infty$ is invertible, it cannot have rank one on a graph with more than one node, so by Theorem~\ref{thm:characterisation} its dispersion is strictly positive: APPNP has a positive dispersion floor by design.
This difference is consistent with the weaker correlation of the lazy-walk $\mathcal{D}_k$ on APPNP (Table~\ref{tab:per_arch}). Computing $\mathcal{D}_k$ from $\mathbf{Q}_k$ instead raises its observed mean absolute correlation from $0.552$ to $0.624$ (\S\ref{app:matched-operator}). The released implementation uses the standard APPNP restart with symmetric GCN normalization.

\subsubsection{Learned attention: what survives without reversibility}
\label{app:gen-attention}

Each realized attention matrix $\mathbf{T}^{(\ell)}$ of a GAT head, with attention dropout disabled, is row-stochastic, learned, layer-dependent, and generally non-reversible, so neither the spectral comparison of Theorem~\ref{thm:lazy-universality} nor the fixed-operator data-processing argument of Theorem~\ref{thm:monotone} applies as stated.
Write $\mathbf{M}_k=\mathbf{T}^{(k)}\cdots\mathbf{T}^{(1)}$ for the depth-$k$ coefficient product, $p_v^{(k)}$ for its rows, and $\mathcal{D}_k$ for their dispersion.
The guarantees sort into three groups.

\paragraph{Transfers exactly.}
Theorem~\ref{thm:characterisation} uses only row-stochasticity, so $\mathcal{D}_k=0$ if and only if $\mathbf{M}_k$ has rank one.

\paragraph{Transfers in weakened form.}
Monotonicity of the mean dispersion does not transfer.
A new layer multiplies the product on the \emph{left}, so each new row is a convex combination of old rows,
$p_v^{(k+1)}=\sum_uT^{(k+1)}_{vu}p_u^{(k)}$;
this is not the right-multiplication covered by the data-processing inequality.
A four-node example shows that the mean dispersion can increase.
Let the rows of $\mathbf{M}_k$ be $p_1=p_2=p_3=e_1$ and $p_4=e_2$, so that $6$ of the $16$ ordered pairs have $H^2=1$ and $\mathcal{D}_k=6/16$.
If in the next layer rows $1$ and $3$ attend to node $1$ and rows $2$ and $4$ attend to node $4$, the new rows are $e_1,e_2,e_1,e_2$, with $8$ ordered pairs at $H^2=1$ and $\mathcal{D}_{k+1}=8/16>\mathcal{D}_k$.
By continuity, the same strict increase occurs for strictly positive attention weights close to these.

What does hold is monotonicity of the \emph{worst-pair} dispersion
$\tilde{\mathcal{D}}_k
=\max_{u,v}H^2(p_u^{(k)},p_v^{(k)})
\geq\mathcal{D}_k$.
Because $H^2$ is jointly convex,
\[
    H^2(p_u^{(k+1)},p_v^{(k+1)})
    \leq
    \sum_{i,j}T^{(k+1)}_{ui}T^{(k+1)}_{vj}
    H^2(p_i^{(k)},p_j^{(k)})
    \leq \tilde{\mathcal{D}}_k,
\]
so $\tilde{\mathcal{D}}_{k+1}\leq\tilde{\mathcal{D}}_k$.

Collapse can be guaranteed under explicit conditions.
Assume that all layers share a fixed connected edge support that includes self-loops, and that every attention coefficient on that support is at least $a_{\min}>0$ at every depth; for softmax attention with logits bounded by $B$ in absolute value, one can take
$a_{\min}=e^{-2B}/(d_{\max}+1)$.
Then any product of $\operatorname{diam}(G)$ consecutive layers has all entries at least $a_{\min}^{\operatorname{diam}(G)}$, so any two of its rows share at least $n\,a_{\min}^{\operatorname{diam}(G)}$ of their mass.
Left-multiplication by such a block contracts the largest pairwise total-variation distance among the rows by the factor
$1-n\,a_{\min}^{\operatorname{diam}(G)}$
(a Dobrushin-type coupling argument), and with
$H^2\leq\mathrm{TV}$ (Lemma~\ref{lem:hellinger-tv}),
\[
    \mathcal{D}_k
    \leq \tilde{\mathcal{D}}_k
    \leq
    \Bigl(1-n\,a_{\min}^{\operatorname{diam}(G)}\Bigr)^{
        \lfloor k/\operatorname{diam}(G)\rfloor}.
\]
This is an asymptotic statement about the coefficient product.
It does not say that any finite product has rank one, and it does not describe the complete nonlinear network (feature maps, activations, feature-dependent attention, skip paths).
The GAT and GATv2 runs in our experiments impose no minimum-attention bound, so we do not present it as evidence that these conditions held in those runs.

\paragraph{Does not transfer.}
The matching lower bound of Theorem~\ref{thm:decay}(ii) assumes a fixed reversible operator.
A two-sided rate for attention would need additional control of the layer-dependent operators and is outside the scope of this paper. Empirically, the lazy-walk $\mathcal{D}_k$ nonetheless has the highest observed mean absolute correlation on both attention architectures ($0.785$ on GATv1, next best $E_{\mathrm{Dir,n}}$ at $0.715$; $0.798$ on GATv2, next best $E_{\mathrm{Proj,n}}$ at $0.784$; Table~\ref{tab:per_arch}). These results support the lazy walk as a surrogate for the evaluated attention models.

\subsection{Sum aggregation: why GIN falls outside the framework}
\label{app:gen-sum}

GIN \citep{xu2019powerful} departs from the operators above in a way that is fundamental to its design rather than incidental. For fixed $\varepsilon$, its update rule is
\begin{equation}
    \mathbf{H}^{(\ell+1)}
    =
    \mathrm{MLP}^{(\ell)}\!\left(
        (1+\varepsilon)\mathbf{H}^{(\ell)}
        +\mathbf{A}\mathbf{H}^{(\ell)}
    \right),
\end{equation}
where the aggregation operator $(1+\varepsilon)\mathbf{I}+\mathbf{A}$ uses a \emph{sum} over neighbors rather than a mean.
This choice is the central motivation of GIN: \citet{xu2019powerful} prove that mean and max aggregators cannot distinguish certain non-isomorphic neighborhoods that the sum aggregator can, and that under their injectivity assumptions on aggregation, updates, and readout, GIN can match the discriminative power of the Weisfeiler--Lehman test.
The price of this expressivity is that the aggregation operator is no longer row-stochastic.

This has a direct consequence for the theory developed in \S\ref{sec:theory}.

\textit{The rows of $\mathbf{P}_{\mathrm{GIN}}^k$ need not be probability distributions.}
The square-root embedding $\phi$ is defined on the simplex $\Delta^{n-1}$, and the diagnostic $\mathcal{D}_k$ measures the variance of the embedded point cloud on $S_+^{n-1}$.
Without row-stochasticity, the rows of
$\mathbf{P}_{\mathrm{GIN}}^k=((1+\varepsilon)\mathbf{I}+\mathbf{A})^k$
have $\ell_1$ norms that are not constrained to one and, for fixed $\varepsilon\geq0$ on a connected graph with $n\geq2$, grow exponentially in $k$.
Renormalizing each row to unit $\ell_1$ norm would produce a row-stochastic surrogate, but this surrogate is no longer the operator GIN uses; it discards the magnitude information that the sum aggregator was designed to preserve.

\subsection{An augmented-scale diagnostic for sum aggregation}
\label{app:gen-aug}

Sum aggregation preserves count information that mean aggregation discards, motivating a diagnostic for GIN that tracks both the angular structure of the propagated rows and their magnitudes.
We construct $\mathcal{D}_k^{\mathrm{aug}}$, an augmented-scale variant of the aggregation dispersion designed for non-row-stochastic operators, by lifting each row of the propagation matrix into a higher-dimensional simplex with an explicit mass coordinate.

\paragraph{Definition.}
Let $\mathbf{M}_{\varepsilon}=(1+\varepsilon)\mathbf{I}+\mathbf{A}$ denote the GIN aggregation operator on a connected undirected graph with $n\geq2$ and fixed $\varepsilon\geq0$, and $\lambda_1=\lambda_1(\mathbf{M}_{\varepsilon})$ its leading eigenvalue.
For each node $v$, let
$s_v^{(k)}=\|\mathbf{M}_{\varepsilon}^k e_v\|_1$
denote the $\ell_1$ mass of the $v$-th column at depth $k$ (equal to that of the $v$-th row by symmetry), and define the rescaled walk vector
\begin{equation}
    \tilde{p}_v^{(k)}
    = \frac{\mathbf{M}_{\varepsilon}^k e_v}{\lambda_1^k}
    \in\mathbb{R}_{\geq0}^n.
\end{equation}
The rescaling by $\lambda_1^k$ stabilizes the magnitudes against the exponential growth at rate $\lambda_1^k$ (\S\ref{app:gen-sum}), placing the rescaled vectors in a bounded regime regardless of depth.
We then form the augmented walk
\begin{equation}
\label{eq:aug-embedding}
    p_v^{(k),\mathrm{aug}}
    =
    \frac{1}{s_v^{(k)}/\lambda_1^k+1}
    \begin{pmatrix}
        \tilde{p}_v^{(k)}\\
        1
    \end{pmatrix}
    \in\Delta^n,
\end{equation}
which appends a mass coordinate equal to $1$ before normalization, then divides by the total to land on the $n$-dimensional simplex $\Delta^n$.
Two nodes with identical angular structure but different total mass produce walks that differ in the appended coordinate; two nodes with identical mass but different angular structure produce walks that differ in the first $n$ coordinates.
The augmented dispersion is then the standard construction applied to these augmented walks:
\begin{equation}
\label{eq:dk-aug}
    \mathcal{D}_k^{\mathrm{aug}}
    =
    1-\left\|
        \frac{1}{n}\sum_v\sqrt{p_v^{(k),\mathrm{aug}}}
    \right\|_2^2.
\end{equation}

\paragraph{Properties.}
The augmented walk $p_v^{(k),\mathrm{aug}}$ lies on the standard simplex $\Delta^n$ by construction, so $\mathcal{D}_k^{\mathrm{aug}}$ has the following properties:

\textit{Collapse characterization.}
$\mathcal{D}_k^{\mathrm{aug}}=0$ if and only if all augmented embeddings coincide, which by \eqref{eq:aug-embedding} requires both that all nodes have identical row directions \emph{and} identical total mass.

\textit{Closed-form centroid.}
The reformulation \eqref{eq:dk-aug} retains the $O(n^2)$ centroid evaluation of the original $\mathcal{D}_k$, since the augmented embedding adds only a single coordinate per node.

\textit{Concentration.}
For $s\geq2$ independent uniform source nodes sampled with replacement, the McDiarmid argument of Appendix~\ref{app:estimator-concentration} transfers directly: each $p_v^{(k),\mathrm{aug}}$ has unit $\ell_1$ norm, so the bounded-differences constant remains $4/(s-1)$ and the sample complexity bound $s=O(\epsilon^{-2}\log(1/\delta))$ is unchanged.

\paragraph{Quasi-monotonicity.}
The augmented walk evolves under a state-dependent projective rescaling rather than a fixed Markov kernel, so the data-processing inequality does not apply directly.
We instead prove a relaxed monotonicity in which $\mathcal{D}_k^{\mathrm{aug}}$ is non-increasing up to a correction that decays geometrically at a rate controlled by the normalized spectrum of $\mathbf{M}_{\varepsilon}$.

\begin{proposition}[Quasi-monotonicity of $\mathcal{D}_k^{\mathrm{aug}}$]
\label{prop:dk-aug-quasi-monotone}
Let $\mathbf{M}_{\varepsilon}=(1+\varepsilon)\mathbf{I}+\mathbf{A}$ be the GIN aggregation operator on a connected undirected graph with $\varepsilon\geq0$,
let $\lambda_1>|\lambda_2|\geq\cdots\geq|\lambda_n|$ denote its eigenvalues, and set
\[
    \kappa:=\frac{|\lambda_2|}{\lambda_1}\in[0,1).
\]
Using $0^0=1$ when $\kappa=0$, there exists a constant $C>0$ depending only on the graph such that
\begin{equation}
\label{eq:quasi-monotone}
    \mathcal{D}_{k+1}^{\mathrm{aug}}
    \leq \mathcal{D}_k^{\mathrm{aug}}+C\kappa^{k/2}
    \qquad\text{for all }k\geq0.
\end{equation}
Moreover, there exist $k_0\in\mathbb{N}$ and $C'>0$ such that the rate sharpens to
$\mathcal{D}_{k+1}^{\mathrm{aug}}
\leq\mathcal{D}_k^{\mathrm{aug}}+C'\kappa^k$
for all $k\geq k_0$.
\end{proposition}

\begin{proof}
Write $w_v:=(\mathbf{v}_1)_v$ for the entries of the unit-$\ell_2$ Perron eigenvector and let $K:=\mathbf{M}_{\varepsilon}/\lambda_1$.
Connectedness and $\varepsilon\geq0$ imply $w_v>0$ for every $v$ and
$\kappa=|\lambda_2|/\lambda_1<1$ (Perron--Frobenius).

\smallskip\noindent
\emph{Step 1 (convergence of the augmented walks).}
Symmetry of $\mathbf{M}_{\varepsilon}$ gives the orthonormal eigendecomposition
\[
    K^k e_v
    = w_v\mathbf{v}_1+\mathbf{r}_v^{(k)},
    \qquad
    \|\mathbf{r}_v^{(k)}\|_2\leq\kappa^k,
\]
and consequently $\|\mathbf{r}_v^{(k)}\|_1\leq\sqrt{n}\,\kappa^k$ by Cauchy--Schwarz.
Set $\tilde{p}_v^{(k)}:=K^k e_v$; the total mass
\[
    m_v^{(k)}
    :=\|\tilde{p}_v^{(k)}\|_1
    =m_v^{(\infty)}+e_k,
    \qquad
    m_v^{(\infty)}
    :=w_v\|\mathbf{v}_1\|_1>0,
\]
where $|e_k|\leq\|\mathbf{r}_v^{(k)}\|_1\leq\sqrt{n}\,\kappa^k$.
The augmented distribution is
\[
    p_v^{(k),\mathrm{aug}}
    =
    \frac{1}{m_v^{(k)}+1}
    \begin{pmatrix}
        \tilde{p}_v^{(k)}\\
        1
    \end{pmatrix},
    \qquad
    p_v^{(\infty),\mathrm{aug}}
    =
    \frac{1}{m_v^{(\infty)}+1}
    \begin{pmatrix}
        \tilde{p}_v^{(\infty)}\\
        1
    \end{pmatrix},
\]
with $\tilde{p}_v^{(\infty)}:=w_v\mathbf{v}_1$ and
$m_v^{(\infty)}=\|\tilde{p}_v^{(\infty)}\|_1$.

We bound the $\ell_1$ distance between these two $(n+1)$-vectors.
For the first $n$ coordinates:
\[
\begin{aligned}
\left\|
    \frac{\tilde{p}_v^{(k)}}{m_v^{(k)}+1}
    -\frac{\tilde{p}_v^{(\infty)}}{m_v^{(\infty)}+1}
\right\|_1
&\leq
\left|
    \frac{1}{m_v^{(k)}+1}
    -\frac{1}{m_v^{(\infty)}+1}
\right|
\|\tilde{p}_v^{(k)}\|_1\\
&\quad+
\frac{\|\tilde{p}_v^{(k)}-\tilde{p}_v^{(\infty)}\|_1}
     {m_v^{(\infty)}+1}.
\end{aligned}
\]

The scalar inequality
\[
    |(a+1)^{-1}-(b+1)^{-1}|
    =\frac{|a-b|}{(a+1)(b+1)}
    \leq|a-b|
    \qquad(a,b\geq0)
\]
yields
\[
    \left|
        \frac{1}{m_v^{(k)}+1}
        -\frac{1}{m_v^{(\infty)}+1}
    \right|
    \leq|e_k|
    \leq\sqrt{n}\,\kappa^k.
\]
Because $\tilde{p}_v^{(k)}$ has non-negative entries,
$\|\tilde{p}_v^{(k)}\|_1=m_v^{(k)}$ is uniformly bounded by some constant $M$ independent of $v$ and $k$ (one can take $M=\sqrt{n}$, since $\|K^k\|_{\mathrm{op}}\leq1$).
Hence the first part is at most $M\sqrt{n}\,\kappa^k$.

The second part is simply
$\|\mathbf{r}_v^{(k)}\|_1/(m_v^{(\infty)}+1)
\leq\sqrt{n}(m_v^{(\infty)}+1)^{-1}\kappa^k$.

For the appended coordinate,
\[
    \left|
        \frac{1}{m_v^{(k)}+1}
        -\frac{1}{m_v^{(\infty)}+1}
    \right|
    \leq\sqrt{n}\,\kappa^k.
\]

Adding the three contributions we obtain
\[
    \|p_v^{(k),\mathrm{aug}}
      -p_v^{(\infty),\mathrm{aug}}\|_1
    \leq C_1\kappa^k,
\]
with the graph-dependent constant
\[
    C_1
    :=\sqrt{n}\left(
        M+\frac{1}{\min_u(m_u^{(\infty)}+1)}+1
    \right).
\]

\smallskip\noindent
\emph{Step 2 (from $\ell_1$ convergence to dispersion).}
For any $p,q\in\Delta^n$, the elementary inequality
$(\sqrt{a}-\sqrt{b})^2\leq|a-b|$ summed over coordinates gives
$\|\sqrt{p}-\sqrt{q}\|_2\leq\sqrt{\|p-q\|_1}$.
Write $\mathrm{BC}(p,q)=\langle\sqrt{p},\sqrt{q}\rangle=1-H^2(p,q)$
for the Bhattacharyya coefficient.
By Cauchy--Schwarz and $\|\sqrt{p}\|_2=1$,
\[
\begin{aligned}
    |\mathrm{BC}(p,q)-\mathrm{BC}(p',q')|
    &\leq
    \|\sqrt{p}-\sqrt{p'}\|_2
    +\|\sqrt{q}-\sqrt{q'}\|_2\\
    &\leq
    \sqrt{\|p-p'\|_1}
    +\sqrt{\|q-q'\|_1}.
\end{aligned}
\]
Applying this to the identity
$\mathcal{D}_k^{\mathrm{aug}}
=1-\frac{1}{n^2}\sum_{u,v}
\mathrm{BC}(p_u^{(k),\mathrm{aug}},p_v^{(k),\mathrm{aug}})$,
which is \eqref{eq:dk-aug} with the squared norm expanded, and inserting the bound from Step~1,
\[
    |\mathcal{D}_k^{\mathrm{aug}}
      -\mathcal{D}_{\infty}^{\mathrm{aug}}|
    \leq2\sqrt{C_1}\,\kappa^{k/2},
\]
where
$\mathcal{D}_{\infty}^{\mathrm{aug}}
:=1-\|\frac{1}{n}\sum_v
\sqrt{p_v^{(\infty),\mathrm{aug}}}\|_2^2$.

\smallskip\noindent
\emph{Step 3 (triangle inequality).}
From Step~2,
\[
\begin{aligned}
    \mathcal{D}_{k+1}^{\mathrm{aug}}
    -\mathcal{D}_k^{\mathrm{aug}}
    &=
    \bigl(\mathcal{D}_{k+1}^{\mathrm{aug}}
          -\mathcal{D}_{\infty}^{\mathrm{aug}}\bigr)
    -
    \bigl(\mathcal{D}_k^{\mathrm{aug}}
          -\mathcal{D}_{\infty}^{\mathrm{aug}}\bigr)\\
    &\leq4\sqrt{C_1}\,\kappa^{k/2},
\end{aligned}
\]
which gives
$\mathcal{D}_{k+1}^{\mathrm{aug}}
\leq\mathcal{D}_k^{\mathrm{aug}}+C\kappa^{k/2}$
with $C:=4\sqrt{C_1}$.
This establishes~\eqref{eq:quasi-monotone}.

\smallskip\noindent
The factor $\tfrac12$ in the exponent of Step~2 originates solely from the non-Lipschitz behaviour of $x\mapsto\sqrt{x}$ near zero.
Because $w_v>0$ for every vertex, the limit vector
$p_v^{(\infty),\mathrm{aug}}$ has all coordinates bounded below by some $\delta_\star>0$.
By Step~1 there exists $k_0$ such that for all $k\geq k_0$ every coordinate of
$p_v^{(k),\mathrm{aug}}$ exceeds $\delta_\star/2$.
On the interval $[\delta_\star/2,1]$ the square root is
$(2\sqrt{\delta_\star/2})^{-1}$-Lipschitz, so the bound in Step~2 sharpens to
\[
    |\mathcal{D}_k^{\mathrm{aug}}
      -\mathcal{D}_{\infty}^{\mathrm{aug}}|
    \leq C''\kappa^k,
    \qquad k\geq k_0,
\]
for a suitable $C''$.
Repeating the triangle inequality of the previous step gives
$\mathcal{D}_{k+1}^{\mathrm{aug}}
\leq\mathcal{D}_k^{\mathrm{aug}}+C'\kappa^k$
with $C'=2C''$ for all $k\geq k_0$.
\end{proof}

\paragraph{Empirical validation on GIN.}
We evaluate $\mathcal{D}_k^{\mathrm{aug}}$ on the GIN configurations of \S\ref{app:gin-results}; the full per-$\varepsilon$ results are in Table~\ref{tab:gin-results}.
The augmented variant has the highest observed mean absolute correlation at every $\varepsilon$ setting and Overall, with a $+0.158$ improvement over the standard $\mathcal{D}_k$ and $+0.033$ over the strongest post-training baseline (Erank).
The improvement over the standard $\mathcal{D}_k$ is consistent with the hypothesis that retaining propagation magnitudes provides useful information for GIN (\S\ref{app:gen-sum}).

\subsection{Ablation: choice of propagation operator}
\label{app:gen-ablation}

Theorem~\ref{thm:lazy-universality} shows that, for operators satisfying its conditions, the asymptotic decay rate of their associated $\mathcal{D}_k$ is comparable to that of the lazy walk. We now test whether this translates to similar observed correlations: does the choice of $\mathbf{P}$ underlying $\mathcal{D}_k$ affect its predictive correlation with accuracy degradation?
We vary the laziness coefficient and compare row-stochastic propagation with row-normalized powers of its symmetric counterpart.

\paragraph{Axis 1: laziness coefficient.}
We sweep the self-loop weight in
$\mathbf{P}_{\alpha}
=\alpha\mathbf{I}+(1-\alpha)\mathbf{D}^{-1}\mathbf{A}$
for $\alpha\in\{0,0.25,0.5,0.75\}$.
The canonical lazy walk corresponds to $\alpha=0.5$, the smallest laziness that guarantees a non-negative spectrum on every graph.
The case $\alpha=0$ is the non-lazy walk $\mathbf{D}^{-1}\mathbf{A}$, whose spectrum can be negative (it is periodic on bipartite graphs), so convergence to zero requires a connected non-bipartite graph, with the decay rate governed by $r_\star=\max_{i\geq2}|\lambda_i|$ rather than necessarily $\lambda_2$.

\begin{table}[H]
\centering
\caption{Laziness coefficient $\alpha$ in $\mathbf{P}_{\alpha}=\alpha\mathbf{I}+(1-\alpha)\mathbf{D}^{-1}\mathbf{A}$.
Correlations are computed across eight depths, then averaged over 14 model configurations and the datasets in each row. Mean $|r|$ is non-increasing in $\alpha$ in both regimes, with $\alpha=0$ leading the canonical $\alpha=0.5$ by $+0.013$ Overall. Bold marks the per-row best.}
\label{tab:ablation-alpha}
\renewcommand{\arraystretch}{1.10}
\begin{tabular}{lcccc}
\toprule
$\alpha$ & 0 & 0.25 & 0.5 (canonical) & 0.75 \\
\midrule
Homophilic   & \textbf{0.788} & 0.780 & 0.762 & 0.732 \\
Heterophilic & \textbf{0.702} & 0.702 & 0.698 & 0.687 \\
\midrule
Overall      & \textbf{0.734} & 0.730 & 0.721 & 0.703 \\
\bottomrule
\end{tabular}
\end{table}

The per-dataset breakdown (Table~\ref{tab:ablation-alpha-perdataset}) shows that the trend holds on most datasets: 7 of 11 datasets are best at $\alpha=0$, the per-dataset range is at most $0.093$ (ogbn-arxiv) and typically below $0.05$, and Chameleon is the sole dataset where $\alpha=0.5$ (lazy walk) wins.

\begin{table}[H]
\centering
\caption{Per-dataset breakdown of the laziness sweep. Each cell is the mean $|r|$ computed across eight depths and averaged over 14 configurations. The Range column reports the gap between best and worst $\alpha$.}
\label{tab:ablation-alpha-perdataset}
\renewcommand{\arraystretch}{1.05}
\small
\begin{tabular}{lccccc}
\toprule
Dataset & $\alpha=0$ & $\alpha=0.25$ & $\alpha=0.5$ & $\alpha=0.75$ & Range \\
\midrule
\multicolumn{6}{l}{\textit{Homophilic}} \\
CiteSeer      & \textbf{0.859} & 0.853 & 0.842 & 0.826 & 0.033 \\
Cora          & \textbf{0.773} & 0.763 & 0.744 & 0.718 & 0.055 \\
PubMed        & \textbf{0.703} & 0.696 & 0.681 & 0.657 & 0.046 \\
ogbn-arxiv    & \textbf{0.819} & 0.808 & 0.781 & 0.726 & 0.093 \\
\midrule
\multicolumn{6}{l}{\textit{Heterophilic}} \\
AmazonRatings & \textbf{0.759} & 0.751 & 0.743 & 0.737 & 0.022 \\
Chameleon     & 0.872 & 0.874 & \textbf{0.878} & 0.868 & 0.010 \\
Cornell       & 0.533 & \textbf{0.543} & 0.542 & 0.540 & 0.010 \\
RomanEmpire   & \textbf{0.840} & 0.833 & 0.829 & 0.819 & 0.021 \\
Squirrel      & \textbf{0.732} & 0.730 & 0.723 & 0.698 & 0.034 \\
Texas         & 0.528 & \textbf{0.530} & 0.521 & 0.512 & 0.018 \\
Wisconsin     & 0.652 & \textbf{0.655} & 0.649 & 0.636 & 0.020 \\
\bottomrule
\end{tabular}
\end{table}
In the random-walk interpretation, each layer keeps the walk at its current node with probability $\alpha$ and moves it to a neighbor with probability $1-\alpha$. Thus, over $k$ layers, the expected number of neighbor moves is $k(1-\alpha)$. For example, at $\alpha=0.5$, sixteen layers yield eight neighbor moves on average, but are not equivalent to eight non-lazy propagation steps: the number of moves varies across walks. Across the eight tested depths between $2$ and $24$, smaller $\alpha$ yielded higher mean correlations in both dataset regimes. The greater expected number of neighbor moves offers intuition for this empirical trend, but does not establish it theoretically.

\paragraph{Axis 2: symmetric vs.\ row-stochastic.}
The row-stochastic operator
$\mathbf{T}=\tfrac{1}{2}(\mathbf{I}+\mathbf{D}^{-1}\mathbf{A})$
and the symmetric operator
$\mathbf{S}=\tfrac{1}{2}(\mathbf{I}+\mathbf{D}^{-1/2}\mathbf{A}\mathbf{D}^{-1/2})$
are related by the diagonal similarity
$\mathbf{S}=\mathbf{D}^{1/2}\mathbf{T}\mathbf{D}^{-1/2}$. They share the same spectrum and therefore the same spectral convergence rate, but not the same dispersion values (Remark~\ref{rem:gcn-bridge}).
We test how close the two are in practice.

\textbf{A technical note:}
the centroid computation of $\mathcal{D}_k$ (Equation~\ref{eq:dispersion_def}, Lemma~\ref{lem:equivalence}) requires probability rows, and $\mathbf{S}^k$ is not generally row-stochastic:
$[\mathbf{S}^k]_{vw}
=\sqrt{d_v/d_w}\,[\mathbf{T}^k]_{vw}$.
We therefore normalize each row of $\mathbf{S}^k$ to unit $\ell_1$ norm before applying the centroid formula.
This removes the row factor $\sqrt{d_v}$ but not the column weights $d_w^{-1/2}$, so the normalized rows are degree-reweighted versions of the walk distributions $p_v^{(k)}$ rather than the walk distributions themselves, and the comparison below is an empirical one.

\begin{table}[H]
\centering
\caption{Symmetric vs.\ row-stochastic propagation at $\alpha=0.5$. Mean absolute correlations differ by $0.003$ Overall and at most $0.005$ across the two dataset regimes.}
\label{tab:ablation-sym}
\renewcommand{\arraystretch}{1.10}
\begin{tabular}{lcc}
\toprule
Dataset regime & Row-stochastic $\mathbf{T}$ & Symmetric $\mathbf{S}$ \\
\midrule
Homophilic   & \textbf{0.763} & 0.758 \\
Heterophilic & \textbf{0.697} & 0.696 \\
\midrule
Overall      & \textbf{0.721} & 0.718 \\
\bottomrule
\end{tabular}
\end{table}

The row-stochastic form has the higher mean on 9 of 11 datasets, but the gaps are uniformly below $0.01$ and the Overall difference is $0.003$. The two constructions therefore have similar observed correlations, although this does not establish statistical equivalence.
We use the row-stochastic form throughout the main paper because it admits the direct probabilistic interpretation of walk distributions on $\Delta^{n-1}$ that motivates the construction.

\paragraph{Implications.}
Two findings emerge.
First, the diagnostic has similar Overall correlations across the tested constructions: across the laziness sweep ($\alpha=0$ to $\alpha=0.75$, a range of $0.031$ Overall) and the symmetric/row-stochastic comparison ($0.003$ Overall), no choice has the highest mean on every dataset, and every variant has a higher Overall mean than every post-training baseline of Table~\ref{tab:ranking}.
Second, smaller laziness has a slightly higher observed Overall correlation, with $\alpha=0$ leading $\alpha=0.5$ by $+0.013$ Overall.
We retain $\alpha=0.5$ as the canonical choice in the main paper for two reasons: it is the smallest laziness for which the spectrum is non-negative on every graph, so that Theorem~\ref{thm:decay} holds without a non-bipartiteness assumption, and it is the standard convention in the Markov-chain literature~\citep{levin2017markov}. Among the tested values $\alpha\in\{0,0.25,0.5\}$, Overall correlations differ by at most $0.013$. For $0<\alpha<1/2$, negative eigenvalues are possible, but convergence still holds on connected bipartite graphs; the rate is governed by $\max_{i\geq2}|\lambda_i|$. Non-bipartiteness is needed only at $\alpha=0$.

\section{Extended empirical results}
\label{app:extended-results}

This appendix collects the empirical results that support and qualify the headline numbers of \S\ref{sec:experiments}, organised by the question each experiment answers. \S\ref{app:headline-robustness} asks how robust the headline comparison of Table~\ref{tab:ranking} is: statistical significance, choice of correlation measure, activation function, and per-dataset behaviour. \S\ref{app:depth-controlled} asks whether $\mathcal{D}_k$ is more than a proxy for depth. \S\ref{app:arch-matched} recomputes the diagnostic from operators associated with each architecture, including GIN. \S\ref{app:mitigations} evaluates it under oversmoothing mitigation techniques. \S\ref{app:practical-use} examines how far the diagnostic can be put to practical use: depth selection, detection, and regularization. Unless stated otherwise, the planned benchmark design is that of Table~\ref{tab:ranking} ($11$ datasets $\times$ $14$ configurations $\times$ $8$ depths $\times$ $10$ seeds).

\subsection{Robustness of the headline comparison}
\label{app:headline-robustness}

\subsubsection{Statistical significance}
\label{app:significance}

Table~\ref{tab:ranking} reports means of per-trajectory correlations. The 14 model configurations evaluated on one dataset share the same graph and hence the same $\mathcal{D}_k$ trajectory, so they are not independent observations of the diagnostic. Following \citet{demsar2006statistical}, we therefore take the \emph{dataset} as the unit of analysis: for each diagnostic we first average the $14$ configuration-level $|r|$ values within each dataset, which yields $N = 11$ paired observations per diagnostic, and run all tests in this subsection on these.

\paragraph{Friedman test with Holm-corrected post-hoc comparisons.}
A Friedman test over the eight diagnostics rejects equal performance ($\chi^2(7) = 14.94$, $p = 0.037$). Holm-corrected post-hoc comparisons of mean ranks with $\mathcal{D}_k$ as the control (Table~\ref{tab:friedman-holm}) show that $\mathcal{D}_k$ has the best average rank ($2.64$ of $8$) and is significantly better than MAD ($p = 0.012$) and $E_{\mathrm{Proj}}$ ($p = 0.024$); the remaining comparisons, including the one against Erank, are not significant.

\begin{table}[H]
\centering
\small
\caption{Friedman test ($\chi^2(7)=14.94$, $p=0.037$, $N=11$ datasets) and Holm-corrected post-hoc comparisons against $\mathcal{D}_k$. Lower average rank is better. ${}^{*}$ significant at $\alpha = 0.05$; n.s.: not significant.}
\label{tab:friedman-holm}
\renewcommand{\arraystretch}{1.1}
\begin{tabular}{lcc}
\toprule
Diagnostic & Average rank & Holm $p$ vs.\ $\mathcal{D}_k$ \\
\midrule
$\mathcal{D}_k$ (ours) & \textbf{2.64} & -- \\
Erank                  & 3.55 & n.s. \\
$E_{\mathrm{Proj,n}}$  & 4.18 & n.s. \\
NumRank                & 4.27 & n.s. \\
$E_{\mathrm{Dir,n}}$   & 4.91 & n.s. \\
$E_{\mathrm{Dir}}$     & 4.91 & n.s. \\
$E_{\mathrm{Proj}}$    & 5.64 & $0.024^{*}$ \\
MAD                    & 5.91 & $0.012^{*}$ \\
\bottomrule
\end{tabular}
\end{table}

\paragraph{Pairwise Wilcoxon signed-rank tests.}
Table~\ref{tab:wilcoxon-pairwise} reports exact two-sided paired Wilcoxon signed-rank tests of $\mathcal{D}_k$ against each baseline over the $11$ dataset-level values, with Holm correction over the seven comparisons. Against Erank, $\mathcal{D}_k$ is higher on $8$ of $11$ datasets with an unrounded mean difference of $+0.0336$ ($0.72117$ vs.\ $0.68759$), but the test does not reject the null ($p = 0.1016$ raw, $p = 0.3047$ after Holm correction). 

\begin{table}[H]
\centering
\small
\caption{Pairwise exact Wilcoxon signed-rank tests of $\mathcal{D}_k$ against each baseline over the $11$ dataset-level mean correlations (two-sided; Holm correction over seven comparisons). ``Wins'' counts datasets on which $\mathcal{D}_k$ has the higher mean. $^*p<0.05$; $^{**}p<0.01$.}
\label{tab:wilcoxon-pairwise}
\renewcommand{\arraystretch}{1.1}
\begin{tabular}{lccccccc}
\toprule
 & Erank & NumRank & $E_{\mathrm{Dir}}$ & $E_{\mathrm{Dir,n}}$ & $E_{\mathrm{Proj}}$ & $E_{\mathrm{Proj,n}}$ & MAD \\
\midrule
Dataset mean        & 0.688 & 0.647 & 0.629 & 0.647 & 0.614 & 0.674 & 0.617 \\
$\mathcal{D}_k$ wins & 8/11 & 6/11 & 9/11 & 8/11 & 9/11 & 9/11 & 10/11 \\
$W$                 & 14    & 18    & 8     & 10    & 8     & 15    & 1 \\
$p$ (raw)           & 0.102 & 0.206 & $0.024^{*}$ & $0.042^{*}$ & $0.024^{*}$ & 0.123 & $0.002^{**}$ \\
$p$ (Holm)          & 0.305 & 0.305 & 0.146 & 0.168 & 0.146 & 0.305 & $0.014^{*}$ \\
\bottomrule
\end{tabular}
\end{table}

\paragraph{Hierarchical bootstrap.}
We estimate uncertainty in the mean difference between $\mathcal{D}_k$ and Erank using $10{,}000$ hierarchical bootstrap replicates (seed $42$). Each replicate resamples the $11$ datasets with replacement, then resamples the $14$ configurations within each selected dataset. We compute both diagnostics' overall means from the same resampled cells and record their difference. The $2.5$th and $97.5$th percentiles give a $95\%$ interval of $[-0.010,\,0.079]$.

$\mathcal{D}_k$ has the highest observed mean among the eight diagnostics. The Friedman post-hoc comparisons identify differences against MAD and $E_{\mathrm{Proj}}$; the Holm-corrected paired Wilcoxon tests identify a difference only against MAD. Neither procedure establishes superiority or equivalence to Erank. Our main finding is that $\mathcal{D}_k$ achieves similar observed correlations to the strongest post-training diagnostics at substantially lower cost.

\subsubsection{Choice of correlation measure}
\label{app:rank-correlations}

The headline numbers in \S\ref{sec:experiments} use the absolute Pearson correlation $|r|$ between $\log(\mathrm{metric})$ and test accuracy. To check the leading comparison under other association measures, Table~\ref{tab:rank-correlations} repeats the comparison under Spearman and Kendall rank correlations.

\begin{table}[H]
\centering
\caption{Mean absolute correlation between each diagnostic and test accuracy, computed across eight depths and averaged over all (dataset, model) pairs. $\mathcal{D}_k$ has the highest observed mean under every correlation type. The Pearson row uses the aggregates in Table~\ref{tab:ranking}. Bold marks the best value per row; underline marks the second best.}
\label{tab:rank-correlations}
\renewcommand{\arraystretch}{1}
\small
\begin{tabular}{lcccccccc}
\toprule
Correlation       & $\mathcal{D}_k$ & Erank & EProj,n & MAD & NumRank & EDir,n & EDir & EProj \\
\midrule
Pearson (log) & \textbf{0.721} & \underline{0.688} & 0.674 & 0.617 & 0.647 & 0.647 & 0.629 & 0.614 \\
Spearman          & \textbf{0.7515} & \underline{0.6810} & 0.5727 & 0.6573 & 0.6628 & 0.6106 & 0.5438 & 0.5276 \\
Kendall           & \textbf{0.6731} & \underline{0.5918} & 0.4691 & 0.5621 & 0.5651 & 0.5207 & 0.4519 & 0.4303 \\
\bottomrule
\end{tabular}
\end{table}

The two leading diagnostics are unchanged: $\mathcal{D}_k$ has the highest mean under all three measures and Erank is consistently second, with a gap that is larger under the rank correlations ($+0.071$ under Spearman, $+0.081$ under Kendall) than under Pearson ($+0.034$). The remaining diagnostics change order. These are differences in observed means; \S\ref{app:significance} tests the Pearson comparison only. When $\mathcal{D}_k$ is strictly decreasing over the evaluated depths, its absolute Spearman and Kendall correlations equal those of depth itself. Thus these rank correlations demonstrate monotone association, rather than information beyond depth.

\subsubsection{Per-activation breakdown}
\label{app:activations}

Tables~\ref{tab:activations-lrelu} and~\ref{tab:activations-tanh} report the mean absolute correlation per (architecture, activation) cell, computed across eight depths and averaged over the 11 datasets.

\begin{table}[H]
\centering
\caption{LeakyReLU: Mean absolute correlation $|r|$ between each diagnostic and GNN accuracy, computed across eight depths and averaged over datasets. $\mathcal{D}_k$ ties EProj,n at the reported precision at the activation subtotal (both 0.696). Bold = best in row; underline = second best.}
\label{tab:activations-lrelu}
\renewcommand{\arraystretch}{1.10}
\small
\begin{tabular}{lcccccccc}
\toprule
Architecture & $\mathcal{D}_k$ & Erank & NumRank & EDir & EDir,n & EProj & EProj,n & MAD \\
\midrule
GCN     & \underline{0.756} & 0.657 & 0.653 & 0.559 & \textbf{0.808} & 0.543 & 0.560 & 0.614 \\
GATv1   & \underline{0.764} & 0.653 & 0.508 & 0.700 & \textbf{0.773} & 0.729 & 0.751 & 0.547 \\
GATv2   & 0.784 & 0.743 & 0.611 & 0.745 & \textbf{0.844} & 0.769 & \underline{0.816} & 0.619 \\
SAGE    & 0.777 & \textbf{0.868} & 0.687 & 0.735 & 0.702 & 0.762 & \underline{0.807} & 0.712 \\
ChebNet & \underline{0.781} & \textbf{0.849} & 0.752 & 0.666 & 0.544 & 0.704 & 0.715 & 0.661 \\
JKNet   & 0.407 & 0.397 & 0.445 & 0.340 & \textbf{0.499} & 0.444 & \underline{0.468} & 0.415 \\
APPNP   & 0.604 & 0.501 & 0.620 & \underline{0.645} & 0.556 & 0.623 & \textbf{0.754} & 0.619 \\
\midrule
\textit{LReLU avg.} & \textbf{0.696} & 0.667 & 0.611 & 0.627 & \underline{0.675} & 0.653 & \textbf{0.696} & 0.598 \\
\bottomrule
\end{tabular}
\end{table}

\begin{table}[H]
\centering
\caption{Tanh: Mean absolute correlation $|r|$ between each diagnostic and GNN accuracy, computed across eight depths and averaged over datasets. $\mathcal{D}_k$ leads the activation subtotal (0.747 vs.\ Erank's 0.709), having the highest observed mean on four of seven architectures (GCN, GATv1, GATv2, ChebNet). Bold = best in row; underline = second best.}
\label{tab:activations-tanh}
\renewcommand{\arraystretch}{1.10}
\small
\begin{tabular}{lcccccccc}
\toprule
Architecture & $\mathcal{D}_k$ & Erank & NumRank & EDir & EDir,n & EProj & EProj,n & MAD \\
\midrule
GCN     & \textbf{0.830} & \underline{0.753} & 0.689 & 0.548 & 0.631 & 0.449 & 0.529 & 0.554 \\
GATv1   & \textbf{0.805} & \underline{0.722} & 0.658 & 0.565 & 0.657 & 0.488 & 0.648 & 0.698 \\
GATv2   & \textbf{0.813} & \underline{0.785} & 0.726 & 0.573 & 0.711 & 0.593 & 0.753 & 0.722 \\
SAGE    & \underline{0.847} & \textbf{0.901} & 0.721 & 0.743 & 0.802 & 0.705 & 0.812 & 0.810 \\
ChebNet & \textbf{0.830} & 0.720 & \underline{0.729} & 0.717 & 0.403 & 0.590 & 0.579 & 0.470 \\
JKNet   & 0.600 & 0.582 & \textbf{0.686} & 0.608 & 0.622 & 0.532 & 0.579 & \underline{0.645} \\
APPNP   & 0.501 & 0.497 & 0.575 & \underline{0.661} & 0.506 & \textbf{0.670} & \underline{0.661} & 0.552 \\
\midrule
\textit{Tanh avg.} & \textbf{0.747} & \underline{0.709} & 0.683 & 0.631 & 0.619 & 0.575 & 0.652 & 0.636 \\
\midrule
\textbf{Overall} & \textbf{0.721} & \underline{0.688} & 0.647 & 0.629 & 0.647 & 0.614 & 0.674 & 0.617 \\
\bottomrule
\end{tabular}
\end{table}

The activation breakdown shows that $\mathcal{D}_k$ achieves its strongest advantage under Tanh, where it attains the highest activation-level average (0.747), compared to Erank (0.709). In this regime, it is the best-performing diagnostic on four of seven architectures (GCN, GATv1, GATv2, ChebNet) and second on SAGE. Under LeakyReLU, the leading activation-level means are closer. At the activation-level average, $\mathcal{D}_k$ and EProj,n are tied (both 0.696).

Across both activation regimes, several diagnostics have lower mean correlations on JKNet and APPNP, consistent with what we discussed in \S\ref{ssec:per_arch}.

\subsubsection{Per-dataset dispersion-vs-depth curves}
\label{app:scatter-plots}

\begin{figure}[H]
\centering
 \includegraphics[width=\textwidth]{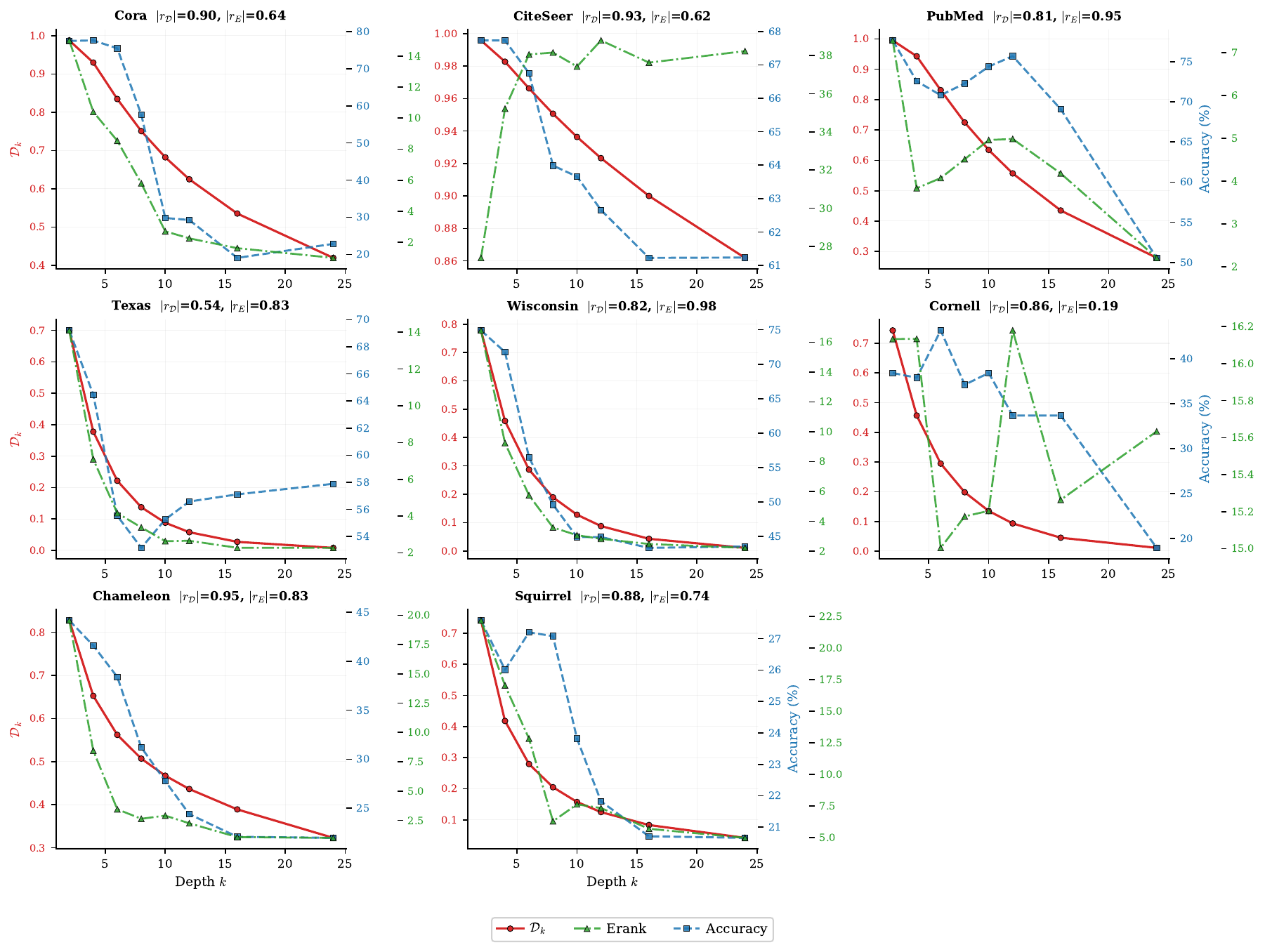}
\caption{
\textbf{Dispersion, effective rank, and accuracy vs.\ depth.}
For each dataset we plot $\mathcal{D}_k$ (red, left), $\mathrm{Erank}(X^{(k)})$
(green, offset right), and test accuracy (blue dashed, right) against depth $k\in\{2,4,6,8,10,12,16,24\}$, using $P=D^{-1}A$.}
\label{fig:scatter-perdataset}
\end{figure}

Figure~\ref{fig:scatter-perdataset} shows, for each of the eight homophilic and heterophilic benchmarks shown, how the proposed dispersion metric $\mathcal{D}_k$ (red, left axis), the effective rank $\mathrm{Erank}(X^{(k)})$ (green, offset right axis), and the test accuracy of the trained model (blue dashed, right axis) jointly evolve as a function of the propagation depth~$k\in\{2,4,6,8,10,12,16,24\}$. The $\mathcal{D}_k$ curves in this figure are computed on the non-lazy walk $P=D^{-1}A$, the $\alpha=0$ operator of the laziness ablation in Appendix~\ref{app:gen-ablation}; accuracy and Erank are computed from the trained models.

\paragraph{Configuration selection.}
For each dataset we sample one (architecture, activation) combination uniformly at random from the 14 model$\times$activation pairs described in Section~\ref{sec:experiments}, to avoid selecting panels by diagnostic performance. Both $\mathcal{D}_k$ and $\mathrm{Erank}$ are plotted on their natural (raw) scale, and the per-panel Pearson correlations $|r_D|$ and $|r_E|$ between $\log$-metric and accuracy are reported in each title.

$\mathcal{D}_k$  decreases across the plotted depths, although several panels remain far from zero and accuracy need not decrease monotonically. Its per-panel Pearson correlation with accuracy ranges from $0.54$ (\textsc{Texas}) to $0.95$ (\textsc{Chameleon}). The $\mathrm{Erank}$ curve is markedly less consistent: it tracks accuracy almost perfectly on \textsc{Wisconsin} ($|r_E|\!=\!0.98$) and \textsc{PubMed} ($0.95$), but collapses to $|r_E|\!=\!0.19$ on \textsc{Cornell} and stays below $0.65$ on \textsc{Cora} and \textsc{CiteSeer}. Aggregated over the eight panels, $\mathcal{D}_k$ wins on five (\textsc{Cora}, \textsc{CiteSeer}, \textsc{Cornell}, \textsc{Chameleon}, \textsc{Squirrel}) and Erank on three (\textsc{PubMed}, \textsc{Texas}, \textsc{Wisconsin}), illustrating variation in their relative performance across datasets.

The per-panel picture illustrates that neither diagnostic dominates on every selected trajectory. These panels use the non-lazy operator, whereas Table~\ref{tab:ranking} uses the canonical lazy walk. Both versions of $\mathcal{D}_k$ require no trained representations; computational costs are reported in Appendix~\ref{app:estimator-runtime}.

\subsection{Depth dependence and operator-dependent mixing}
\label{app:depth-controlled}

At a given depth, $\mathcal{D}_k$ measures the remaining
diversity of walk distributions for a particular graph
and propagation operator. Different graphs or operators
can therefore produce different dispersion values at the
same depth. We examine whether this structural information
helps explain accuracy beyond specified depth trends.

Within-graph correlation alone cannot answer this question.
For example, if we assume that the dispersion decays exactly exponentially,
$\log\mathcal{D}_k$ is linear in depth and has the same
absolute Pearson correlation with accuracy as depth itself,
regardless of the decay rate. We therefore use complementary
controls: regression relative to specified linear depth
models (\S\ref{app:added_value}), comparison with structural
baselines (\S\ref{app:structural-baselines}), and experiments
that hold depth fixed while varying the operator or graph
(\S\ref{app:teleport-sweep}--\S\ref{app:fixed-depth}).

\subsubsection{Held-out test with depth and dataset fixed effects}
\label{app:added_value}

\paragraph{Setup.}
After averaging accuracy over the $14$ model configurations of
Table~\ref{tab:ranking}, we obtain $N = 11 \times 8 = 88$ observations indexed
by $(\text{dataset}\;d, \text{depth}\;k)$.  For each diagnostic $M$ we fit
two ordinary-least-squares models,
\begin{align}
\textbf{Baseline:}\quad &\mathrm{acc}_{d,k} = \beta_0 + \beta_1 k + \alpha_d + \varepsilon_{d,k},
\label{eq:baseline}\\[2pt]
\textbf{Full}(M):\quad &\mathrm{acc}_{d,k} = \beta_0 + \beta_1 k + \alpha_d + \beta_M\log M_{d,k} + \varepsilon_{d,k},
\label{eq:full}
\end{align}
where $\{\alpha_d\}$ are dataset fixed effects with one reference dataset. The baseline absorbs the linear effect of depth and every per-dataset constant offset; any additional variance explained by $M$ in Eq.~\eqref{eq:full} is incremental over this specified linear model.

\paragraph{Evaluation.}
We measure whether adding $\log M$ improves the regression in two ways. First, we measure the improvement in fit on the data used to estimate the models:
$\Delta R^2_{\mathrm{in}}=R^2(\text{Full}(M))-R^2(\text{Baseline})$.
Second, we use $5$-fold cross-validation, with each dataset represented in every fold, to measure the change in prediction performance on held-out observations, $\Delta R^2_{\mathrm{cv}}$. This evaluates prediction at held-out depths within the observed datasets. A negative $\Delta R^2_{\mathrm{cv}}$ means that adding the diagnostic makes these predictions worse.

\paragraph{Result.}
The baseline alone attains $R^2_{\mathrm{in}} = 0.7804$ and
$R^2_{\mathrm{cv}} = 0.6285$. Adding $\log\mathcal{D}_k$ improves cross-validated $R^2$ by $0.099$. Table~\ref{tab:added_value} reports the results for all diagnostics.

\begin{table}[H]
\centering
\caption{Changes in $R^2$ from adding $\log M$ to a baseline containing linear depth and dataset fixed effects (Eq.~\eqref{eq:baseline}). $\Delta R^2_{\mathrm{in}}$ measures the improvement in fit on the observed data; $\Delta R^2_{\mathrm{cv}}$ measures the change in prediction performance on held-out depths within the observed datasets, using $5$-fold cross-validation. Bold marks the training-free diagnostic.}
\label{tab:added_value}
\setlength{\tabcolsep}{8pt}
\renewcommand{\arraystretch}{1.05}
\begin{tabular}{lcc}
\toprule
Diagnostic $M$ & $\Delta R^2_{\mathrm{in}}$ & $\Delta R^2_{\mathrm{cv}}$ \\
\midrule
$\mathcal{D}_k$ (ours) & $\mathbf{+0.0530}$ & $\mathbf{+0.0986}$ \\
Erank                 & $+0.0332$ & $+0.0608$ \\
MAD                   & $+0.0112$ & $+0.0233$ \\
NumRank               & $+0.0124$ & $-0.0014$ \\
$E_{\mathrm{Dir}}$    & $+0.0002$ & $-0.0084$ \\
$E_{\mathrm{Dir,n}}$  & $+0.0921$ & $+0.1666$ \\
$E_{\mathrm{Proj}}$   & $+0.0019$ & $-0.0071$ \\
$E_{\mathrm{Proj,n}}$ & $+0.0020$ & $-0.0037$ \\
\bottomrule
\end{tabular}
\end{table}

\paragraph{Dataset-specific depth slopes.}
The baseline in Eq.~\eqref{eq:baseline} shares one depth slope across datasets. Adding a diagnostic can therefore improve prediction by capturing dataset-specific decay rates. We also fit a baseline with dataset-specific intercepts and slopes, $\beta_{0,d}+\beta_{1,d}k$. The spectral control $S_k$ adds $\Delta R^2_{\mathrm{cv}}=0.113$ under the shared-slope baseline and $0.000$ under the dataset-specific baseline; $\log\mathcal{D}_k$ adds $0.008$ under the latter. A dataset-specific intercept and slope absorb any diagnostic satisfying $\log M_{d,k}=a_d+b_dk$, including an exactly exponential dispersion trajectory. The remaining gain therefore reflects predictive value in departures from these log-linear trends. A small incremental gain is compatible with the exponential mixing behavior established theoretically. These results evaluate held-out depths within the observed datasets.

\subsubsection{Training-free structural baselines}
\label{app:structural-baselines}

Commute times, resistance distances, and hitting times characterize graph mixing without training, but in their classical form they are depth-independent: they assign one value per graph or node pair, so their correlation with a depth--accuracy curve is undefined. We therefore evaluate depth-dependent variants, the $k$-truncated hitting time and the $k$-truncated commute time (resistance distance is a rescaling of commute time), extracting at each depth the mean and the variance over node pairs and correlating these statistics with test accuracy across depths. We also include the spectral control $S_k$ defined above, using $\log S_k$. Untruncated hitting and resistance quantities are infinite between components, and $\gamma = 0$ on a disconnected graph makes $S_k \equiv 1$, so these baselines are computed on the largest connected component ($63.7\%$ of CiteSeer, $91.8\%$ of Cora; all other graphs are connected), whereas $\mathcal{D}_k$ is computed directly from the walk distributions on the full graphs. A sensitivity check without CiteSeer and Cora ($N = 9$) preserves the qualitative comparison.

\begin{table}[H]
\centering
\small
\caption{Training-free structural baselines versus $\mathcal{D}_k$, mean $|r|$ over the $11$ datasets. Hitting and commute statistics are untransformed; $\mathcal{D}_k$ and $S_k$ are log-transformed. The mean truncated commute time is twice the mean truncated hitting time, giving identical absolute Pearson correlations. Both are approximately linear in $k$ over the tested depths.}
\label{tab:structural}
\renewcommand{\arraystretch}{1.1}
\begin{tabular}{lcc}
\toprule
Diagnostic & Training-free & Mean $|r|$ \\
\midrule
$\mathcal{D}_k$ (ours)                           & yes & \textbf{0.721} \\
Spectral control $S_k = (1-\gamma)^{2k}$          & yes & 0.701 \\
Mean $k$-truncated hitting time                   & yes & 0.702 \\
Mean $k$-truncated commute time                   & yes & 0.702 \\
Variance of $k$-truncated hitting / commute time  & yes & $\approx 0.51$ \\
\bottomrule
\end{tabular}
\end{table}

The untransformed mean variants are approximately linear in $k$ over the tested range (reported $\operatorname{corr}(\bar H_k,k)=1.000$). Likewise $\log S_k=2k\log(1-\gamma)$ is exactly linear in depth within a dataset. These baselines therefore closely track depth. On identical observations and without numerical clipping, $\log S_k$ has exactly the same absolute Pearson correlation with accuracy as depth itself. For the directional comparison with depth, the reported one-sided exact paired Wilcoxon test over the $11$ dataset-level means gives $p=0.027$ (two-sided $p=0.054$).

\subsubsection{Fixed graph and depth, varying the propagation operator}
\label{app:teleport-sweep}

We fix the graph and the depth and vary the laziness parameter of the walk: $\mathbf{T} = \alpha_{\mathrm{lazy}}\mathbf{I} + (1-\alpha_{\mathrm{lazy}})\hat{\mathbf{D}}^{-1}\hat{\mathbf{A}}$, with $\alpha_{\mathrm{lazy}} \in \{0.05, 0.1, 0.2, 0.3, 0.5\}$. Here $\mathbf{T}$ is the one-step lazy kernel of Appendix~\ref{app:gen-archcheck}, and its $k$-step walk distributions are the rows of $\mathbf{T}^k$. We train models using $\mathbf{T}$ at each propagation step on all $11$ datasets at $k \in \{2,4,8,16,24\}$, compute $\mathcal{D}_k(\mathbf{T})$ before training, and within each of the $55$ (dataset, depth) cells correlate $\log\mathcal{D}_k(\mathbf{T})$ with test accuracy \emph{across the five operators}, so neither graph nor depth varies within a correlation.

\begin{table}[H]
\centering
\small
\caption{Fixed (dataset, depth) cells, varying only the laziness parameter. Mean absolute correlation across the five operators. $\mathcal{D}_k(\mathbf{T})$ is the only training-free entry; its observed difference from $E_{\mathrm{Proj}}$ is positive in $28/55$ cells, with median $+0.004$.}
\label{tab:teleport-sweep}
\renewcommand{\arraystretch}{1.1}
\begin{tabular}{lccccc}
\toprule
 & $\mathcal{D}_k(\mathbf{T})$ & $E_{\mathrm{Proj}}$ & MAD & Erank & NumRank \\
\midrule
Mean $|r|$ & \textbf{0.763} & 0.742 & 0.734 & 0.719 & 0.701 \\
\bottomrule
\end{tabular}
\end{table}

Each cell contains only five operators, so individual correlations are noisy; the observed correlations support an association with accuracy as laziness varies at fixed graph and depth.

\subsubsection{Fixed depth, varying graph and operator}
\label{app:fixed-depth}

Last, we ask whether an architecture-associated dispersion score $M_{d,a,k}$ correlates with \emph{relative} accuracy once shared depth variation is removed. At each fixed depth we correlate $\log M_{d,a,k}$ with $\mathrm{acc}(d,a,k)/\max_L\mathrm{acc}(d,a,L)$ across the $132$ (dataset, architecture, activation) combinations from GCN, SGC, GCN+Residual, SAGE, APPNP, and GCN+JKNet, each with two activations on $11$ datasets.

Writing $\mathbf{P}=(\mathbf{D}+\mathbf{I})^{-1}(\mathbf{A}+\mathbf{I})$, GCN and stacked $\mathrm{SGConv}(K=1)$ layers use the dispersion of $\mathbf{P}^k$; GCN+Residual uses $[\tfrac12(\mathbf{I}+\mathbf{P})]^k$; SAGE uses the canonical lazy walk; and APPNP uses $\mathbf{Q}_k$ from Eq.~\eqref{eq:appnp-qk}, with restart probability $0.1$. For GCN+JKNet, we instead use the layer-averaged score
\[
M_{d,\mathrm{JK},k}=\frac1k\sum_{j=1}^{k}\mathcal{D}_j(\mathbf{P}),
\]
which is a summary of layerwise dispersions, not the dispersion of a single propagation matrix.

Averaged over the eight depths, $M_{d,a,k}$ reaches the highest observed mean Pearson correlation ($0.313$ vs.\ $0.288$ for $E_{\mathrm{Dir,n}}$). The observed means are comparable, while our proposed diagnostic requires no training.

\subsection{Architecture-matched operators}
\label{app:arch-matched}

$\mathcal{D}_k$ in Table~\ref{tab:ranking} is computed from the canonical lazy walk for every architecture. This subsection recomputes it from operators associated with each architecture: first for specified row-stochastic operators available without training (\S\ref{app:matched-operator}), then for GIN, whose sum aggregation requires the augmented construction of Appendix~\ref{app:gen-aug} (\S\ref{app:gin-results}).

\subsubsection{Closed-form row-stochastic operators}
\label{app:matched-operator}

We recompute $\mathcal{D}_k$ from specified propagation operators available without training: GCN (row-normalized $\hat{\mathbf{D}}^{-1}\hat{\mathbf{A}}$; the symmetric operator need not have probability rows, Remark~\ref{rem:gcn-bridge}), GraphSAGE-mean (the idealized shared-weight operator $\tfrac12\mathbf{I} + \tfrac12\mathbf{D}^{-1}\mathbf{A}$), APPNP (the restart operator $\mathbf{Q}_k$ of Eq.~\eqref{eq:appnp-qk}), and SGC, which we add for this experiment. GAT and ChebNet are excluded because their learned attention or polynomial filters are not represented here by a fixed training-free stochastic operator. The protocol is otherwise identical to Table~\ref{tab:ranking}.

\begin{table}[H]
\centering
\small
\caption{Mean absolute correlation $|r|$ when $\mathcal{D}_k$ is computed from the lazy walk versus from the specified architecture-associated operator.}
\label{tab:matched-operator}
\renewcommand{\arraystretch}{1.1}
\begin{tabular}{lccc}
\toprule
Architecture & Lazy walk & Matched operator & $\Delta$ \\
\midrule
APPNP & 0.552 & \textbf{0.624} & $+0.072$ \\
SGC   & 0.756 & \textbf{0.779} & $+0.023$ \\
GCN   & 0.793 & \textbf{0.801} & $+0.008$ \\
SAGE  & 0.812 & \textbf{0.813} & $+0.002$ \\
\midrule
Overall & 0.728 & \textbf{0.754} & $+0.026$ \\
\bottomrule
\end{tabular}
\end{table}

APPNP has the largest observed increase ($+0.072$), followed by SGC ($+0.023$) and GCN ($+0.008$). The SAGE difference is implementation noise: its idealized operator is identical to the canonical lazy walk. 

\subsubsection{Sum aggregation: GIN}
\label{app:gin-results}

GIN's sum aggregation produces a non-row-stochastic propagation operator (Appendix~\ref{app:gen-sum}), which places it outside the simplex-geometric framework underlying the lazy-walk $\mathcal{D}_k$. We evaluate the architecture-matched diagnostic $\mathcal{D}_k^{\mathrm{aug}}$ developed in Appendix~\ref{app:gen-aug}.
\begin{table}[H]
\centering
\caption{Mean absolute correlation $|r|$ between $\log(\mathrm{diagnostic})$ and test accuracy, computed across eight depths and averaged over 10 datasets (excluding ogbn-arxiv), for each GIN $\varepsilon$ variant. The augmented diagnostic has the highest observed mean for every tested $\varepsilon$ and Overall. Bold marks the per-row best and underline the second-best.}
\label{tab:gin-results}
\renewcommand{\arraystretch}{1.10}
\small
\begin{tabular}{lccccccccc}
\toprule
GIN variant   & $\mathcal{D}_k^{\mathrm{aug}}$ & $\mathcal{D}_k$ & Erank & NumRank & EDir & EDir,n & EProj & EProj,n & MAD \\
\midrule
$\varepsilon = 0$    & \textbf{0.710} & 0.607 & \underline{0.681} & 0.635 & 0.641 & 0.623 & 0.584 & 0.614 & 0.571 \\
$\varepsilon = 0.5$  & \textbf{0.747} & 0.554 & \underline{0.704} & 0.670 & 0.622 & 0.623 & 0.671 & 0.635 & 0.663 \\
$\varepsilon = 5$    & \textbf{0.751} & 0.573 & \underline{0.724} & 0.666 & 0.609 & 0.677 & 0.598 & 0.618 & 0.658 \\
\midrule
Overall              & \textbf{0.736} & 0.578 & \underline{0.703} & 0.657 & 0.624 & 0.641 & 0.617 & 0.623 & 0.631 \\
\bottomrule
\end{tabular}
\end{table}
The augmented diagnostic has Overall mean $0.736$, compared with $0.578$ for the canonical lazy-walk diagnostic and $0.703$ for Erank, differences of $+0.158$ and $+0.033$, respectively. These results support the augmented construction empirically.

\subsection{Evaluating oversmoothing mitigation techniques}
\label{app:mitigations}

Table~\ref{tab:mitigations-table} reports, for each anti-smoothing variant, the mean absolute correlation between each diagnostic's depth profile and the corresponding depth-accuracy curve, averaged over all 11 datasets and both activations. We stress that this is a \emph{within-variant} correlation. $\mathcal{D}_k$ is defined from the lazy random walk on the test-time graph and is therefore invariant to architectural choices that leave this diagnostic operator unchanged (Bias, PairNorm, LayerNorm, GCNII residuals, JKNet skip-aggregation): the sequence $\{\mathcal{D}_k\}_{k=1}^K$ is identical across these variants on a given dataset. The table consequently does not, and cannot, rank mitigation approaches by $\mathcal{D}_k$. Each column asks a separate question: under this particular mitigation, does the topological dispersion curve still track the depth-accuracy curve?

\begin{table}[H]
\centering
\caption{Mean absolute correlation $|r|$ between each diagnostic and downstream accuracy, computed within each anti-smoothing variant and averaged over all 11 datasets and both activations. Bold marks the per-column best, underline the second-best.}
\label{tab:mitigations-table}
\renewcommand{\arraystretch}{1.10}
\small
\begin{tabular}{lcccccccc}
\toprule
Metric & Vanilla & Bias & PairNorm & GCNII & JKNet & DropEdge & LayerNorm & Average \\
\midrule
$\mathcal{D}_k$ (ours) & \textbf{0.793} & 0.788 & \textbf{0.780} & \textbf{0.816} & 0.503 & \textbf{0.778} & 0.759 & \textbf{0.745} \\
Erank        & 0.705 & \textbf{0.849} & \underline{0.739} & 0.483 & 0.489 & \underline{0.714} & 0.780 & 0.680 \\
NumRank      & 0.671 & \underline{0.823} & 0.733 & 0.633 & \textbf{0.566} & 0.673 & 0.703 & \underline{0.686} \\
EDir         & 0.553 & 0.690 & 0.661 & \underline{0.680} & 0.474 & 0.655 & 0.687 & 0.629 \\
EDir,n       & \underline{0.719}& 0.548 & 0.614 & 0.547 & \underline{0.560} & 0.710 & 0.726 & 0.632 \\
EProj        & 0.496 & 0.632 & 0.614 & 0.577 & 0.488 & 0.554 & \underline{0.816} & 0.597 \\
EProj,n      & 0.544 & 0.717 & 0.614 & 0.540 & 0.523 & 0.604 & \textbf{0.818} & 0.623 \\
MAD          & 0.584 & 0.687 & 0.514 & 0.545 & 0.530 & 0.593 & 0.728 & 0.597 \\
\bottomrule
\end{tabular}
\end{table}

The empirical pattern is that $\mathcal{D}_k$ remains correlated with accuracy across the evaluated variants, with a lower observed mean on JKNet. Feature-side modifications (PairNorm $0.780$, LayerNorm $0.759$, Bias $0.788$), the GCNII residual scheme ($0.816$), and DropEdge ($0.778$) yield mean correlations between $0.759$ and $0.816$, compared with $0.793$ for Vanilla. A possible explanation is that mitigating feature-side collapse makes the effect of propagation-induced mixing on accuracy easier to observe. JKNet combines representations from multiple layers, which can weaken the relationship between deepest-layer mixing and final accuracy; its mean correlations range from $0.474$ to $0.566$ across diagnostics.

Erank, NumRank, EDir, and EProj are computed on trained hidden representations and therefore vary across feature-side mitigations on a fixed graph; $\mathcal{D}_k$ does not. Its highest average correlation across the evaluated variants ($0.745$, versus $0.686$ for NumRank) supports its use as a complementary structural diagnostic. It cannot distinguish feature-side mitigations that leave its chosen propagation operator unchanged.

\subsection{From diagnosis to use}
\label{app:practical-use}

The preceding subsections evaluate $\mathcal{D}_k$ as a diagnostic. This subsection asks how far it supports decisions: choosing a depth (\S\ref{app:depth-selection}, \S\ref{app:controlled-expressivity}), flagging degraded models (\S\ref{app:detection}), and regularizing training (\S\ref{app:regularizer}). Raw values of $\mathcal{D}_k$ have graph-dependent initial and limiting values: they start from an $n$-dependent value, $\mathcal{D}_0 = (n-1)/n$, and on a disconnected graph they converge to a topology-dependent floor rather than to zero (Remark~\ref{rem:components}). The trajectory correlations in Table~\ref{tab:ranking} are computed within each graph. For a common threshold across datasets, we normalize the initial value and limiting floor.

\begin{definition}[Normalized dispersion]
\label{def:normalized-dispersion}
Let $G$ have connected components of sizes $n_1, \ldots, n_M$ with $n = \sum_j n_j$, and let $\mathcal{D}_0 = (n-1)/n$ and $\mathcal{D}_\infty = 1 - \sum_{j=1}^{M} (n_j/n)^2$ denote the initial and limiting dispersion (so $\mathcal{D}_\infty = 0$ when $G$ is connected). Assuming at least one component contains more than one node, the \emph{normalized dispersion} is
\begin{equation}
\label{eq:normalized-dispersion}
    \mathcal{D}_k^{\mathrm{norm}} \;=\; \frac{\mathcal{D}_k - \mathcal{D}_\infty}{\mathcal{D}_0 - \mathcal{D}_\infty} \;\in\; [0, 1].
\end{equation}
\end{definition}

$\mathcal{D}_k^{\mathrm{norm}}$ equals $1$ at depth $0$ and decays to $0$ whenever defined: subtracting $\mathcal{D}_\infty$ removes the between-component contribution, which is constant in $k$, and dividing by $\mathcal{D}_0 - \mathcal{D}_\infty$ rescales the remaining within-component dispersion to a common range. It is the fraction of initial within-component dispersion that remains at depth $k$, and it is undefined only in the degenerate case where every component is a single node. The depth-selection rules below use $\mathcal{D}_k^{\mathrm{norm}}$ throughout.

\subsubsection{Depth selection: a negative result}
\label{app:depth-selection}

A natural question is whether $\mathcal{D}_k$ supports a practical depth-selection rule. We evaluated several candidate rules on the benchmark configurations. None of the evaluated diagnostic-based rules improves on a fixed two-layer choice, so these experiments do not support using $\mathcal{D}_k$ as a standalone depth selector.

\paragraph{Setup.}
For a rule selecting $\hat{k}$ on a (dataset, model) pair, the regret is $\max_k \mathrm{acc}(d,m,k) - \mathrm{acc}(d,m,\hat{k})$, averaged over the $154$ evaluated pairs. $\mathcal{D}_k$-based rules use the normalized trajectory $\mathcal{D}_k^{\mathrm{norm}}$ (abbreviated $\nu_k$ in Table~\ref{tab:depth-rules}). It is evaluated on the fine grid $k = 1, \ldots, K$ and $\hat{k}$ is snapped to the nearest measured depth. Rule parameters are fixed at their proposed values ($\varepsilon = 0.05$, $\tau = 0.5$) and are not tuned on the targets.

\begin{table}[H]
\centering
\small
\caption{Mean regret (pp) over $154$ (dataset, model) pairs. No rule based on $\nu_k = \mathcal{D}_k^{\mathrm{norm}}$ improves on a fixed depth of $2$.}
\label{tab:depth-rules}
\renewcommand{\arraystretch}{1.1}
\begin{tabular}{llc}
\toprule
Rule & Selected depth & Regret \\
\midrule
Fixed $k = 2$   & $\hat{k} = 2$ & \textbf{0.54} \\
Relative decay  & $\hat{k} = \min\{k : (\nu_k - \nu_{k+1})/\nu_k < \varepsilon\}$ & 2.86 \\
Curvature knee  & $\hat{k} = \arg\max_k\,[(\nu_{k-1} - \nu_k) - (\nu_k - \nu_{k+1})]$ & 11.97 \\
Threshold       & $\hat{k} = \max\{k : \nu_k \ge \tau\}$ & 21.14 \\
\bottomrule
\end{tabular}
\end{table}

\paragraph{Results.}
The accuracy-maximizing depth is $k=2$ for $107$ of $154$ pairs, leaving only $0.54$~pp of mean headroom; of the remaining $47$ pairs, $39$ peak at $k=4$. Leave-one-dataset-out calibration of the threshold also fails to improve on fixed depth $2$ in this evaluation: the calibrated threshold selects $k = 4$ on Wisconsin, Cornell, Texas, and Squirrel, yielding $1.05$~pp held-out regret against $0.00$~pp for fixed $k = 2$ on the same datasets. The limitation is not specific to $\mathcal{D}_k$: applying the same evaluation to the seven post-training diagnostics, even granting them the trained models, $0/7$ outperform fixed $k = 2$ (best: Erank, $2.22$~pp). The small advantage available over the fixed two-layer choice limits what depth selection can gain on these benchmarks.

\subsubsection{Controlled feature expressivity and a usable-depth ceiling}
\label{app:controlled-expressivity}

To examine whether feature-side choices contribute to the early accuracy decline, we use orthogonal weight matrices, identity activation, and no dropout, while keeping propagation unchanged. On Cora this raises depth-$24$ accuracy from $14.7\pm0.3$ to $70.2\pm3.0$ ($+55.5$~pp). In this comparison, accuracy declines from $70.2$ to $67.1$ to $32.5$ at depths $24$, $48$, and $96$, respectively. We observe a similar delayed accuracy decline on CiteSeer and PubMed.

In a separate experiment, we ask whether normalized dispersion can identify how deep a controlled linear GCN can go while retaining acceptable accuracy. We use $\mathbf{P}=(\mathbf{D}+\mathbf{I})^{-1}(\mathbf{A}+\mathbf{I})$ and test depths $\{2,4,\ldots,50\}$. For each graph, the target $k^\dagger$ is the deepest tested model whose accuracy is within $5$ percentage points of the best tested accuracy.

The rule selects the largest depth at which $\mathcal{D}_k^{\mathrm{norm}}$ is at least a threshold $\tau$. We check this condition at every integer depth from $1$ to $50$, then round the selected depth to the nearest tested depth to obtain $k_{\max}$. The initial threshold $\tau=0.60$ and the $5$~pp tolerance were chosen after inspecting the results, so this analysis is exploratory.

We then make a retrospective leave-one-graph-out comparison, choosing a threshold for each graph using the other two. For each of those two graphs, we take the normalized dispersion at its reported acceptable depth $k^\dagger$ and use the larger of these two values as the threshold for the held-out graph. For Cora, CiteSeer, and PubMed, the reported acceptable depths are $14$, $50$, and $12$, while the rule selects $10$, $34$, and $12$, respectively. None of the selected depths exceeds the corresponding reported limit. The mean ratio $k^\dagger/k_{\max}$ is $1.29$, where $1$ would indicate an exact match. 

These results motivate further study of depth selection when feature transformations are constrained, but do not establish a general depth guarantee. More generally, large $\mathcal{D}_k$ with collapsed representations means that the walk distributions remain diverse while the learned features do not; this can be caused by a degeneracy of the learned weights or caused by the initial node features. Small $\mathcal{D}_k$ indicates substantial walk mixing, which can induce oversmoothing of the representations.

\subsubsection{Detecting accuracy degradation}
\label{app:detection}

We complement the correlation analyses with the decision-level question ``has this model degraded?'' and evaluate whether a score ranks accuracy-degraded observations above the remaining observations using AUROC. We report pooled and stratified comparisons to examine the roles of graph, depth, and operator choice.

\paragraph{Setup.}
A \emph{trajectory} is one configuration on one dataset across depths, with test accuracy averaged over seeds. An observation at depth $k$ is labeled \emph{accuracy-degraded} when its accuracy lies more than $\delta$~pp below the trajectory's best accuracy; we use $\delta=5$ and report sensitivity to $\delta\in\{2,5,10\}$. The $14$ configurations at eight depths across $11$ datasets define $154$ trajectories and $1{,}232$ planned observations. The depth-$24$ observation for LeakyReLU+SAGE on ogbn-arxiv is not included, leaving $1{,}231$ observations, of which $646$ are labeled accuracy-degraded at $\delta=5$.

Scores (larger = more degraded) are the log-decay from the depth-$2$ reference. They use raw dispersion relative to that reference. With $\eta=10^{-15}$, the implemented scores are
\[
\begin{aligned}
s_D(k)&=-\bigl[\log\max\{\mathcal{D}_k,\eta\}-\log\max\{\mathcal{D}_2,\eta\}\bigr],\\
s_E(k)&=-\bigl[\log\max\{E_k-1,\eta\}-\log\max\{E_2-1,\eta\}\bigr],
\end{aligned}
\]
where $E_k$ denotes effective rank. Depth $k$ itself serves as a control. The matched score uses the same log-decay construction with $M_k$, the architecture-associated score of \S\ref{app:fixed-depth}, in place of $\mathcal{D}_k$. Values are floored before taking logarithms, and non-finite scores are excluded. Ties receive half credit. Stratified AUROCs weight each cell by its number of positive--negative observation pairs; cells containing only one label contribute no pairs.

\paragraph{Pooled and depth-stratified comparisons.}
Pooling all observations (Table~\ref{tab:auroc-pooled}), $\mathcal{D}_k$ has the lowest observed AUROC of the three scores. Depth alone reaches $0.780$ per observation and $0.964$ after model averaging. Thus much of the pooled discrimination is associated with depth. The positive-label frequencies are $0.019$, $0.162$, $0.481$, $0.636$, $0.682$, $0.695$, $0.740$, and $0.784$ at depths $2$, $4$, $6$, $8$, $10$, $12$, $16$, and $24$, respectively. The depth and fixed-walk scores are non-decreasing with depth where defined; effective-rank scores from separately trained models need not be.

\paragraph{Stratifying by depth compares observations only at the same depth.} This gives the depth control AUROC $0.500$, while the lazy-walk score has AUROC $0.343$. At a fixed depth, $\mathcal{D}_k$ varies only across graphs. These graphs differ in task difficulty as well as mixing: RomanEmpire retains $\mathcal{D}_{24}^{\mathrm{norm}}=0.998$ and loses approximately $40$~pp, whereas Texas has $\mathcal{D}_{24}^{\mathrm{norm}}=0.048$ and loses approximately $4$~pp. On the same observations, using the architecture-associated score increases pooled AUROC from $0.588$ to $0.613$ and depth-stratified AUROC from $0.378$ to $0.441$ (Table~\ref{tab:auroc-stratified}).

\begin{table}[H]
\centering
\small
\caption{AUROC for detecting accuracy-degraded observations ($\delta=5$~pp). All three scores use the same $1{,}231$ observations; the model-averaged comparison has $88$ groups.}
\label{tab:auroc-pooled}
\renewcommand{\arraystretch}{1.1}
\begin{tabular}{lccc}
\toprule
 & $\mathcal{D}_k$ (lazy) & Erank & Depth (control) \\
\midrule
Pooled, per observation ($1{,}231$) & 0.602 & 0.810 & 0.780 \\
Pooled, model-averaged ($88$ groups) & 0.802 & 0.904 & \textbf{0.964} \\
Depth-stratified & 0.343 & 0.698 & 0.500 \\
\bottomrule
\end{tabular}
\end{table}

\paragraph{Graph- and depth-stratified comparison.}
The comparison that holds both graph and depth fixed asks: among architectures on the same graph at the same depth, does the one whose structural score loses more dispersion relative to depth $2$ degrade more? The lazy-walk score is constant within a (graph, depth) cell, so it ties on every comparison and scores $0.500$ by construction; only the operator-matched form can distinguish observations within these cells. By restricting ourselves to architectures for which we can compute architecture-matched aggregation dispersion, we end up having 703 observations. Table~\ref{tab:auroc-stratified} shows that, when graph and depth are held fixed, the matched score attains observed AUROC $0.752$, against a $0.500$ control. Effective rank remains higher ($0.846$), but it is computed from a trained model's representations, whereas $M_k$ requires only the graph and the specified propagation construction and is available before any training. The ordering is stable across labeling thresholds (Table~\ref{tab:auroc-delta}); both diagnostics improve monotonically as $\delta$ grows, although changing $\delta$ also changes the positive class and the set of cells containing both labels.

\begin{table}[H]
\centering
\small
\caption{AUROC comparisons on the same $703$ observations for all four scores ($\delta=5$). $\dagger$: constant within every cell, hence $0.500$ by construction. Bold marks the highest training-free (graph, depth)-stratified score.}
\label{tab:auroc-stratified}
\renewcommand{\arraystretch}{1.1}
\begin{tabular}{lcccc}
\toprule
Score & Pooled & Depth-strat. & (graph, depth)-strat. & Training-free \\
\midrule
$\mathcal{D}_k$, lazy walk & 0.588 & 0.378 & $0.500^{\dagger}$ & yes \\
$M_k$, matched & 0.613 & 0.441 & \textbf{0.752} & yes \\
Effective rank $E_k$ & 0.841 & 0.798 & 0.846 & no \\
Depth (control) & 0.737 & 0.500 & 0.500 & -- \\
\bottomrule
\end{tabular}
\end{table}

\begin{table}[H]
\centering
\small
\caption{Sensitivity of the (graph, depth)-stratified AUROC to the labeling threshold $\delta$. Observations labeled accuracy-degraded: $439$, $330$, $205$ of $703$; cells containing both labels: $75$, $75$, $57$ of $88$.}
\label{tab:auroc-delta}
\renewcommand{\arraystretch}{1.1}
\begin{tabular}{lccc}
\toprule
Score & $\delta = 2$ & $\delta = 5$ & $\delta = 10$ \\
\midrule
$\mathcal{D}_k$, lazy walk           & $0.500^{\dagger}$ & $0.500^{\dagger}$ & $0.500^{\dagger}$ \\
$M_k$, matched  & 0.709 & 0.752 & 0.765 \\
Effective rank $E_k$                 & 0.754 & 0.846 & 0.875 \\
Depth (control)                      & 0.500 & 0.500 & 0.500 \\
\bottomrule
\end{tabular}
\end{table}

\subsubsection{\texorpdfstring{$\mathcal{D}_k$}{D\_k} as a regularizer}
\label{app:regularizer}

Optimizing $\mathcal{D}_k$ requires its propagation operator to depend on the trained parameters: for fixed propagation (GCN, GraphSAGE, APPNP) it depends only on the graph and fixed propagation settings, so $\nabla_\theta\mathcal{D}_k=0$. For GAT we compute dispersion from learned attention operators and add the hinge penalty $\lambda\,\mathrm{ReLU}(\tau - \mathcal{D})$ with $\tau = 0.5$ in three variants: (i) \emph{endpoint}, penalizing $\mathcal{D}_L$ only; (ii) \emph{prefix}, penalizing several intermediate depths; and (iii) \emph{scheduled prefix}.

Under pure dispersion optimization at $L = 96$, all three objectives converge to similar endpoint dispersion ($\mathcal{D}_{96} = 0.249$--$0.252$ against the target $0.5$). Prefix regularization increases intermediate dispersion (e.g.\ $0.568 \to 0.661$ at $k = 24$) but not the endpoint; at moderate depth ($L = 24$) the same optimization reaches the target. This is consistent with depth-dependent optimization difficulty, potentially related to attention's dependence on the hidden representations. Regularizing every layer was computationally costly in our implementation, and sampling intermediate depths did not resolve the endpoint shortfall. With the task loss included, across $\lambda \in \{0.1, 1, 10, 50\}$ no accuracy improvement is significant (Welch $p \ge 0.18$), large penalties reduce performance at $L = 96$, at $L = 24$ the constraint is already nearly satisfied without the penalty ($\mathcal{D} = 0.485$), and at $L = 96$ dispersion increases by only $\approx 0.005$ even with the strongest penalty. The dispersion target was easier to attain at moderate depth under pure dispersion optimization, but the supervised experiments do not establish an accuracy benefit in the tested settings.

\section{Limitations}
\label{app:limitations}

Aggregation dispersion has two main limitations. First, the original construction requires the propagation operator to be row-stochastic, so architectures with sum aggregation (e.g., GIN) fall outside the framework and need an architecture-matched variant (Appendix~\ref{app:gen-aug}), whose guarantees differ
from those of the original diagnostic; skip-connection architectures such as JKNet and APPNP can weaken the relationship between walk mixing and accuracy, reducing the informativeness of oversmoothing diagnostics including $\mathcal{D}_k$. Second, the theoretical guarantees are conditional: the nonlinear
bound (Theorem~\ref{thm:nonlinear}) assumes $1$-Lipschitz activations and $\|\mathbf{W}^{(\ell)}\|_{\mathrm{op}} \leq 1$, sufficient non-expansiveness conditions that are
not necessarily enforced during training; the linear representation lower bound in Theorem~\ref{thm:linear_bound} is vacuous when $d<n-1$;
and the transfer from stationary-weighted to uniformly
weighted representation dispersion incurs a degree-heterogeneity factor $\rho = \pi_{\max}/\pi_{\min}$
that can be large on heavy-tailed graphs. These theoretical guarantees concern representation
dispersion and do not by themselves imply accuracy guarantees.

These limitations affect architectures differently, depending on how they propagate information and combine representations across layers. Indeed, the guarantees for $\mathcal{D}_k$ hold with different strength for different architectures. Theorem~\ref{thm:lazy-universality} extends the spectral rate comparison beyond the canonical lazy walk under its stated graph assumptions, conditions (C1)--(C3), either (C4) or (C4$'$), and (C5). This gives four cases.

\emph{Covered by theory:}
For example, the idealized GraphSAGE-mean operator with equal root and neighbor weights and a shared feature transformation, and row-normalized GCN and the lazy kernel when the required spectral conditions hold (Appendix~\ref{app:gen-archcheck}); for GCN with symmetric normalization, the bridge transfers rates of convergence to rank-one limits, rather than dispersion values (Remark~\ref{rem:gcn-bridge}).

\emph{Partial theoretical coverage:}
learned attention operators (GATv1, GATv2), which can be non-reversible and change across layers. The collapse characterization applies to row-stochastic attention products, but the fixed-operator spectral comparison does not apply, and mean dispersion need not decrease with depth (Appendix~\ref{app:gen-attention}). Empirically, the canonical lazy-walk $\mathcal{D}_k$ has the highest observed mean absolute correlation on both attention architectures
(Table~\ref{tab:per_arch}).

\emph{Requiring separate constructions:}
non-row-stochastic propagation such as GIN's sum aggregation, which needs the architecture-matched $\mathcal{D}_k^{\mathrm{aug}}$
(Appendix~\ref{app:gen-aug}), and APPNP's restart propagation, which is not a power of a fixed operator. APPNP instead requires a depth-dependent operator $\mathbf{Q}_k$, whose row-stochastic form still admits the Hellinger diagnostic but has a positive limiting dispersion
(Appendix~\ref{app:gen-appnp}).

\emph{Multi-layer readouts:}
JKNet combines representations from multiple layers, which can weaken the relationship between deepest-layer mixing and final accuracy.

\end{document}